\RequirePackage{fix-cm}
\documentclass[smallextended]{svjour3}      

\usepackage{amsmath,amsfonts,bm}

\def\eqref#1{equation~\ref{#1}}

\def\1{\bm{1}}

\DeclareMathAlphabet{\mathsfit}{\encodingdefault}{\sfdefault}{m}{sl}
\SetMathAlphabet{\mathsfit}{bold}{\encodingdefault}{\sfdefault}{bx}{n}

\usepackage{hyperref}
\usepackage{url}
\usepackage[final]{graphicx}      
\usepackage{subcaption}
\usepackage{wrapfig}
\usepackage{color}
\usepackage{amssymb}
\usepackage[misc]{ifsym}
\newcommand{\citep}[1]{\cite{#1}}

\begin{document}
    
\title{Exploring the Common Principal Subspace of Deep Features in Neural Networks}
\author{Haoran Liu\and Haoyi Xiong\and Yaqing Wang\and Haozhe An\and Dongrui Wu\and Dejing Dou
\thanks{The first author is affiliated to Texas A\&M University, College Station, TX at present. This work was done when the first author was an intern under supervision from the second author at Baidu Research. Corresponding Author: Haoyi Xiong, xionghaoyi@baidu.com.}}

\authorrunning{H. Liu, H. Xiong, Y. Wang, H. An, D. Wu, and D. Dou} 

\institute{
            H. Liu\and H. Xiong~\Letter \and Y. Wang \and D. Dou \at
              Big Data Lab, Baidu Research \\
              \email{xionghaoyi@baidu.com}           
           \and
           H. An\at
           Department of Computer Science,\\
           University of Maryland, College Park, MD
           \and
           D. Wu \at
           School of Artificial Intelligence and Automation\\
             Huazhong University of Science and Technology
}

\date{Received: date / Accepted: date}
\maketitle
\begin{abstract}
We find that different Deep Neural Networks (DNNs) trained with the same dataset share a common principal subspace in latent spaces, no matter in which architectures (e.g., Convolutional Neural Networks (CNNs), Multi-Layer Preceptors (MLPs) and Autoencoders (AEs)) the DNNs were built or even whether labels have been used in training (e.g., supervised, unsupervised, and self-supervised learning). Specifically, we design a new metric $\mathcal{P}$-vector to represent the principal subspace of deep features learned in a DNN, and propose to measure angles between the principal subspaces using $\mathcal{P}$-vectors. Small angles (with cosine close to $1.0$) have been found in the comparisons between any two DNNs trained with different algorithms/architectures. Furthermore, during the training procedure from random scratch, the angle decrease from a larger one ($70^\circ-80^\circ$ usually) to the small one, which coincides the progress of feature space learning from scratch to convergence. Then, we carry out case studies to measure the angle between the $\mathcal{P}$-vector and the principal subspace of training dataset, and connect such angle with generalization performance. Extensive experiments with practically-used Multi-Layer Perceptron (MLPs), AEs and CNNs for classification, image reconstruction, and self-supervised learning tasks on MNIST, CIFAR-10 and CIFAR-100 datasets have been done to support our claims with solid evidences.

\end{abstract}

\section{Introduction}
Blessed by the capacities of feature learning, deep neural networks~\citep{lecun2015deep} have been widely used to perform learning in various learning settings (e.g., supervised, unsupervised, and self-supervised learning), ranging from classification, to generation~\citep{goodfellow2014generative,radford2015unsupervised}. To better euclid the features learned by deep models, numerous works have studied on interpreting the features spaces of the well-trained models as follows.

\paragraph{\bf\em Backgrounds and Related Works.}
As early as 2013,~\citep{simonyan2013deep} proposed to visualize the features learned by deep convolutional neural networks (CNN) and made sense of discriminative learning via deep feature extraction. 
For generative models,~\citep{white2016sampling,zhu2016generative} studied the interpolation of latent spaces while~\citep{zhu2016generative} discovered an user-controlled way to manipulate the images generated through the surrogation of latent spaces via manifolds.
Later,~\citep{netdissect2017} presented the visual concepts learned in the feature spaces of discriminative models through network dissection on specific datasets while the same group of researchers also proposed GAN dissection~\citep{bau2019gandissect} -- an interactive way to manipulate the semantics and style of image synthesis.
~\citep{richardson2018gans} compared GAN and Gaussian Mixture Models (GMMs) to understand the capacity of distribution learning in GAN. ~\citep{berthelot2018understanding} proposed to improve understanding and interpolation of Autoencoders using adversarial regularizer while~\citep{spinner2018towards} compared AEs with its variational derivatives to interpret the latent spaces.
More recently, the authors in~\citep{Jahanian2020On} studied the ``steerability'' of GAN, where authors discovered point-to-point editing paths for content/style manipulation.~\citep{Zhang2020Empirical} uncovered the phenomena that DNN classifiers with piecewise linear activation tend to map the input data to linear subregions. Apparently, many impressive studies are not well discussed here~\citep{nguyen2016synthesizing,arvanitidis2018latent,sercu2019interactive}.

While existing studies primarily focus on the interpolation of a given model to discover mappings from the feature space to outputs of the model (e.g., classification~\citep{netdissect2017} and generation~\citep{Jahanian2020On}), the work is so few that compares the feature spaces learned by deep models of varying architectures (e.g., MLP/CNN classifiers versus Autoencoders) for different learning paradigms (e.g., supervised/self-supervised classification~\citep{chen2020simple,khosla2020supervised}, unsupervised data reconstruction or de-noising~\citep{spinner2018towards}, etc.). Furthermore, the dynamics of feature space learning over the training procedure still has not been known yet. In our research, we aim at compare feature spaces of DNN models based on various architectures/paradigms and try to understand how the space evolves during the training process and how feature space learning connects to data distributions and performance of models.

\paragraph{\bf\em Intuitions and Hypotheses.}  Based on the same training dataset, it is not difficult to imagine well-trained DNN classifiers of various architectures in supervised learning settings would map these samples into feature vectors that share certain linear subspaces. To yield an appropriate analysis, let simply push back from the classification results -- in any well-trained model, the feature vectors should be capable of being projected to the same set of ground-truth labels (as well-trained models fit training datasets well) through a Fully-Connected (FC) Layer (i.e., a linear transform) and/or a Softmax operator, even though the parameters of networks are different. We further doubt that such subspace might be not only shared by supervised learners but also with Autoencoders (AEs) which are trained to reconstruct input data without any label information in an unsupervised manner, or even shared with self-supervised DNN classifiers (e.g., SimCLR~\citep{chen2020simple}) which train CNN feature extractor and the classifiers separately in an ad-hoc manner.

We specifically hypothesize that (\textbf{H.I}:) there exists certain {\emph{common subspace}} shared by the feature spaces of well-trained deep models using the same training datasets, even though the architectures (MLPs, CNNs, and AEs) and the learning paradigms (supervised, unsupervised, and self-supervised) are significantly different. As the training procedure usually initializes the DNN models from random weights, we further hypothesize that (\textbf{H.II}:) there might exist a process of convergence in feature space learning over training iterations, where feature spaces in the early stage of training procedure would vary from each other for different random initializations however they would gradually evolve to shape the common subspace for convergence. Finally, we hypothesize that (\textbf{H.III}:) the convergence to the shared common subspace would connect to the data distribution and performance of models, as such behavior indicates how well the features are learned from data.

\paragraph{\bf\em Results and Contributions.} 
To test above three hypotheses, this work makes contributions in proposing new measures to the DNN features and conducting extensive experiments for empirical studies. Given a set of feature vectors extracted from a DNN using either the training or testing datasets, we form a \#samples $\times$ \#features matrix\footnote{\#samples and \#features refer to the numbers of samples and features respectively.}, then we use the top left singular vector of such matrix as the vector representing the the principal subspace~\cite{abdi2010principal}  (for more about feature extraction from different architectures and computation, please refer to \textbf{Section 2}).
For convenient, we name the principal subspace as the \textbf{$\mathcal{P}$-vector} of the model based on training/testing sets accordingly.

We train deep models using various DNN architectures, multiple learning paradigms, and datasets, with the checkpoint restored per epoch. Then, we use these models to discover some interesting phenomenons. Results and evidences to support the three hypotheses are summarized as follows.
 
\emph{(1)~Small Angles between Principal Subspaces of Deep Features Learned by Various DNNs (\textbf{H.I}).}~
%
Given multiple DNNs trained using the same dataset, we estimate their $\mathcal{P}$-vectors and calculate the angles between their $\mathcal{P}$-vectors to measure the similarity in their deep feature spaces. Our experiments find that the angles between $\mathcal{P}$-vectors of these DNNs are small with a high cosine similarity close to $1$, no matter how DNNs are trained or modeled. For example, Figure 1 (a)--(b).~show the high cosine measure (close to $1$) for angles between $\mathcal{P}$-vectors of any two well-trained DNNs (CIFAR-10) for eight different architectures/tasks. The experiment results confirm the existence of common subspaces in deep features learned from the same training set, which is invariant under various architectures/tasks.

\begin{figure*}
\centering
\subfloat[Cosine (Train)]{\includegraphics[width=0.5\textwidth]{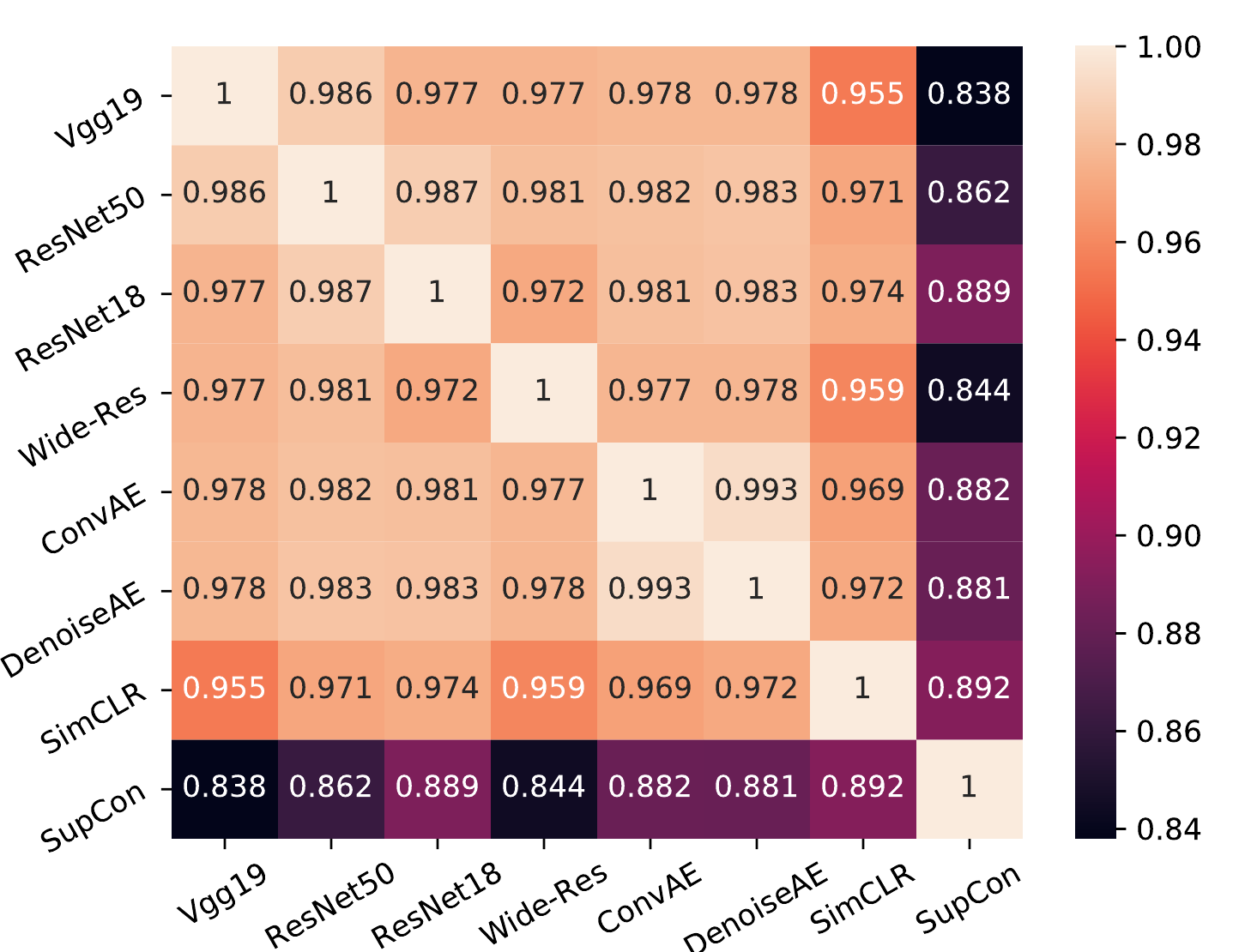}}
\subfloat[Cosine (Test)]{\includegraphics[width=0.5\textwidth]{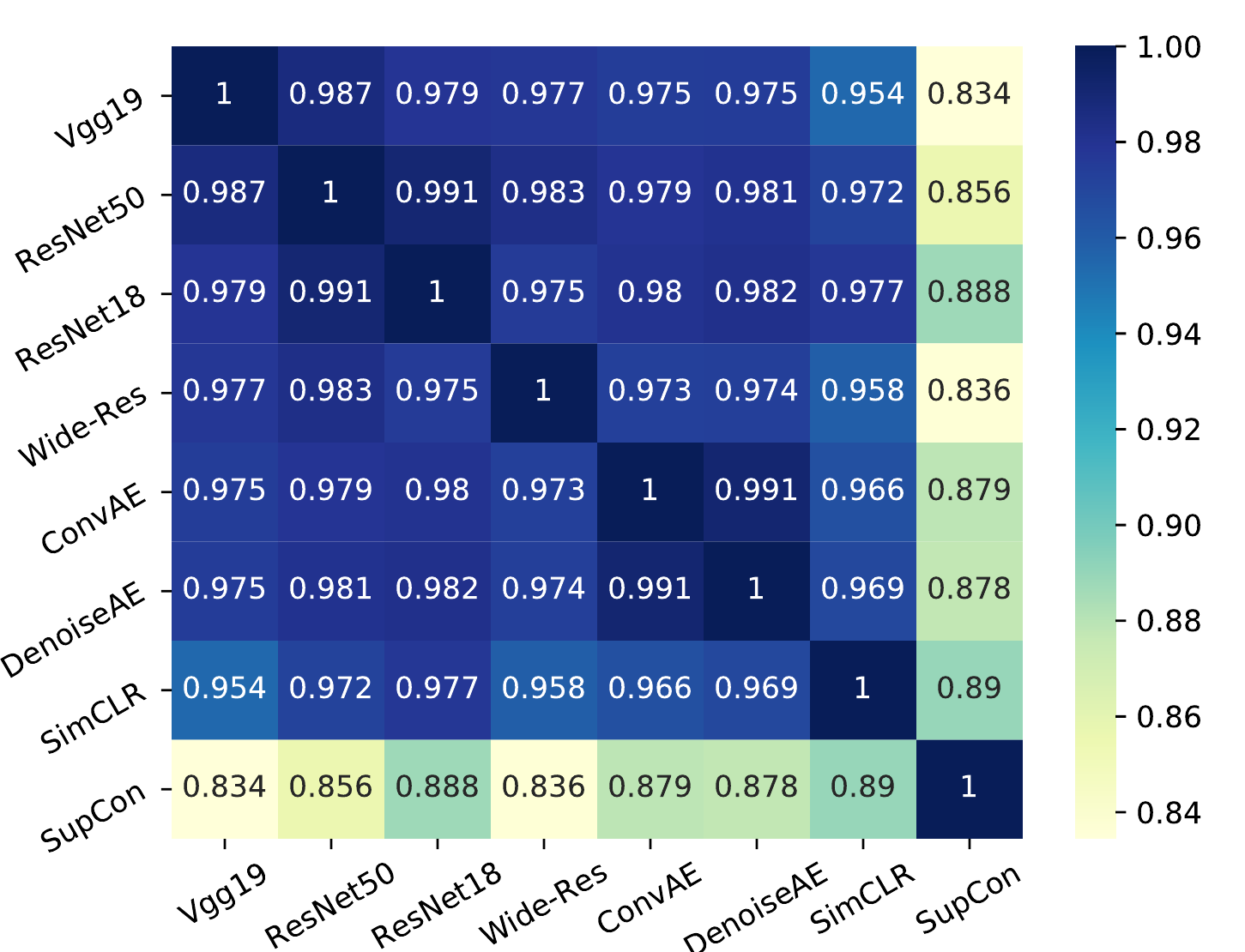}}\vspace{-2mm}
\caption{\textbf{Common Principal Subspaces.} We present cosine (in the range of [0,1]) of angles between principal subspaces of well-trained models in various architectures under different learning paradigms, where angles are measured as the angle between $\mathcal{P}$-vectors based on training and testing datasets respectively. \emph{A well-trained model here is the one trained under the suggested settings after 200 epochs for supervised/self-supervised CNN classifiers and 100 epochs for unsupervised AEs.} Experiments are carried out in 5 independent trials (5 random seeds) with averaged results reported. 
}\vspace{-5mm}
\label{fig:claim1-intro1}
\end{figure*}

\begin{figure*}
\subfloat[Supervised (Train)]{\includegraphics[width=0.45\textwidth]{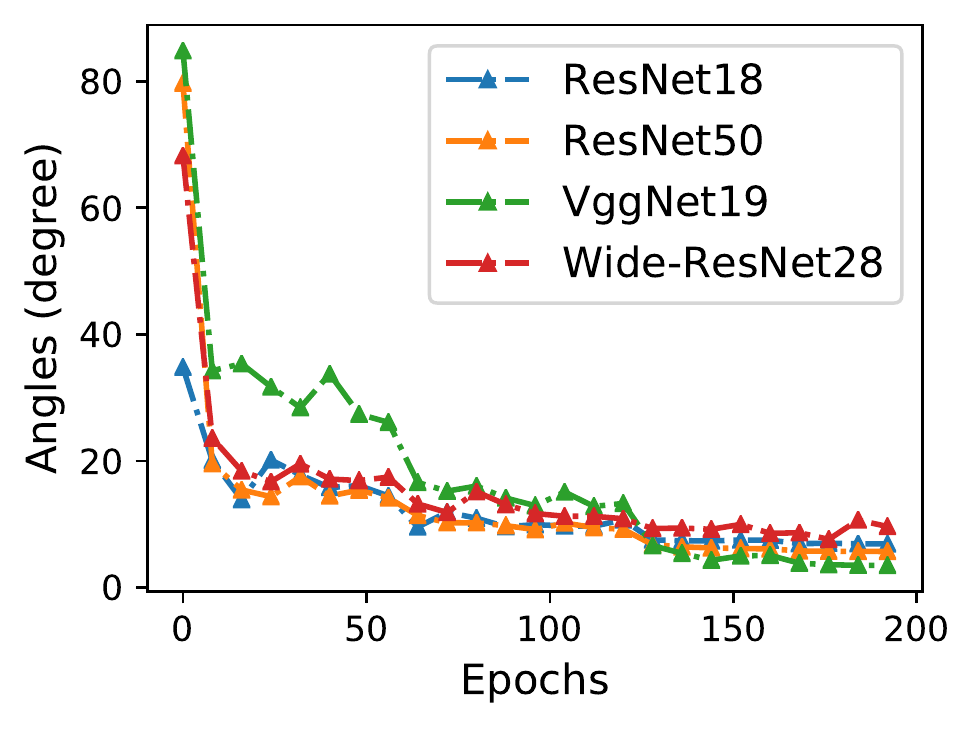}}
\subfloat[Supervised (Test)]{\includegraphics[width=0.45\textwidth]{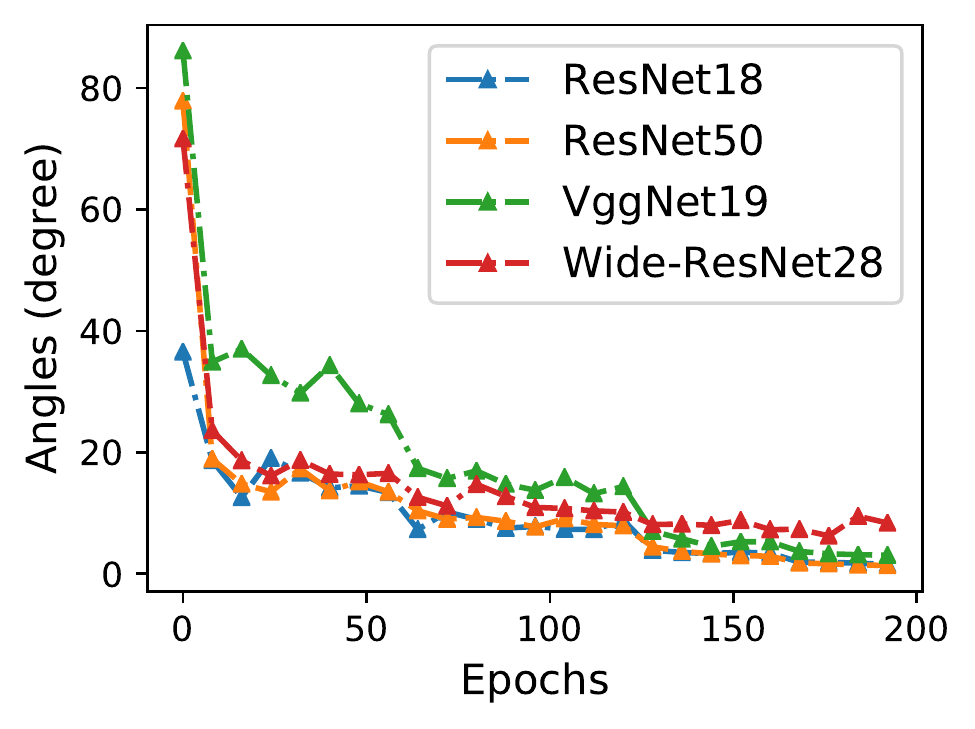}}\\
\subfloat[Unsupervised (Train)]{\includegraphics[width=0.45\textwidth]{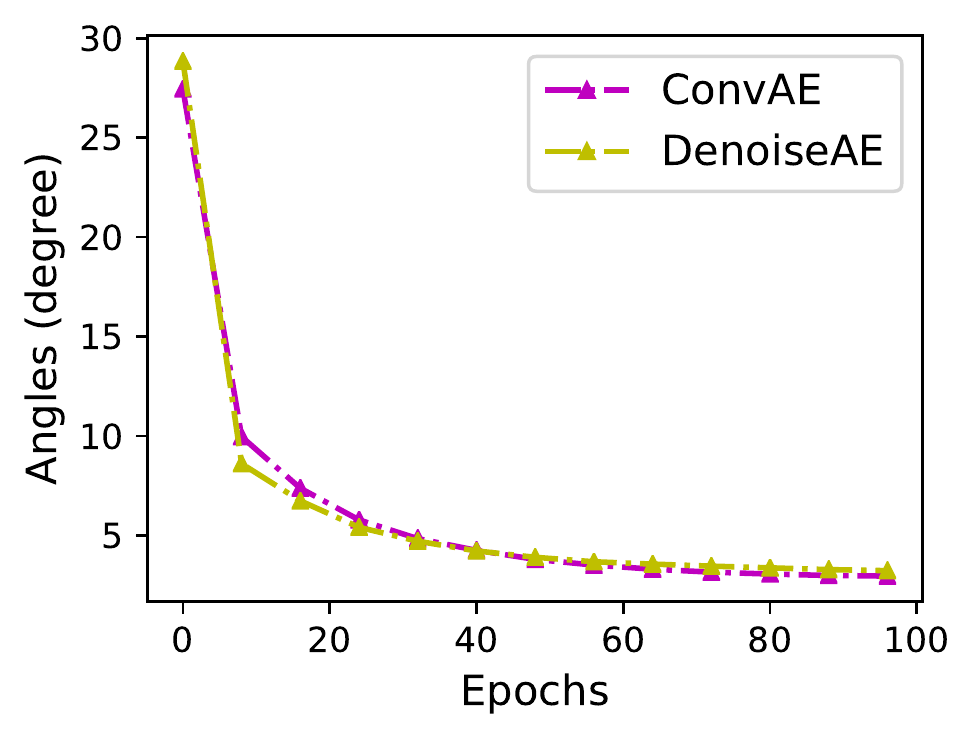}}%
\subfloat[Unsupervised (Test)]{\includegraphics[width=0.45\textwidth]{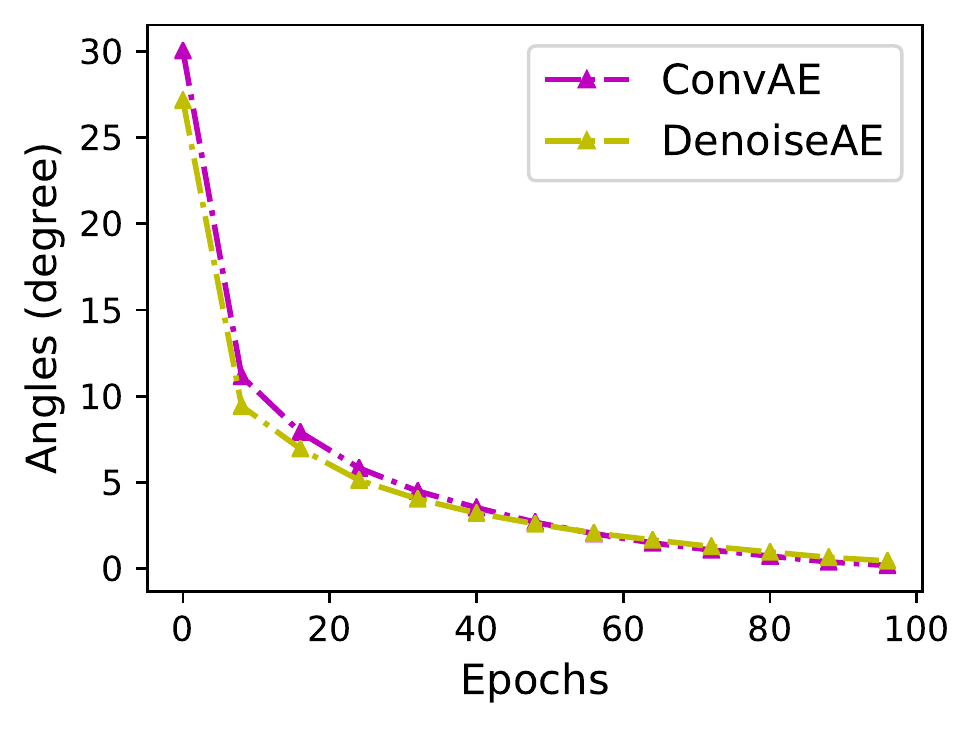}}\\
\subfloat[Self-Sup. (Train)]{\includegraphics[width=0.45\textwidth]{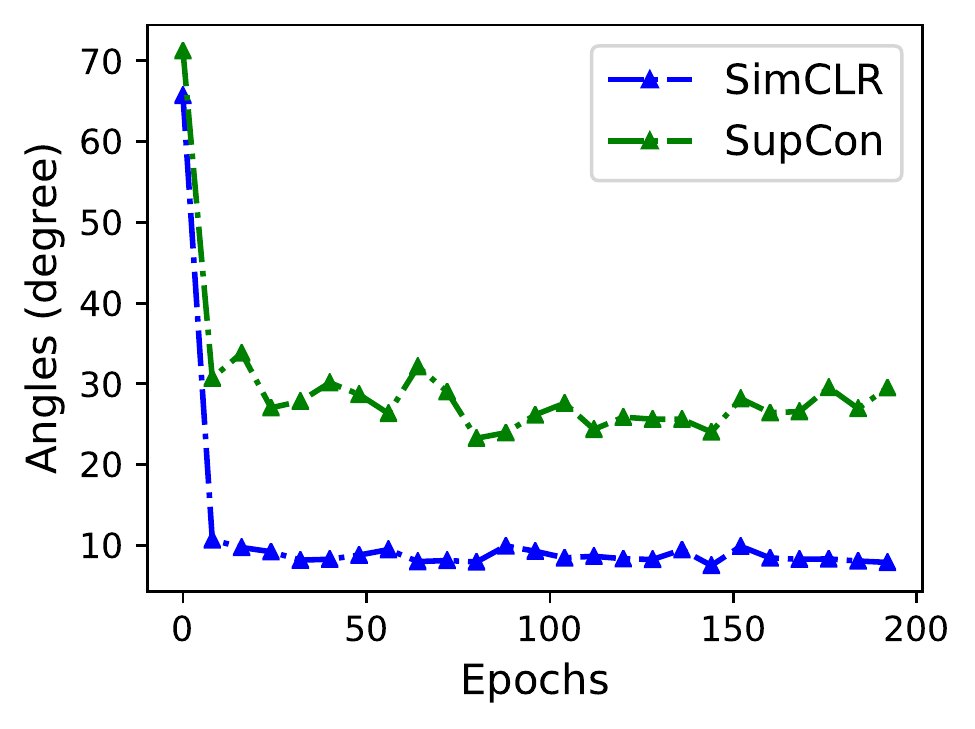}}
\subfloat[Self-Sup. (Test)]{\includegraphics[width=0.45\textwidth]{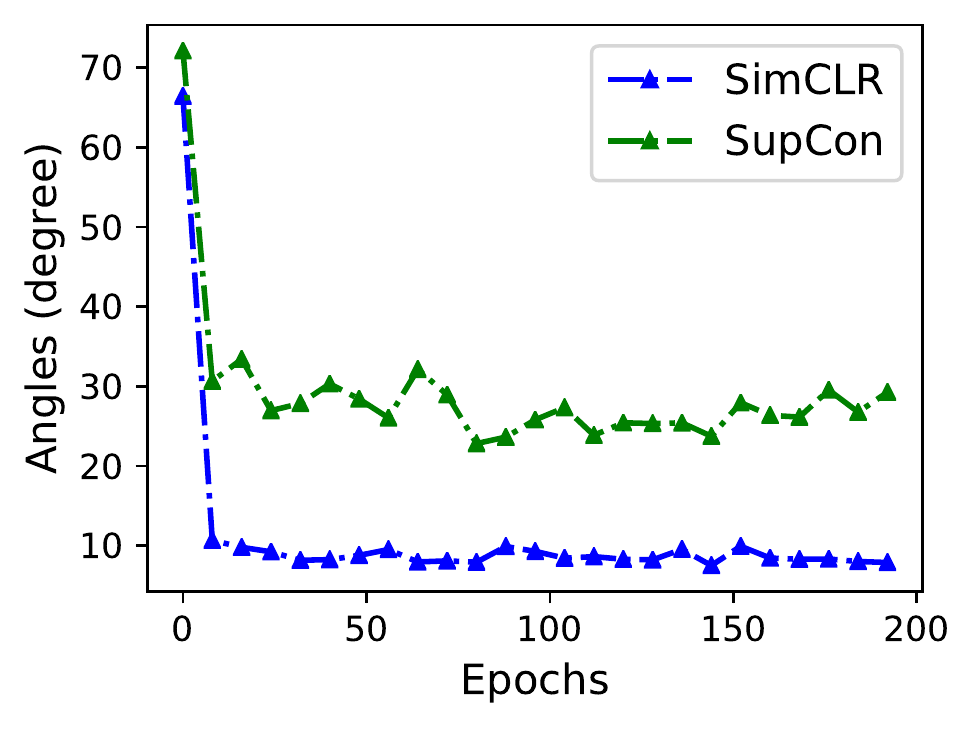}}
\caption{\textbf{Convergence to the Common Subspace measured using Angles between $\mathcal{P}$-vectors based on CIFAR-10.} We present angles between principal subspaces of the well-trained model and the checkpoint per training epoch (from 0$^{th}$--199$^{th}$ epochs) of three learning paradigms.
}
\label{fig:claim1-intro2}
\end{figure*}

\emph{(2)~Converging Trends of Angles over the Number of Training Epochs (\textbf{H.II}).} For any model in the training procedure, we checkout their checkpoints restored after every epoch and extract a $\mathcal{P}$-vector for every checkpoint. Our experiments find that, the principal subspace of deep features at beginning of a training procedure is very different with the well-trained one, but it would slowly converge to the well-trained one and get closer and closer to the common subspace over epochs. For example, Figure~\ref{fig:claim1-intro2} (a)--(f) demonstrate the angles between the $\mathcal{P}$-vectors of the well-trained model and the checkpoint per epoch, where consistent converging trends from large angles to small ones (e.g., $\approx10^\circ$ for supervised CNN classifiers and SimCLR; $\leq 10^\circ$ for ConvAEs/DenoiseAEs; and $\approx30^\circ$ for SupCon~\cite{khosla2020supervised}) could be found in the comparisons among eight different architectures/tasks.

\emph{(3)~The Angle between Principal Subspaces of Deep Features and Data Distribution and its Connections to Generalization (\textbf{H.III}).} We compare principal subspaces between deep features (i.e., $\mathcal{P}$-vector) and raw data (i.e., the top left singular vector of the \#samples $\times$\#dimensions data matrix). For convenience, we name them as model $\mathcal{P}$-vector and data $\mathcal{P}$-vector respectively, and measure the angles between data and model $\mathcal{P}$-vectors, as the proxy measurement of angles between the principal subspaces of deep features and data. We also find the converging trends of such angles over training epochs, where we can again see the well-trained models would incorporate smaller angles ($\approx50^\circ$) than ones in the early stage of training processes ($\approx80^\circ$). We correlate such angles with training/testing accuracy of DNNs, where we observe significant negative correlations in most cases of experiments. We further demonstrate the feasibility of using such angles to predict the generalization performance of a DNN model with their training data only.

\paragraph{\bf\em Discussions.}
The most relevant studies to our work are~\citep{netdissect2017,Zhang2020Empirical,saxe2019mathematical,lee2019mathematical}. For discriminative models,~\citep{netdissect2017} recovered visual features learned by CNN classifiers with a priorly labeled dataset, and quantified then compared the feature learning capacities (namely ``interpretability'' in the work) of different DNN models through patterns matching with the ground truth. Compared to~\citep{netdissect2017}, we carry out the empirical studies on a wide range of datasets without any prior information on their features and observe consistent phenomena in the distribution of samples in the feature spaces. 

Furthermore, while~\citep{Zhang2020Empirical} studied the properties of regions where a supervised DNN classifier with piecewise linear activation behaves linearly, our work observes the common linear subspaces shared by the features learned by the networks that are trained with different architectures (e.g., MLP/CNN classifiers and AEs with ReLU activation) and paradigms (e.g., supervised, unsupervised, and self-supervised learning). Furthermore, both our work and~\citep{saxe2019mathematical} compare feature/representation through SVD, while we perform SVD to investigate the distribution of samples in principal subspace of deep features and~\citep{saxe2019mathematical} uncovered the latent structures in input-and-output. To the best of our knowledge, we make unique contributions compared to the above work.

\section{Measuring Principal Subspaces of Deep Features with $\mathcal{P}$-vectors}

\subsection{Methodology}
\begin{figure*}
    \centering
    \includegraphics[width=0.99\textwidth]{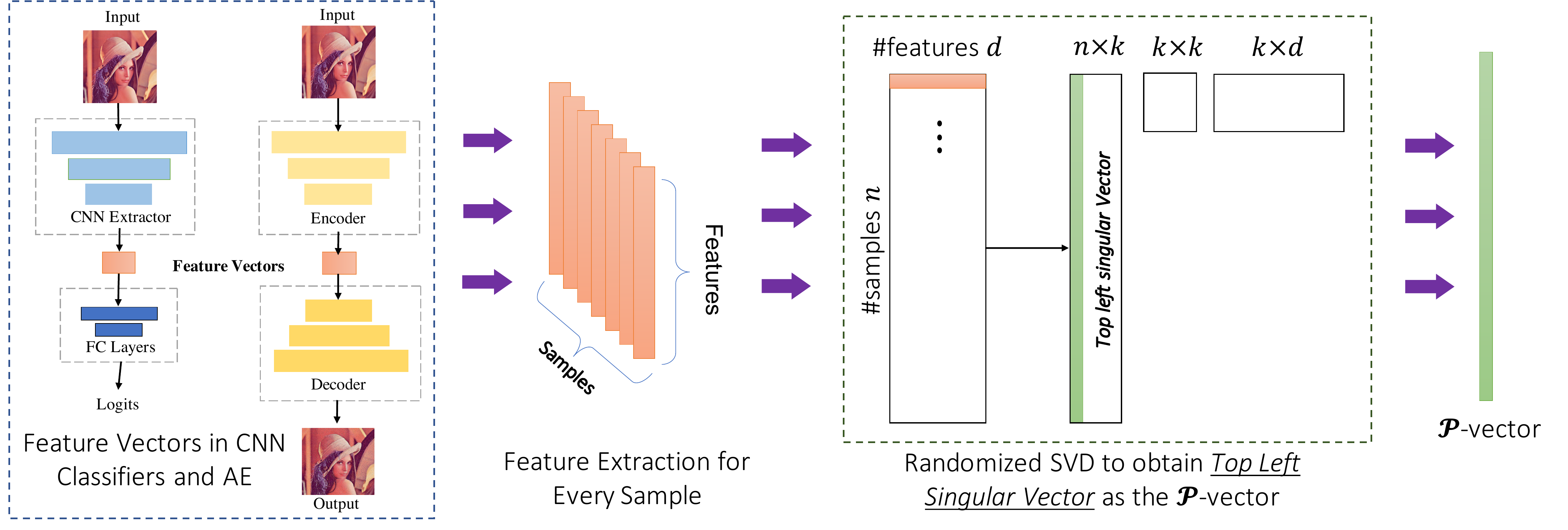}
    \caption{Deep Feature Vectors Extraction and $\mathcal{P}$-vectors Estimation}
    \label{fig:method}
\end{figure*}

As shown in Figure~\ref{fig:method}, we extract deep feature vectors from DNNs and estimate the $\mathcal{P}$-vector in two steps.

\paragraph{\bf\em Feature Vector Extraction.} Given a model, either the well-trained one or a checkpoint obtained during the training process, we extract the feature vector for every sample, with respect to the architectures. For DNN classifiers (either under supervised/self-supervised learning), we use the output of CNN feature extractor (i.e., the input to the Fully-Connected Layer) as the feature vector of the given sample, while we vectorize the output bottleneck layer as the feature vector for AEs. Note that, in our research, we consider AEs with symmetric architectures of encoders and decoders only. 


\paragraph{\bf\em Singular Value Decomposition for $\mathcal{P}$-Vector estimation.} Given the feature vector for every sample, we form a \#samples$\times$\#features matrix and perform SVD to obtain the top left singular vector as the $\mathcal{P}$-vector. Furthermore, there is no need to solve the singular vectors of the complete spectrum, as only the top singular vector is requested for $\mathcal{P}$-vector estimation. In this way, we propose to use Randomized SVD~\citep{halko2011finding} that compresses the feature domain and approximate the low-rank structure of SVD for acceleration purpose. Actually, we compare the numerical solution of Randomized SVD and Common SVD for ResNet-50 on CIFAR-10 dataset (\#features=256 and \#samples= 50,000), where we need to perform SVD on a \#samples$\times$\#features matrix and the $\mathcal{P}$-vector/top left singular vector should be with  dimensions. Compared to the vanilla SVD, around 109x (from 44.88 seconds to 0.41 seconds) speedup has been achieved by Randomized SVD while no significant numerical errors having been found.




\subsection{Measuring Angles between $\mathcal{P}$-Vectors} 
Given the same set of samples, $\mathcal{P}$-vectors of any two DNNs should be in the equal length, as they are both the top singular vectors in the sample side. Thus, we can measure the angle between two $\mathcal{P}$-vectors as a proxy of the angle between two principal subspaces. A larger cosine of the angle between two vectors (e.g., close to 1.0) usually refers to the evidence that the two DNNs share common principal subspace in their features. 

Note that, in high-dimensional spaces, the chance of orthogonality between two random vectors appears more frequently, due to the curse of dimensionality~\citep{pestov1999geometry}. Thus, given a sample set such as CIFAR-10 with 50,000 samples, when the cosine measure close to 1.0 or the angle between the two $\mathcal{P}$-vectors (with 50,000 dimensions) is small, we can conclude that the two networks would share a subspace in the feature spaces in high confidence. 
Of-course, there exists other ways estimating angles between principal subspaces~\cite{bjorck1973numerical}.

\begin{figure}
\centering
\subfloat[CIFAR-10]{\includegraphics[width=0.45\textwidth]{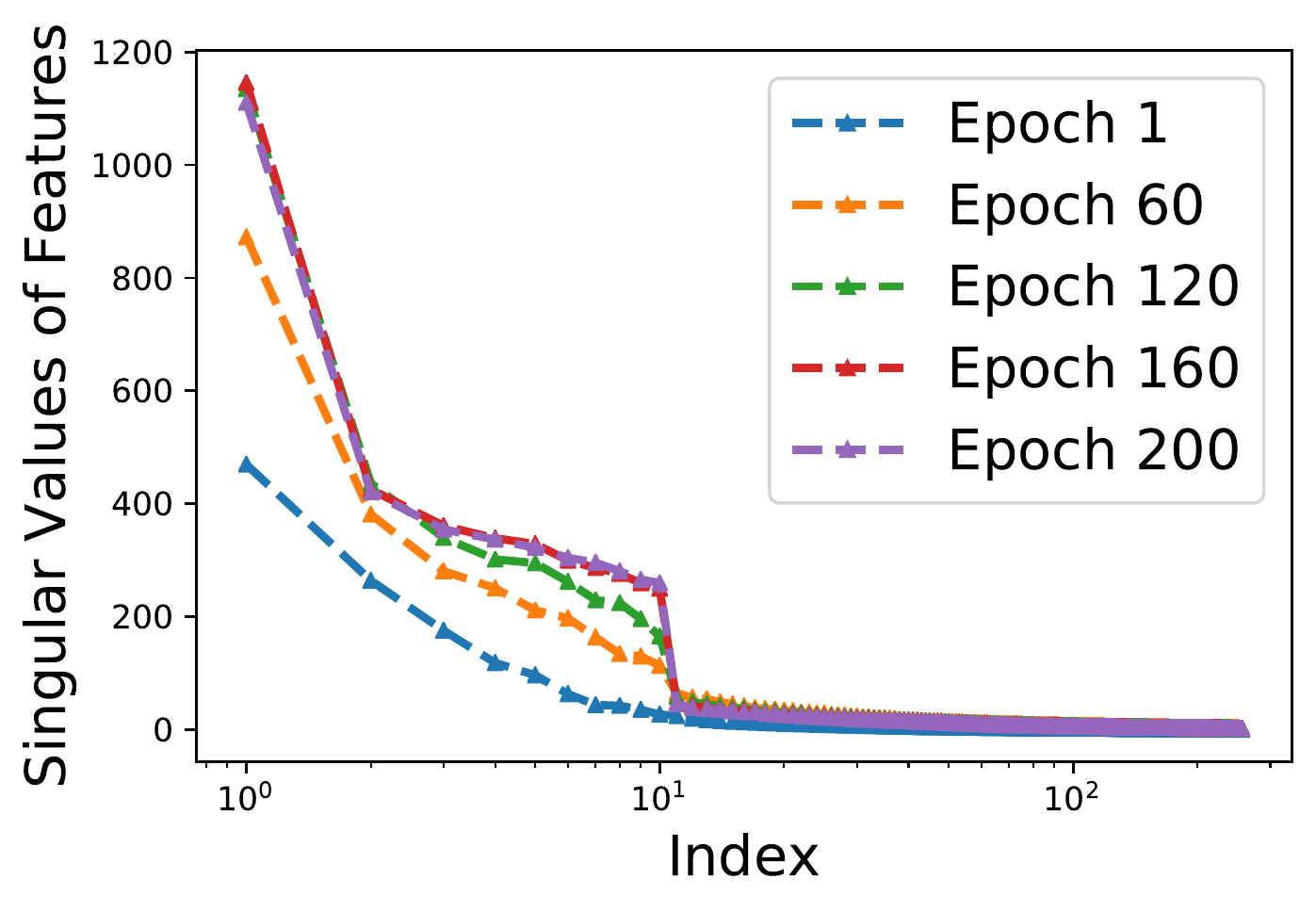}}\
\subfloat[CIFAR-100]{\includegraphics[width=0.45\textwidth]{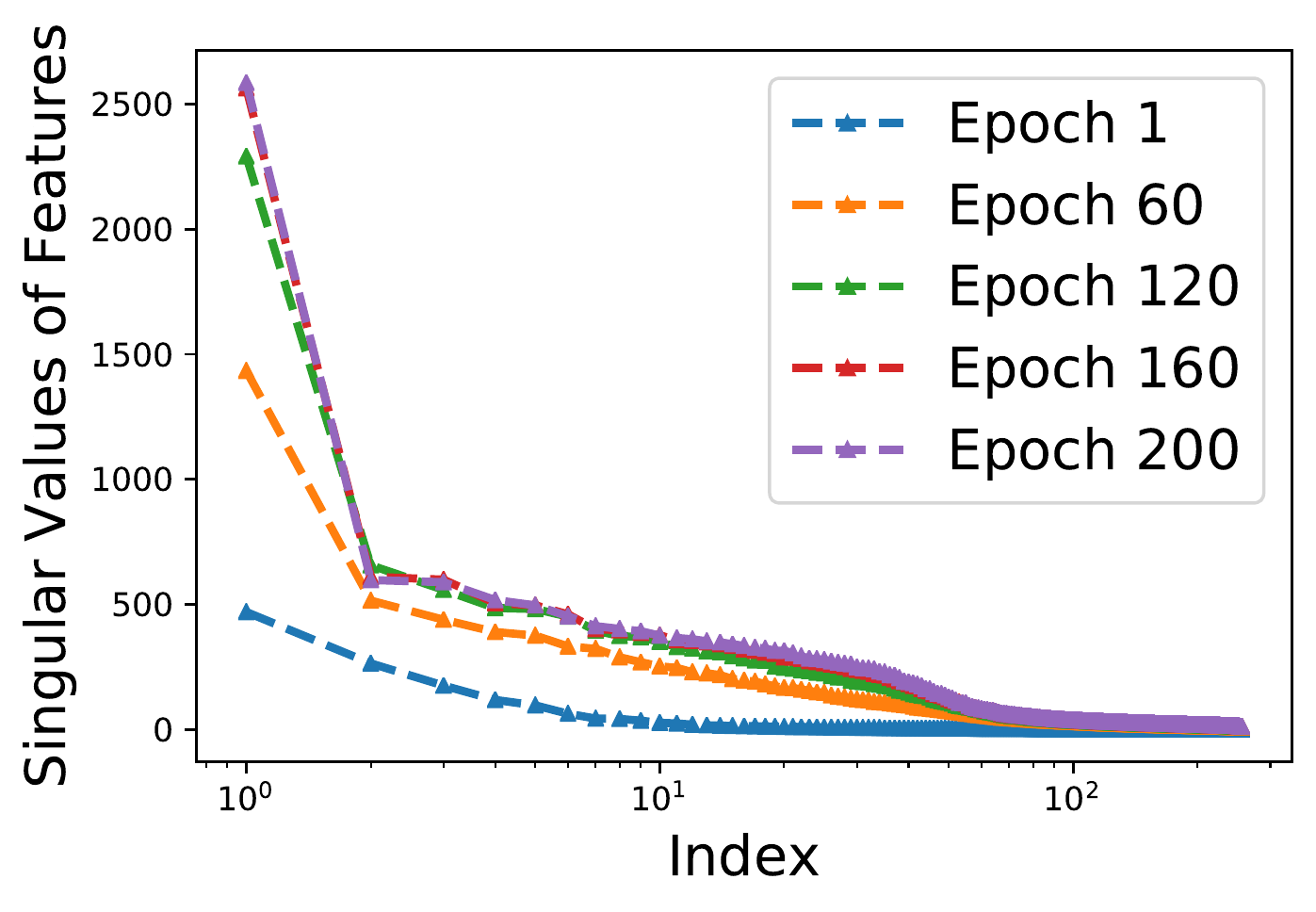}}\\
\subfloat[Ratios]{\includegraphics[width=0.45\textwidth]{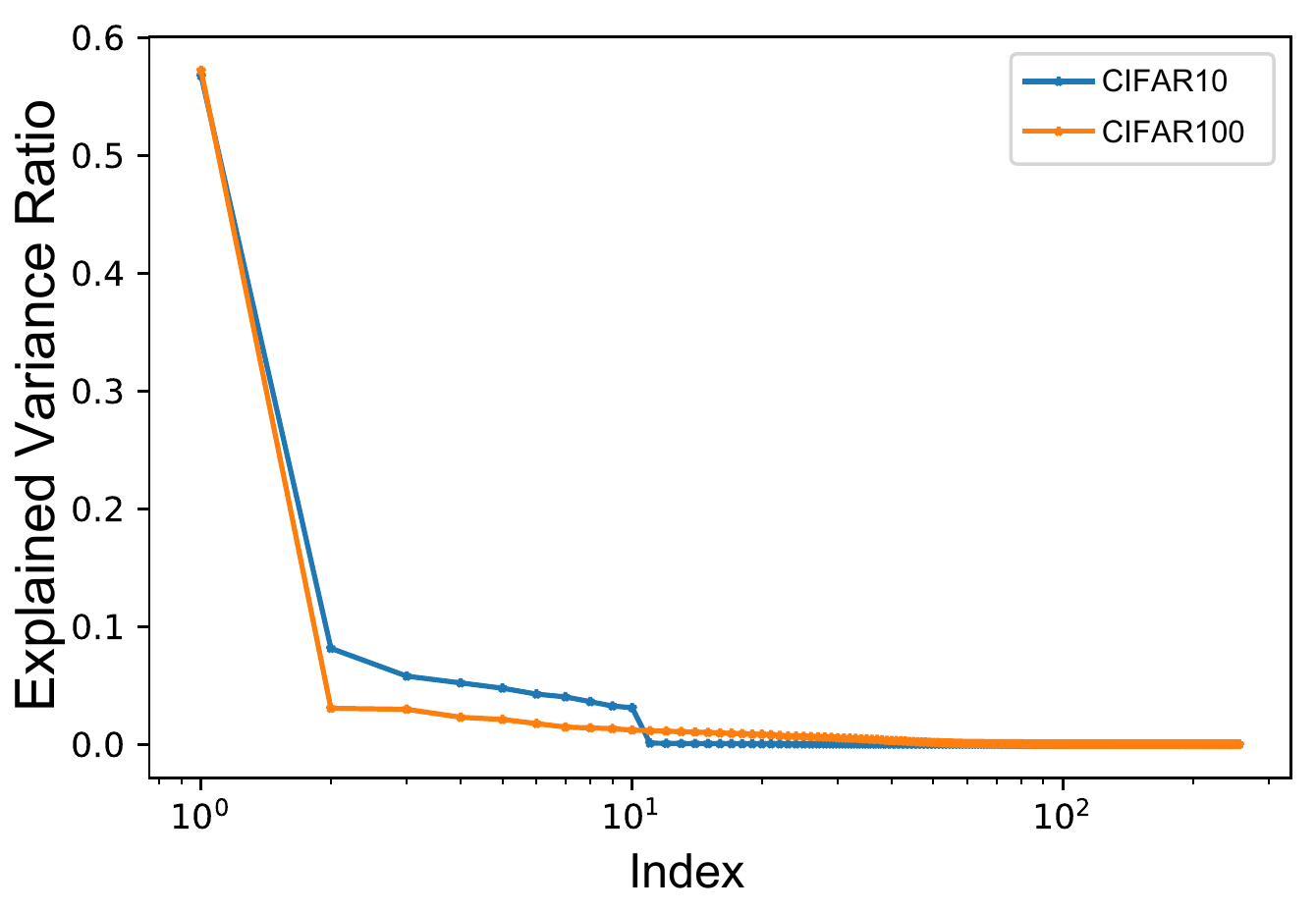}}
\caption{Singular Value Distribution and Explained Variance Ratios of the Matrix of Feature Vectors}
\label{fig:svd}
\end{figure}

\subsection{Statistical Properties of $\mathcal{P}$-vectors}
A possible threat to validity of using $\mathcal{P}$-vectors to analyze the feature space of DNN is that $\mathcal{P}$-vectors might fail to capture the necessary information to represent the features learned. To validate the relevance of $\mathcal{P}$-vectors, we analyze the statistical properties of $\mathcal{P}$-vectors as follows.

\paragraph{\bf\em Significance of $\mathcal{P}$-vectors.} We carried out Singular Value Decomposition on the matrix of feature vectors, using CIFAR-10 and CIFAR-100 datasets both based on ResNet-50 models, and obtain the distribution of singular values over indices. More specifically, we compute the distributions of singular values for the feature matrices obtained in the 1$^{st}$, 60$^{th}$, 120$^{th}$, 160$^{th}$ and 200$^{th}$ epochs to monitor the change of singular value distributions throughout the training procedure. It has been observed in Figures~\ref{fig:svd}(a)~and~(b) that a ``cliff'' pattern in the distribution becomes more and more significant after epochs of training for both CIFAR-10 and CIFAR-100 datasets -- a very small number (less than 10) of top singular values might dominate the whole distribution. In Figure~\ref{fig:svd}(c), we plot the curve of explained variance ratio $\sigma_k^2/{\sum_{j=1}^d\sigma_j^2}$ for every pair of singular vectors, using well-trained models of 200 epochs based on CIFAR-10 and CIFAR-100, where $\sigma_k$ refers to the $k^{th}$ singular value and $d$ is the rank of matrix. The explained variance ratio of the top-1 singular vectors (i.e., the $\mathcal{P}$-vector and the top-1 right singular vector) is more than 50\% while the second top singular vectors are less than 10\%. Results show the use of $\mathcal{P}$-vectors could represent the principal subspace of deep features learned. In addition to top-1 singular vectors (the $\mathcal{P}$-vector), Appendix (A.7) also presents the results of using top-$2,3,\dots$ singular vectors, no consistent observations have been obtained (not representative). 


    
\paragraph{\bf\em Distribution of Values in a $\mathcal{P}$-vector.} A potential threat to validity in this work is that \emph{when $\mathcal{P}$-vectors were trivial with identical values in every dimension, the angles between these vectors would close to zero}. To make our analysis more rigorous, we take a closer look at the $\mathcal{P}$-vector. In A.6 of Appendix, we include a study to observe the distribution of values in every dimension of the  $\mathcal{P}$-vector. With $\mathcal{P}$-vectors extracted from ResNet50, We first retrieve the value on every dimension of the vector and count the frequency that each value appears in the vector. In Figures~\ref{fig:Frequency} (a)--(e) and (f)--(j), we plot the histograms characterizing the frequency of values appeared in every dimension of the P-vector with smoothed curves  for CIFAR-10 and CIFAR-100 training datasets respectively, where results of models trained after the 0$^{th}$ (prior to the training procedure), 60$^{th}$, 120$^{th}$, 160$^{th}$, and 200$^{th}$ epoch have been presented. It is obvious that the values in every dimension of the $\mathcal{P}$-vectors are not identical. Though the shapes of histograms are similar, the frequency distributions are quite different from the ranges and values' perspectives. Thus, we can conclude that $\mathcal{P}$-vectors are different for different datasets. In this way, we can conclude that the $\mathcal{P}$-vectors for the same model in various training epochs are different; the $\mathcal{P}$-vectors for various datasets are different; and $\mathcal{P}$-vectors are not trivial while values are not identical.

\section{Uncovering Common Principal Subspace using Angles between $\mathcal{P}$-vectors}

\begin{figure}
\centering
\subfloat[MNIST (Train)]{\includegraphics[width=0.5\textwidth]{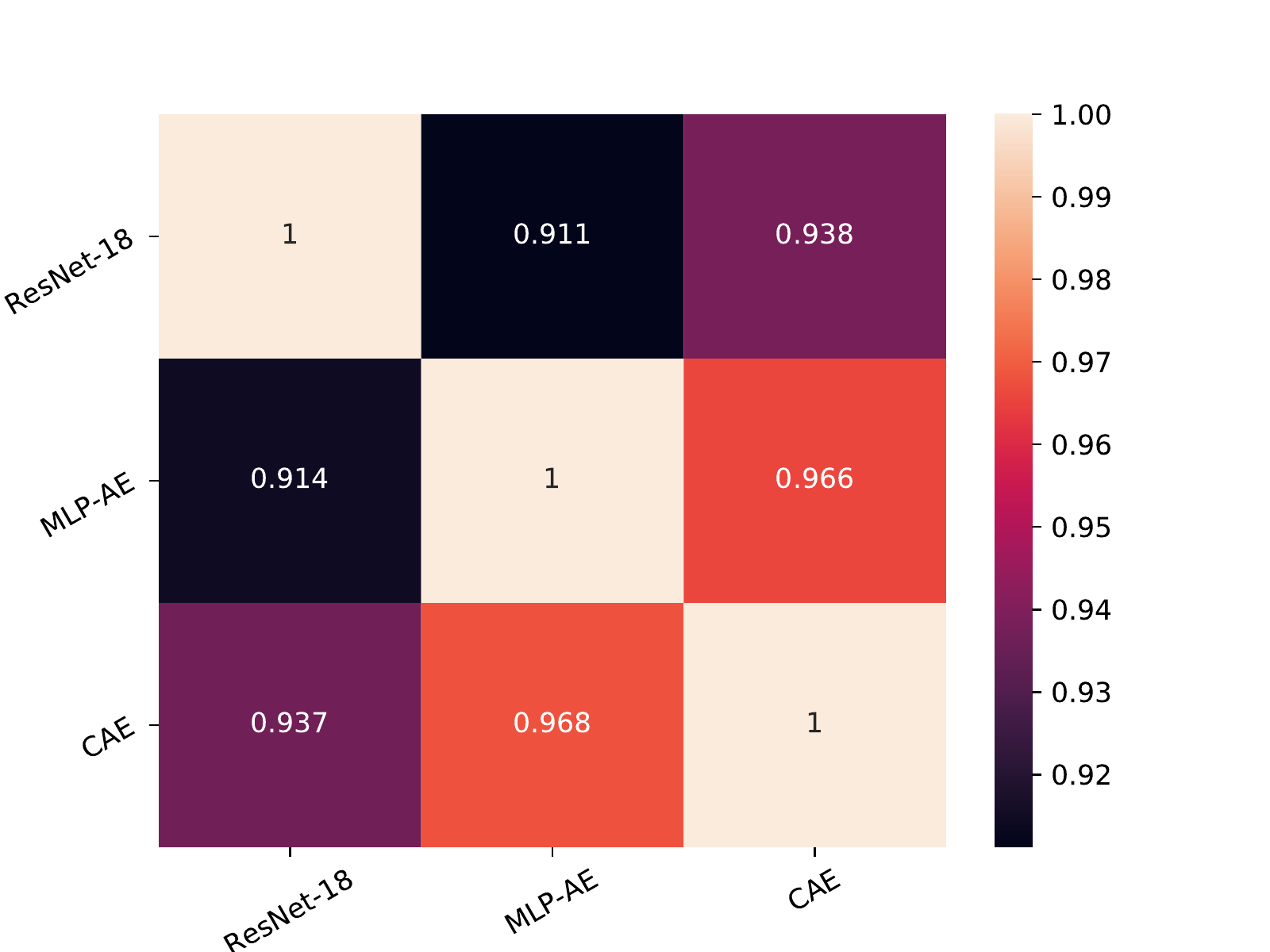}}
\subfloat[MNIST (Test)]{\includegraphics[width=0.5\textwidth]{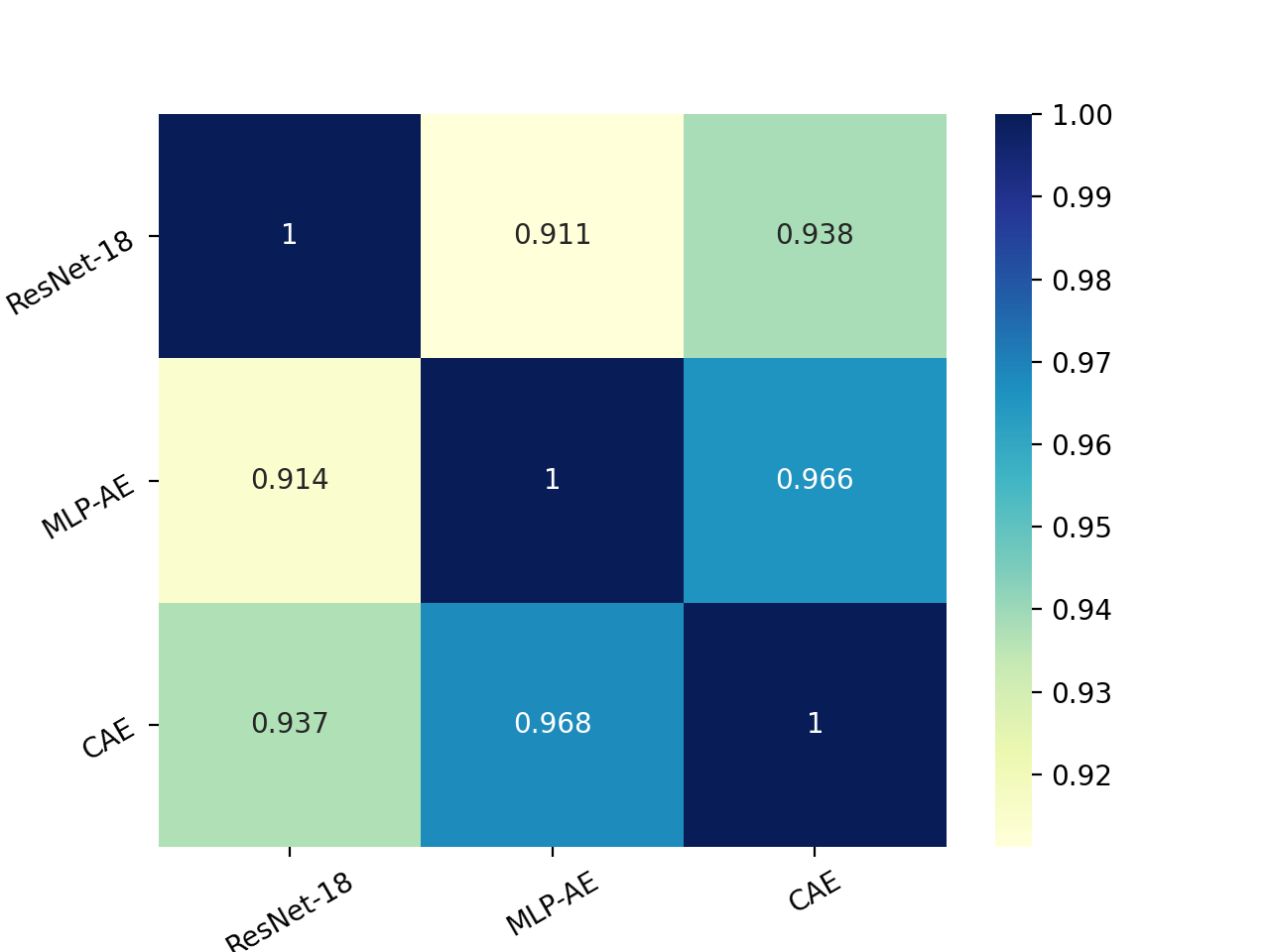}}\\
\subfloat[CIFAR-100 (Train)]{\includegraphics[width=0.5\textwidth]{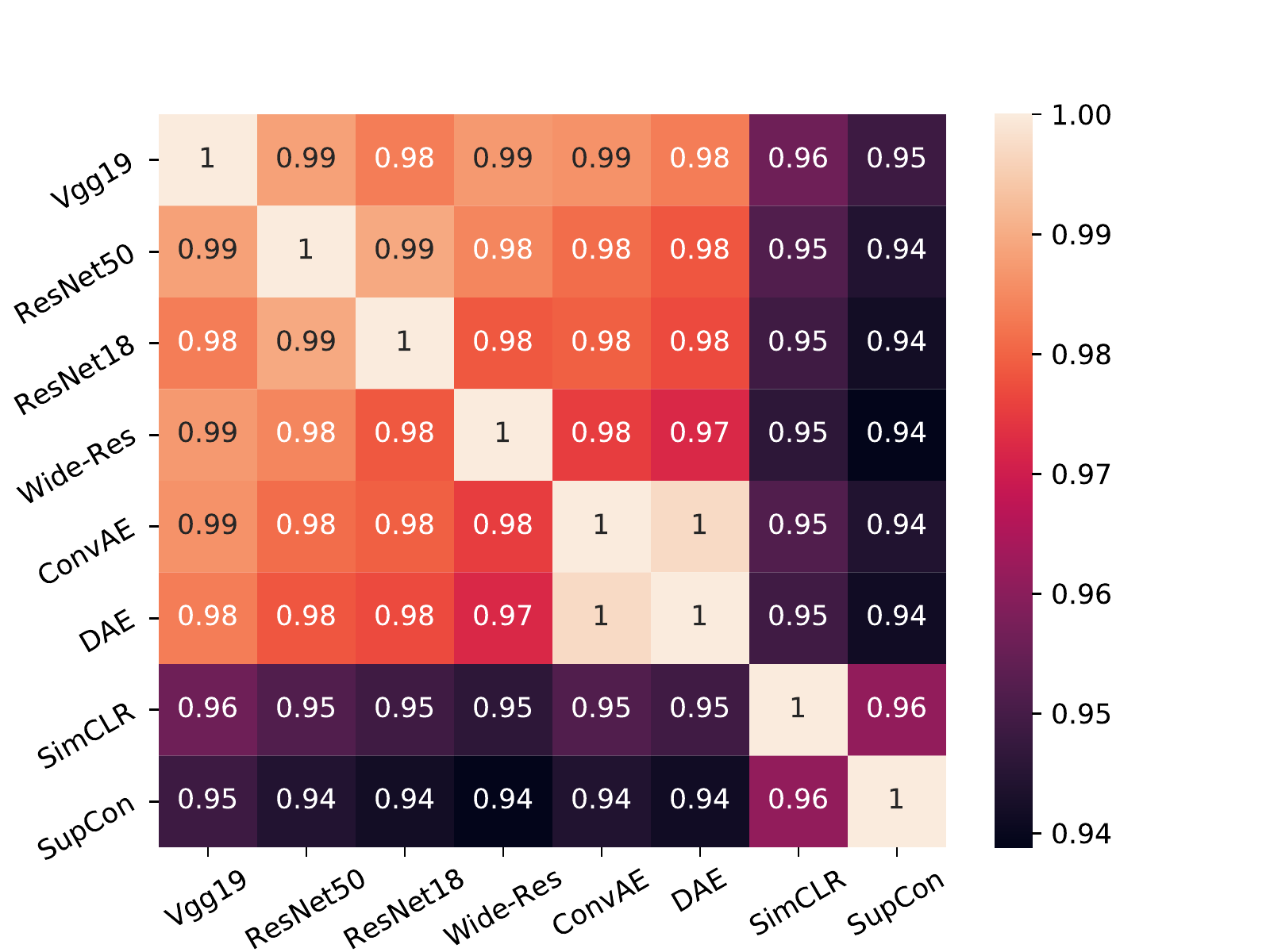}}
\subfloat[CIFAR-100 (Test)]{\includegraphics[width=0.5\textwidth]{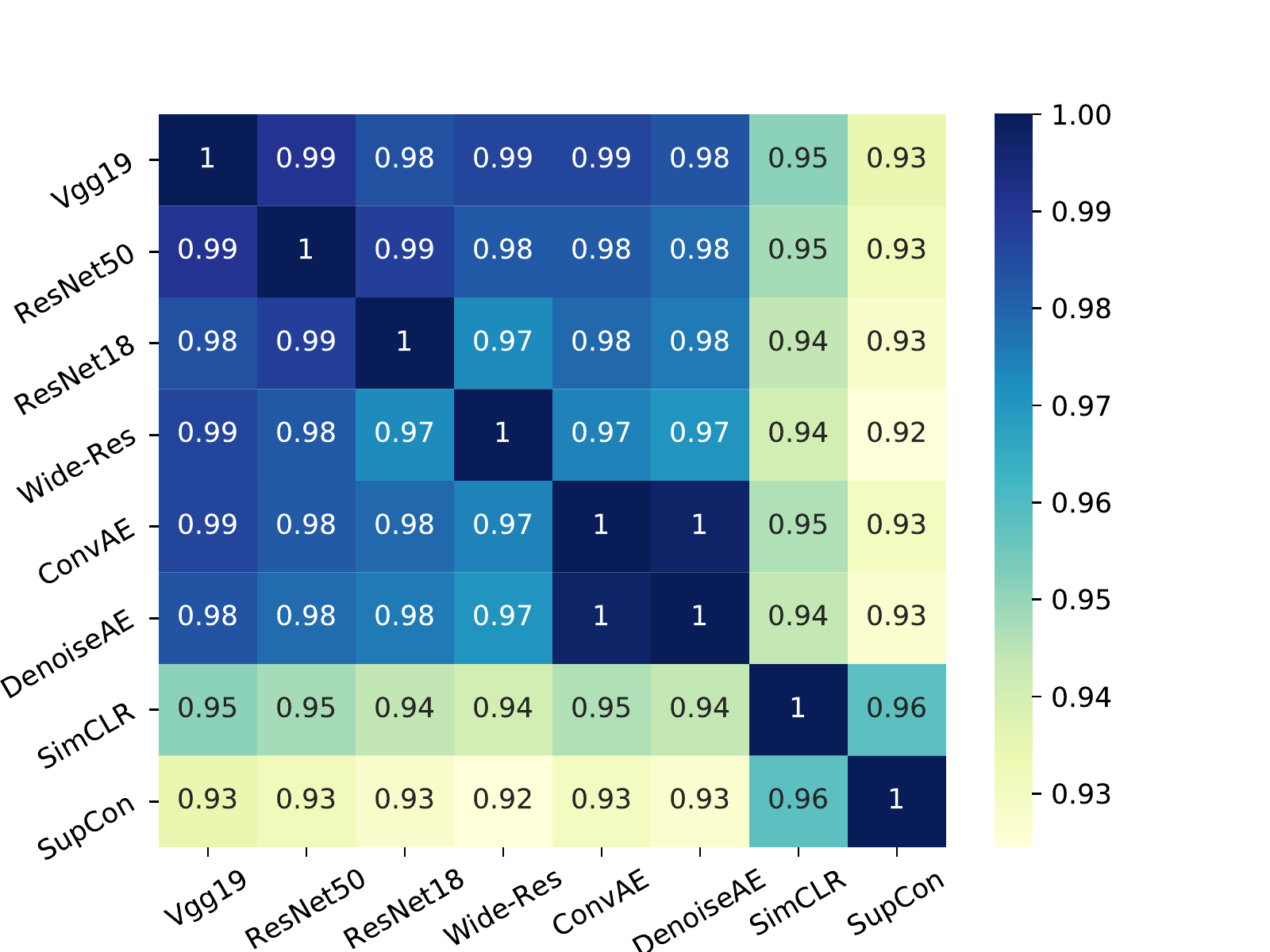}}
\caption{Cosine of angles between principal subspaces of deep features, measured using $\mathcal{P}$-vectors, for models trained under default settings.}
\label{fig:claim1}
\end{figure}

In this section, we elaborate the phenomena that we mentioned in Figures~\ref{fig:claim1-intro1} and~\ref{fig:claim1-intro2} of Section 1 using rich datasets, and provide details of evidences to support the hypotheses \textbf{H.I} and \textbf{H.II}.

\paragraph{\bf\em Common Principal Subspaces.} To demonstrate the existence of common subspaces, we propose to measure the angles between principal subspaces of deep features learned using the same datasets with various architectures under different tasks. Specifically, as was discussed in Section 3, we use the angle between $\mathcal{P}$-vectors as the proxy of measurements. 
Figure~\ref{fig:claim1-intro1}~(a)--(b) presents the cosine of angles between the $\mathcal{P}$-vectors of well-trained models (trained with various architectures and tasks using CIFAR-10 dataset). 
We carry out the same experiments on CIFAR-10 and CIFAR-100 datasets, and present the cosine of angles between the $\mathcal{P}$-vectors of well-trained models in Figure~\ref{fig:claim1}. All experiments based on both training and testing datasets demonstrate that, with a relatively large cosine (close to $1.0$), the models well-trained using the same dataset, no matter what types of architectures or whether labels has been used in the feature learning of various tasks, share a common principal subspace. In addition to above results, we also include the experiment results based on ImageNet~\cite{krizhevsky2012imagenet} dataset, which is sufficiently large and further confirm our observations. 


\paragraph{\bf\em Converging Trends.} To understand the dynamics of feature learning that shapes the common principal subspace from scratch, we measure the change of angles over epochs between the $\mathcal{P}$-vectors of training models (i.e., checkpoint for every epoch) and well-trained ones in comparisons. In Figure~\ref{fig:claim1-intro2}~(a)--(f), for every model of various tasks, we present angles between the $\mathcal{P}$-vectors of the well-trained model and itself's training checkpoint per epoch. A consistent convergence could be observed. As the $\mathcal{P}$-vectors of well-trained models are close to each other (see in Figures~\ref{fig:claim1-intro2} and~\ref{fig:claim1}), we can conclude the feature vectors extracted by these models would evolve on time and gradually converge to share the common subspace during the learning procedure. 

\begin{figure}
\centering
\subfloat[\footnotesize Sup. vs. Sup.]{\includegraphics[width=0.33\textwidth]{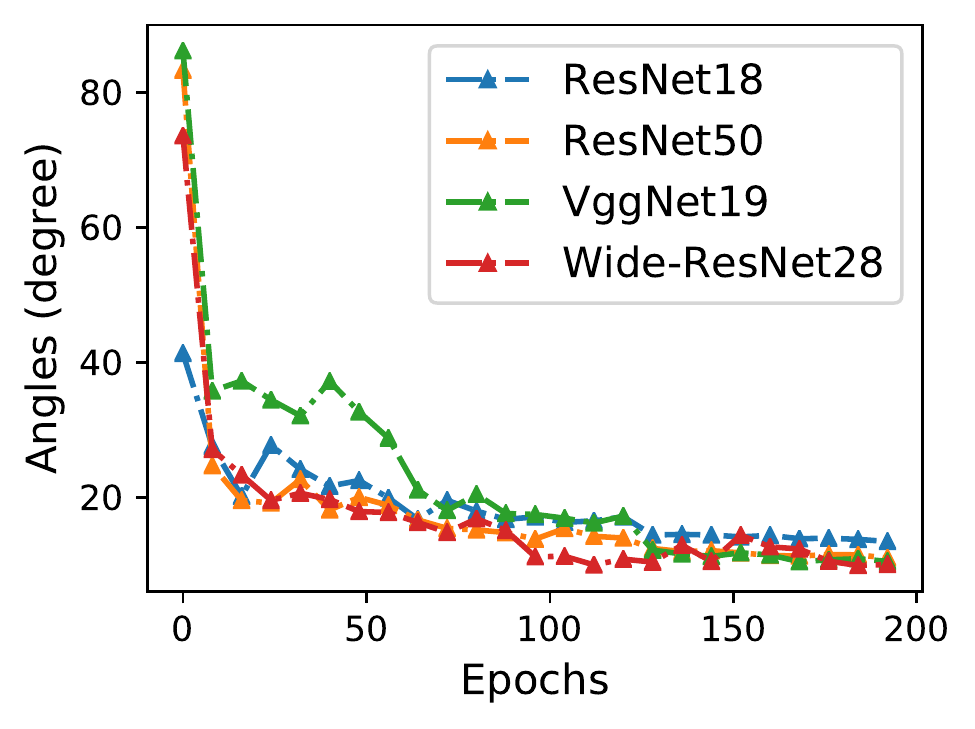}}
\subfloat[\footnotesize Sup. vs. Unsup.]{\includegraphics[width=0.33\textwidth]{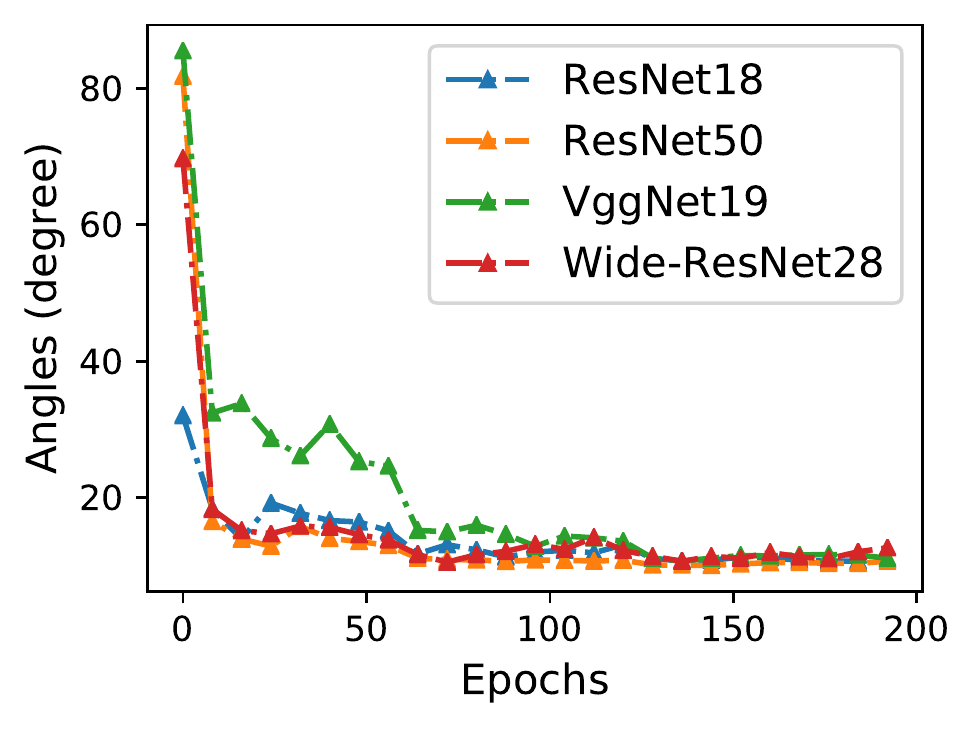}}
\subfloat[\footnotesize Sup. vs. Self-Sup.]{\includegraphics[width=0.33\textwidth]{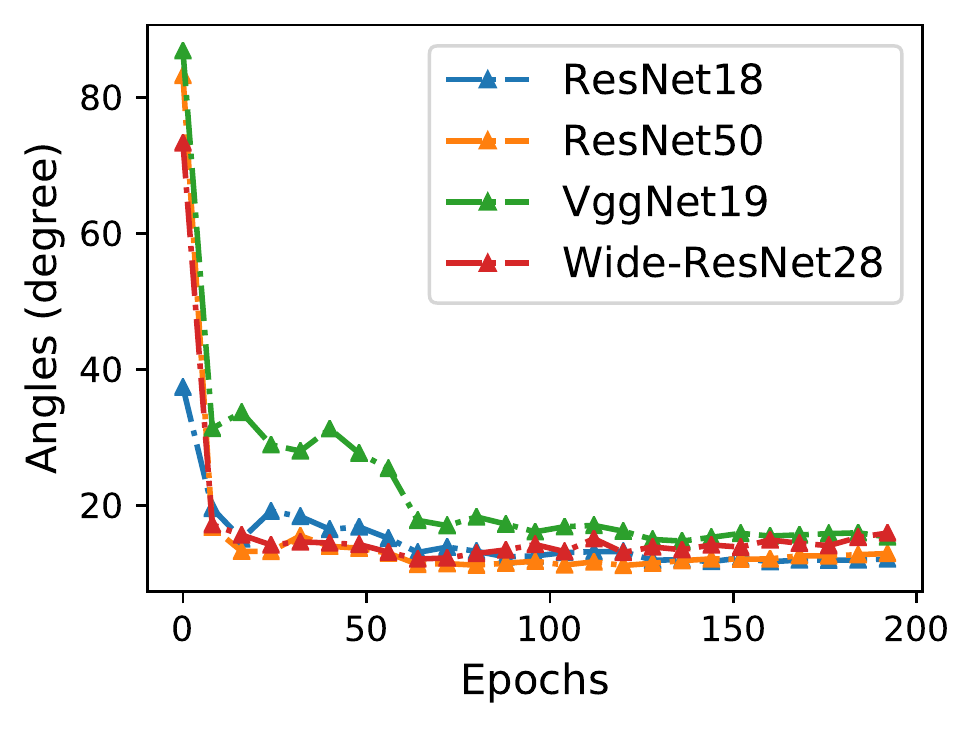}}\\
\subfloat[\footnotesize Unsup. vs. Sup.]{\includegraphics[width=0.33\textwidth]{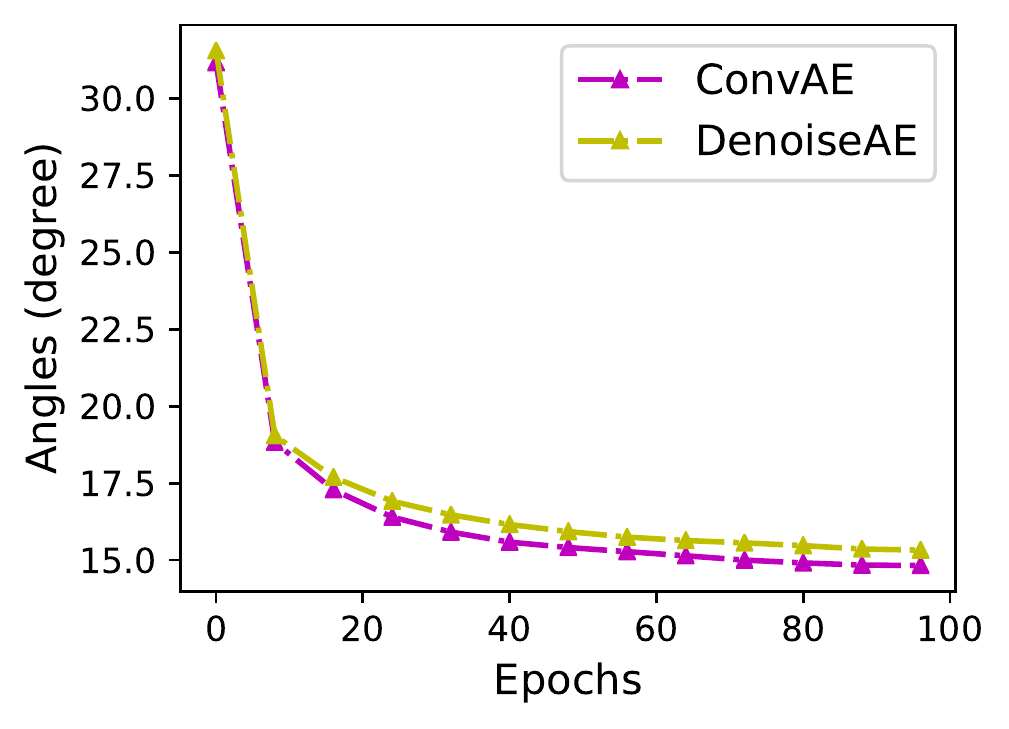}}
\subfloat[\scriptsize Unsup. vs. Unsup.]{\includegraphics[width=0.33\textwidth]{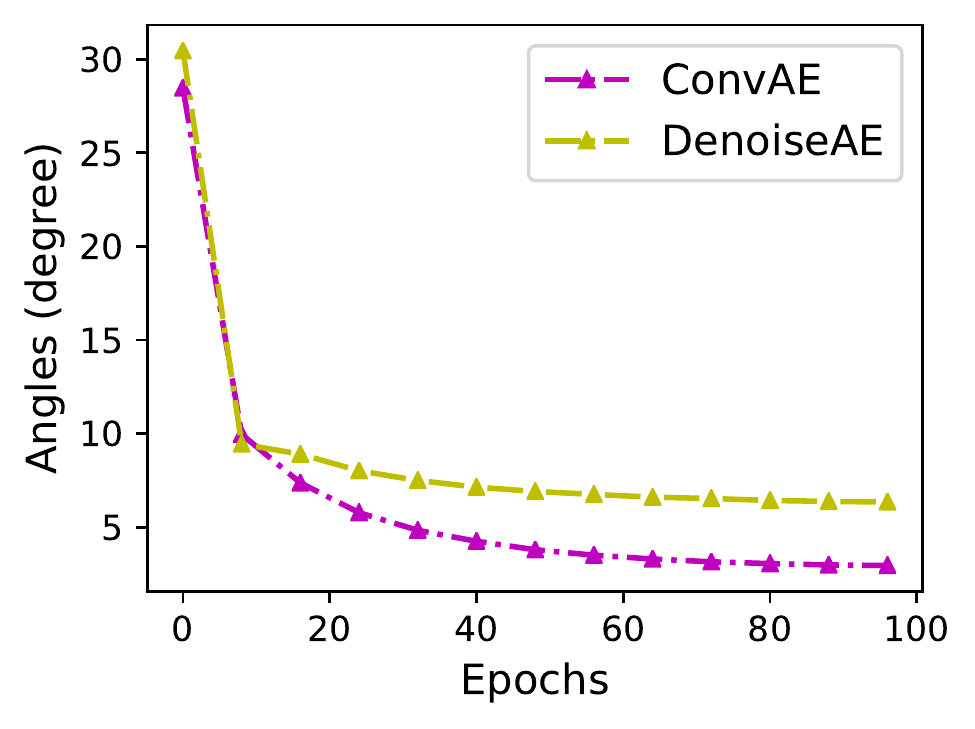}}
\subfloat[\footnotesize Unsup vs Self-Sup]{\includegraphics[width=0.33\textwidth]{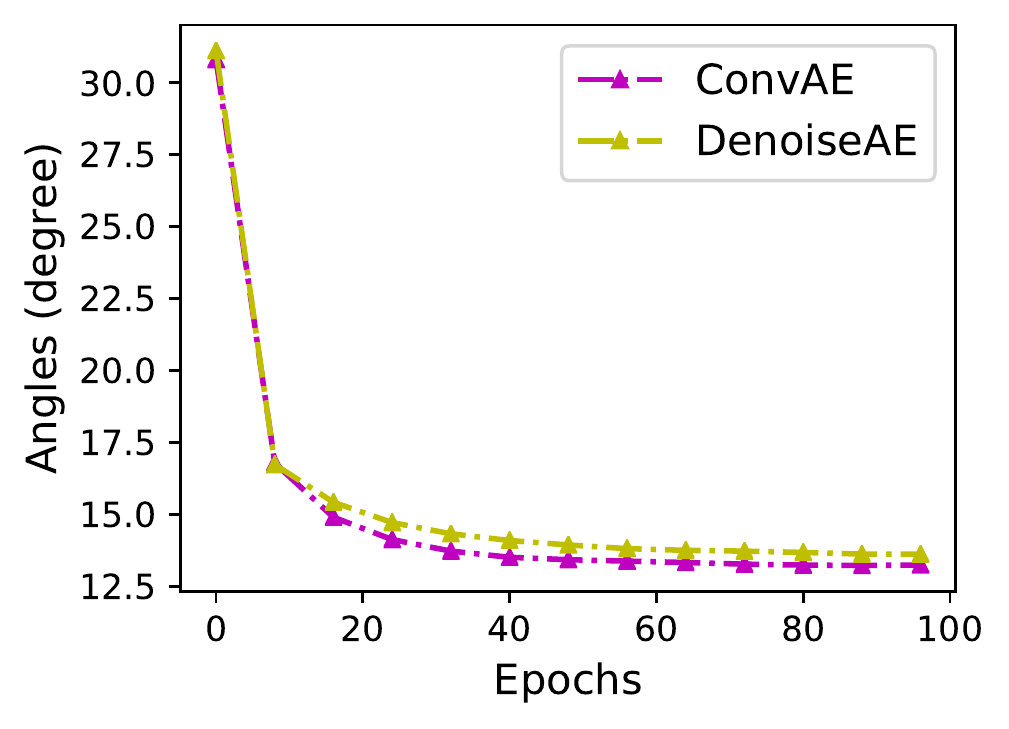}}\\
\subfloat[\footnotesize Self. vs Sup.]{\includegraphics[width=0.33\textwidth]{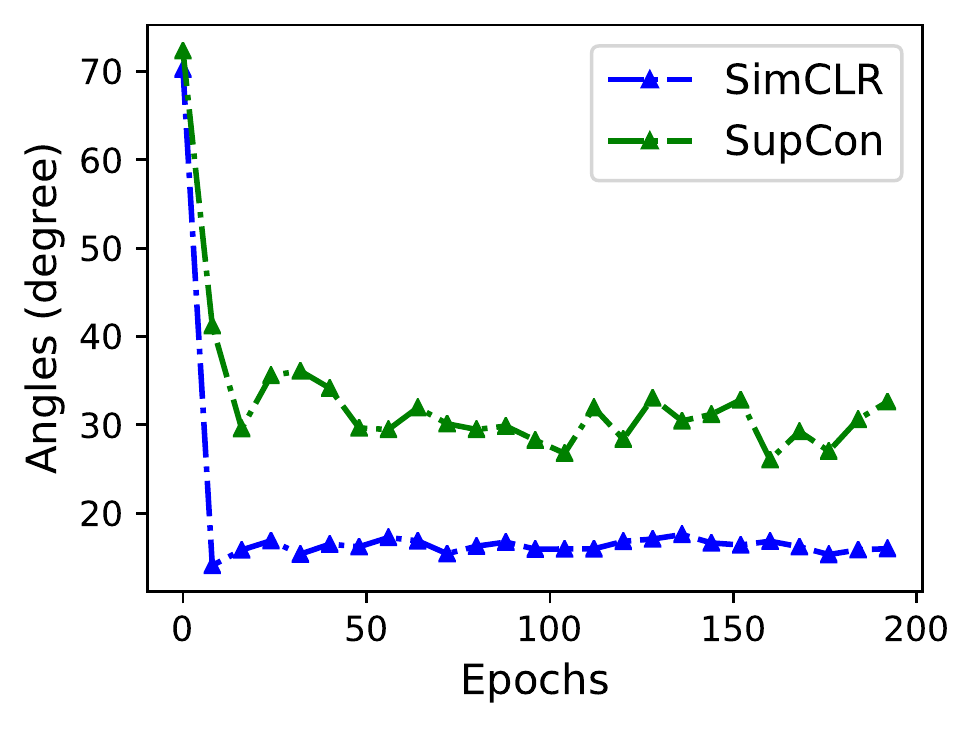}}
\subfloat[\footnotesize Self. vs Unsup.]{\includegraphics[width=0.33\textwidth]{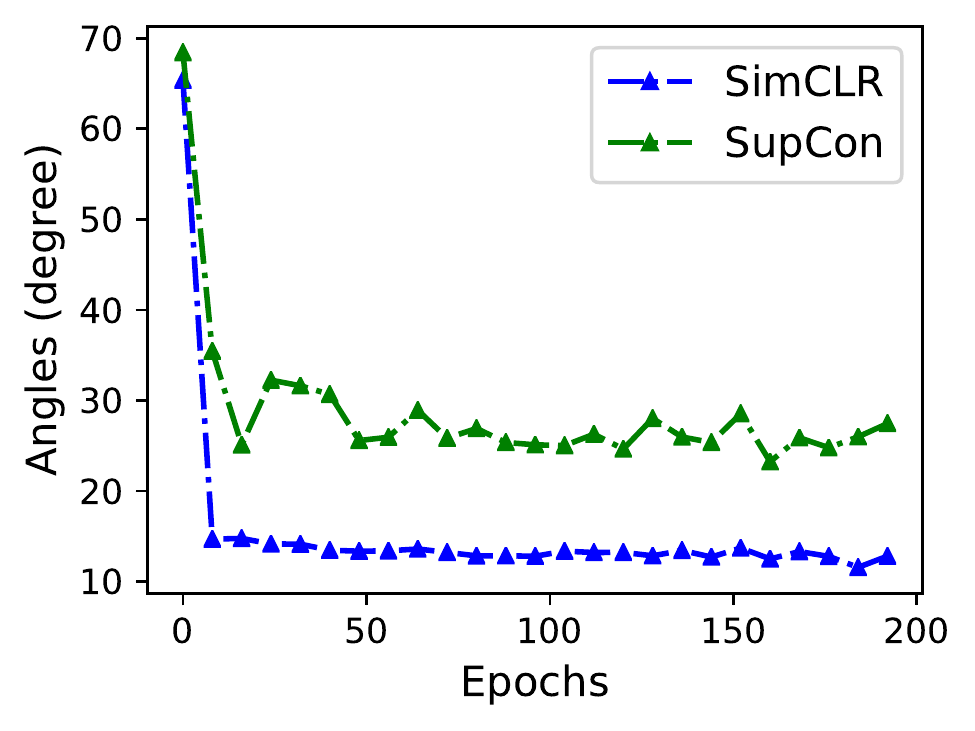}}
\subfloat[\footnotesize Self. vs Self-Sup.]{\includegraphics[width=0.33\textwidth]{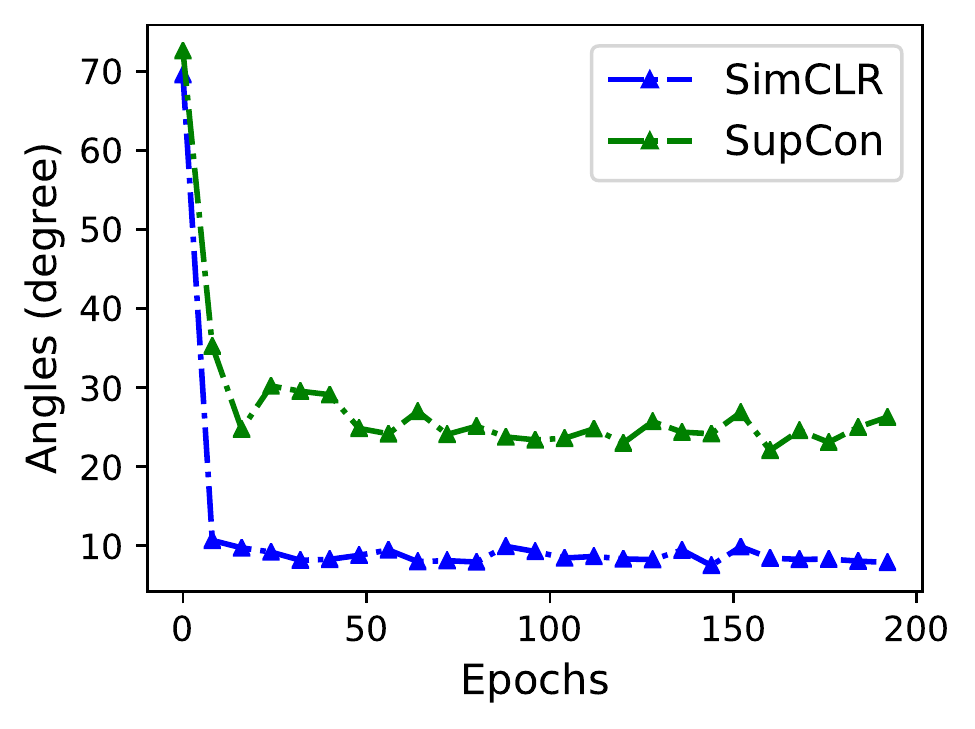}}
\caption{{Angles between the principal subspaces of the well-trained model and the checkpoint per training epoch using $\mathcal{P}$-vectors based on CIFAR-10.} }
\label{fig:claim2}
\end{figure}

We carry out a more comprehensive model-to-model comparison using CIFAR-10 dataset. Figures.~\ref{fig:claim2}~(a)--(c) present the converging trends of $\mathcal{P}$-vector angles between the well-trained supervised models and the checkpoints per epoch of supervised, unsupervised, and self-supervised learning models, where we use the well-trained Wide-ResNet28 (trained with 200 epochs under suggest settings) as the reference of supervised models. Figures.~\ref{fig:claim2}~(d)--(f) present $\mathcal{P}$-vector angles between the well-trained ConvAE (trained with 100 epochs under suggest settings as the reference of unsupervised learning) and the checkpoints per epoch of supervised, unsupervised, and self-supervised learning models. Figures.~\ref{fig:claim2}~(g)--(i) present the $\mathcal{P}$-vector angles between the well-trained SimCLR representations (trained under suggest settings~\citep{chen2020simple} as the reference of self-supervised learning) and the checkpoints per epoch for supervised classification, unsupervised image reconstruction, and self-supervised contrastive learning tasks. The converging trends in all comparison further validate our hypotheses. 

Note that, even though a model has been selected as the reference of well-trained models for the comparison in every setting, the angle between $\mathcal{P}$-vectors of the model to itself would not converge to zero in Figure~\ref{fig:claim2}. As was stated in the caption of Figure~\ref{fig:claim1-intro1}, we carried out experiments with 5 independent trails with different random seeds. Among these models, only one would be selected as the reference. In Figure.~\ref{fig:claim2}, we plot the average angles for the 5 trails, where the angles might not be able to converge to zero even when the target and the reference are with the same architectures. 

\begin{figure}
\centering
\subfloat[Supervised CIFAR-10]{\includegraphics[width=0.45\textwidth]{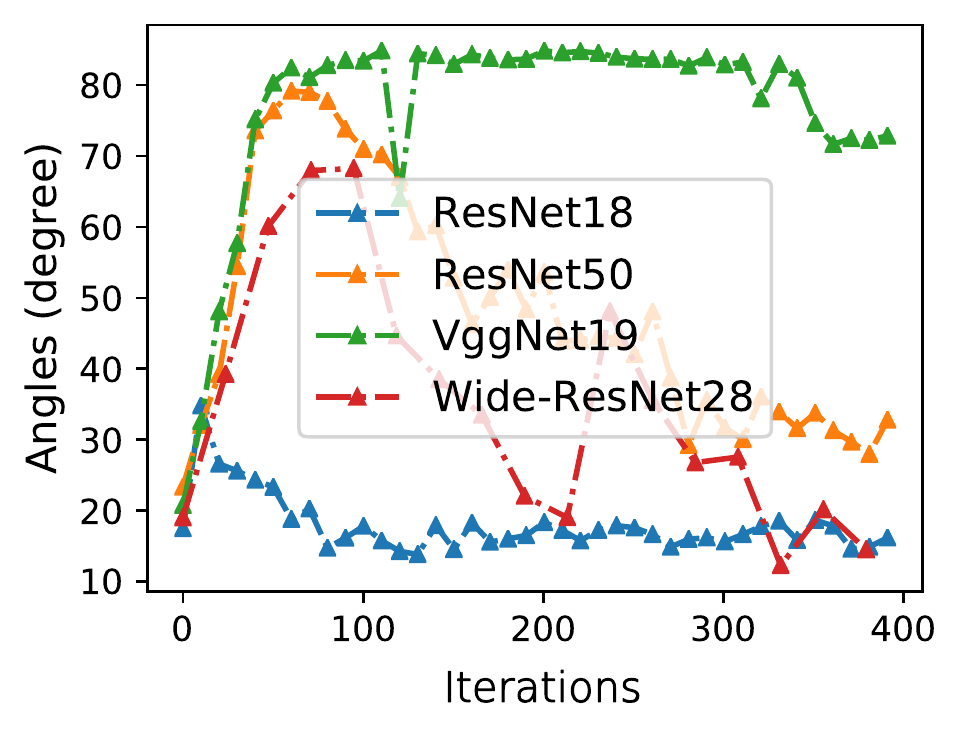}}
\subfloat[Supervised CIFAR-100]{\includegraphics[width=0.45\textwidth]{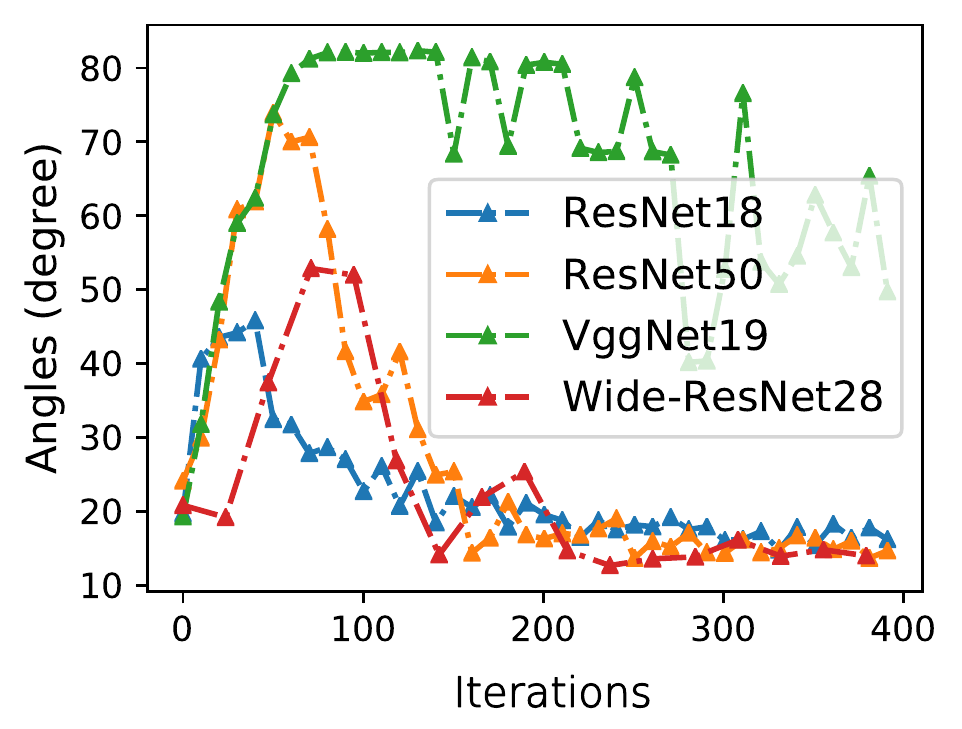}}\\
\subfloat[Unsupervised CIFAR-10]{\includegraphics[width=0.45\textwidth]{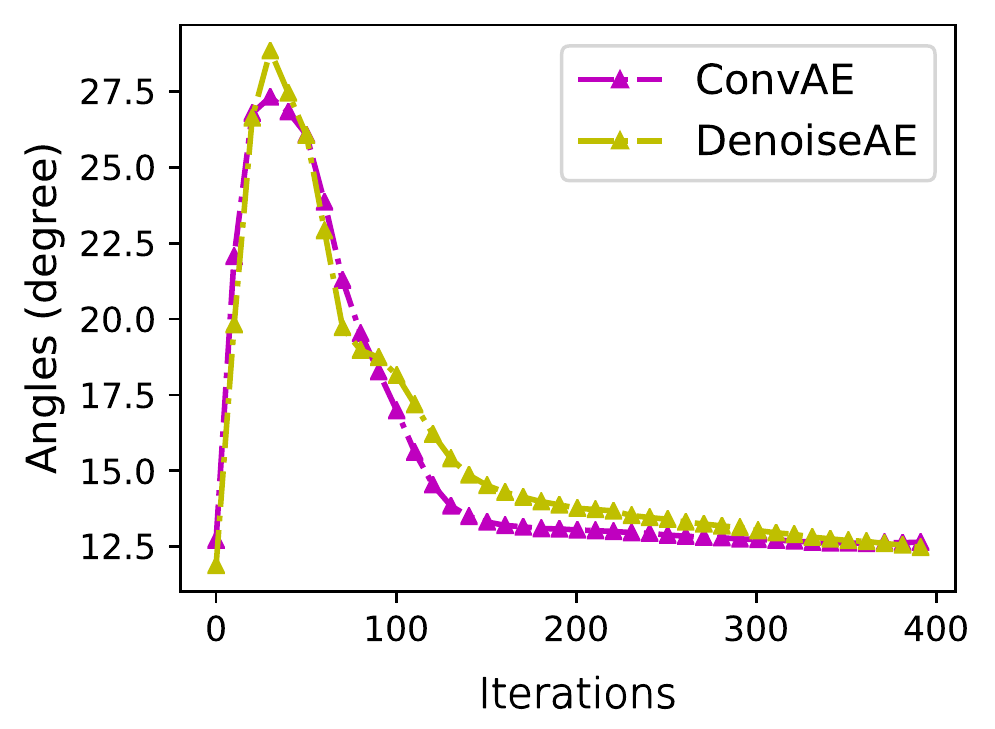}}
\subfloat[Unsupervised CIFAR-100]{\includegraphics[width=0.45\textwidth]{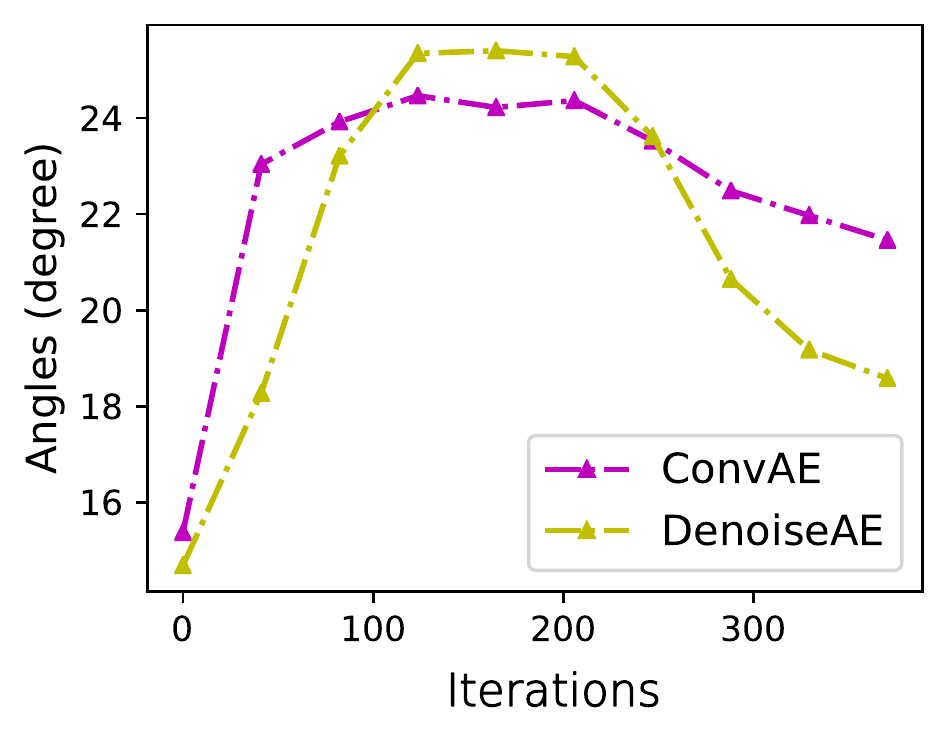}}\\
\subfloat[Self-Supervised CIFAR-10]{\includegraphics[width=0.45\textwidth]{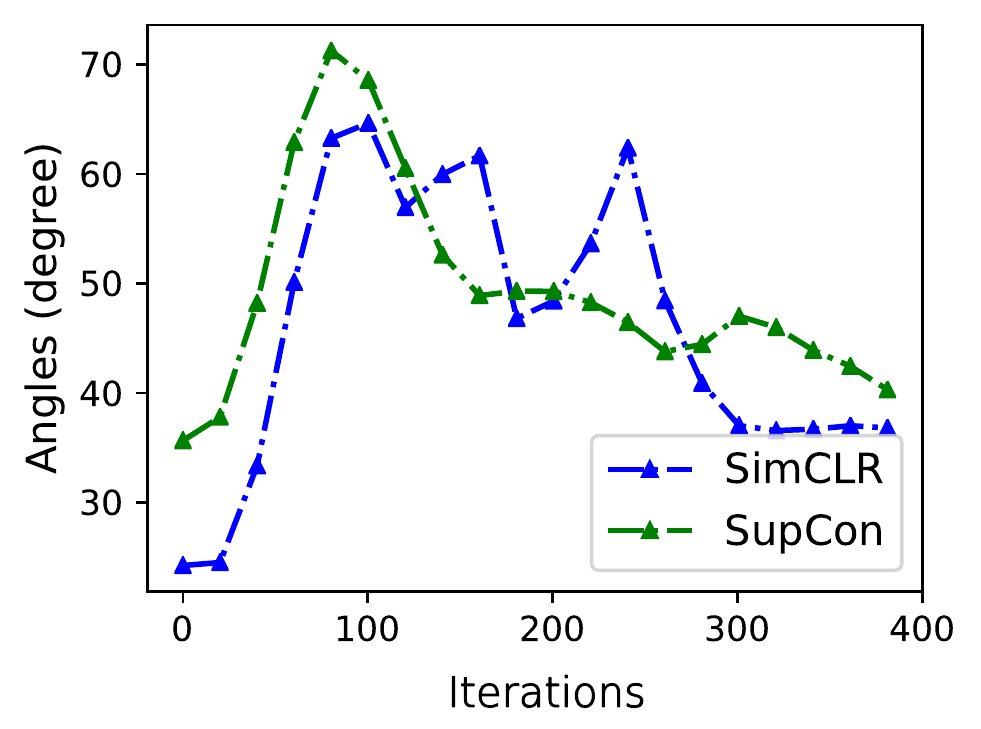}}
\subfloat[Self-Supervised CIFAR-100]{\includegraphics[width=0.45\textwidth]{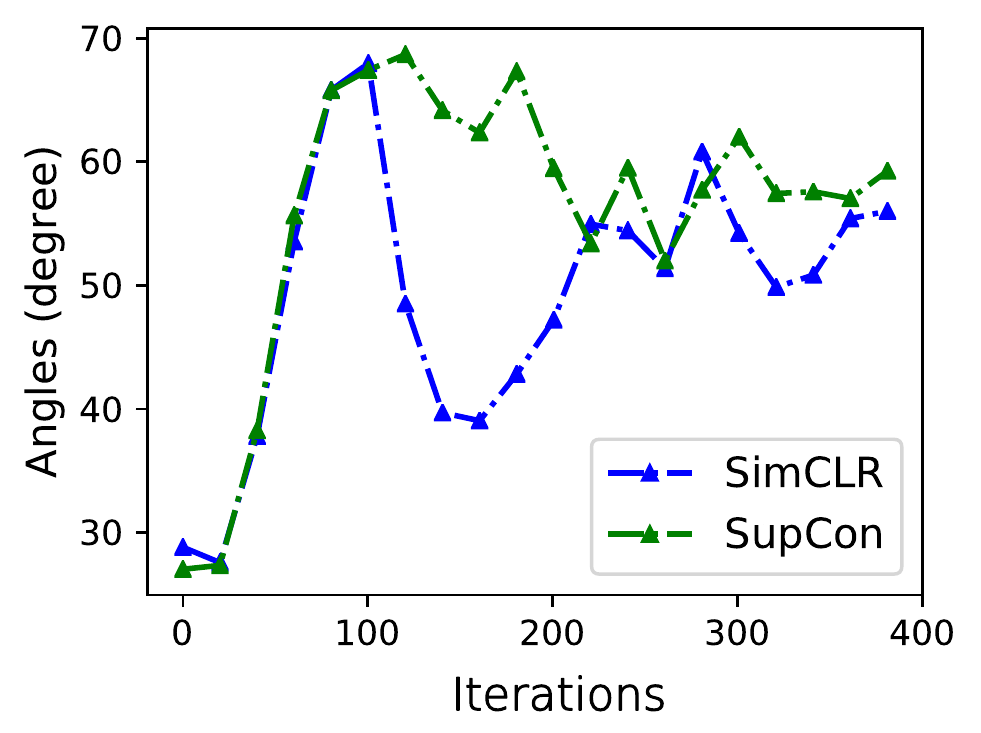}}
\caption{{Angles between principal subspaces, measured using $\mathcal{P}$-vectors based on CIFAR-10 and CIFAR-100, between well-trained models and checkpoints per training iteration in the first epoch.} }
\label{fig:epoch0}
\end{figure}

\paragraph{\bf\em Discussion.} To better visualize every comparison, we use the maximal angles achieved during the first epoch to represent the angles between $\mathcal{P}$-vectors corresponding to the first epoch in Figures~\ref{fig:claim1-intro2} and~\ref{fig:claim2}. Actually, in the first epoch, there would involve some non-monotonic trends for the angles varying over the number of iterations. Figure~\ref{fig:epoch0} presents the angles between the $\mathcal{P}$-vectors of the training and well-trained models over the number of iterations in the first epoch in three settings of learning, where we use the training set of CIFAR-10 and CIFAR-100 for the experiments (Please refer to the results based on the testing set of CIFAR-10 and CIFAR-100 in Appendix A.2). While such ``\emph{zigzag}'' curves could be found in the first few training epochs, the angles between the $\mathcal{P}$-vectors drop down and converge to smaller ones over the number of iterations in the rest of learning procedure.
%

Note that in Appendix (A.7), we include the comparison between the top singular vectors other than the $\mathcal{P}$-vectors, i.e., the angles of top-2, 3, 4, 5 and 6 left singular vectors between models, where we cannot observe the phenomena.


\section{Predicting Generalization Performance using $\mathcal{P}$-vectors}
In this section, we study the angles between the principal subspaces of (raw) data and deep features, and use such angles to predict the generalization performance of models (\textbf{H.III}).

\subsection{Measuring Angles between Principal Subspace of Raw Data and Deep Features with $\mathcal{P}$-vectors} 
Given the raw data matrix, i.e., a \#samples$\times$\#data dimensions matrix, we obtain the \underline{Data $\mathcal{P}$-vector}\footnote{We use the term ``model $\mathcal{P}$-vector'' to represent a $\mathcal{P}$-vector estimated using feature vectors of a deep model, while using ``data $\mathcal{P}$-vector'' as the top left singular vector of the raw data matrix.} of these samples using the top left singular vector of the raw data matrix, which represents the principal subspace of raw features (or the position of every sample projected by the principal component of the data). To understand the connection between models and data, We carry out case studies using CIFAR-10 dataset, including (1) measuring the angles between the principal subspace of deep features and the raw data and (2) understanding the changes of angles over training epochs, using model and data $\mathcal{P}$-vectors.

\begin{figure}
\centering
\subfloat[Supervised CIFAR-10]{\includegraphics[width=0.5\textwidth]{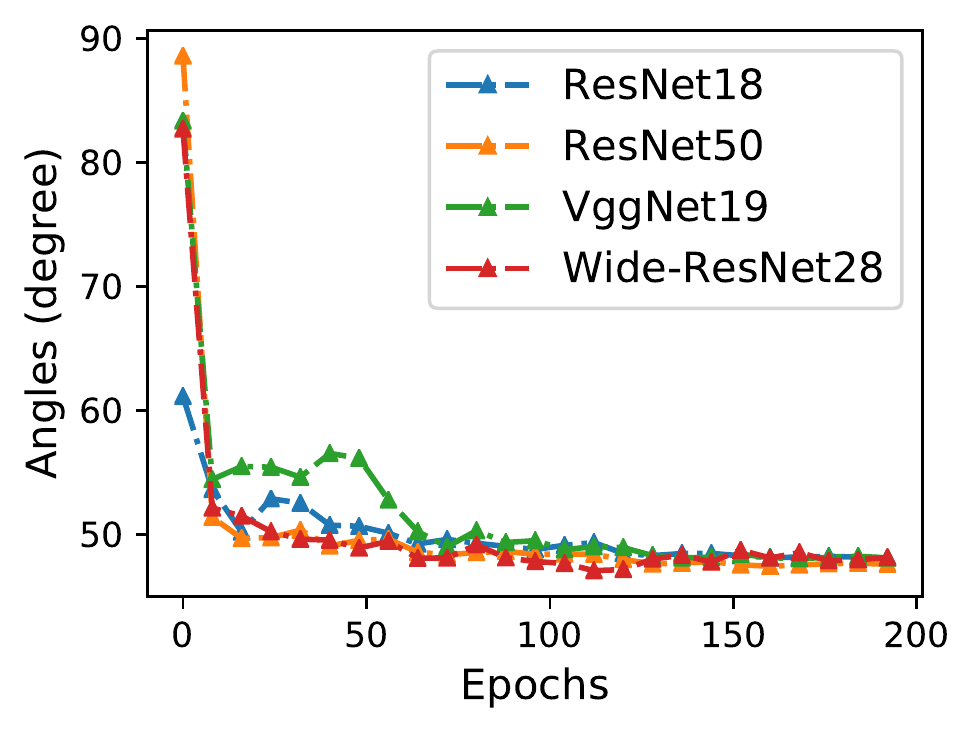}}
\subfloat[Supervised CIFAR-100]{\includegraphics[width=0.5\textwidth]{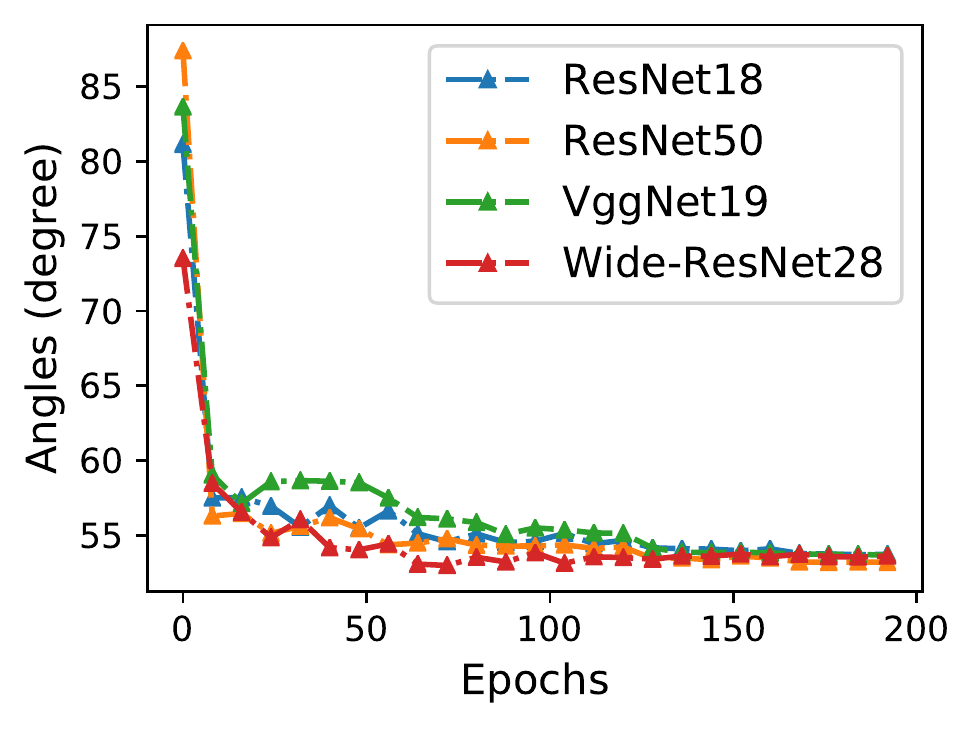}}\\
\subfloat[Unsupervised CIFAR-10]{\includegraphics[width=0.5\textwidth]{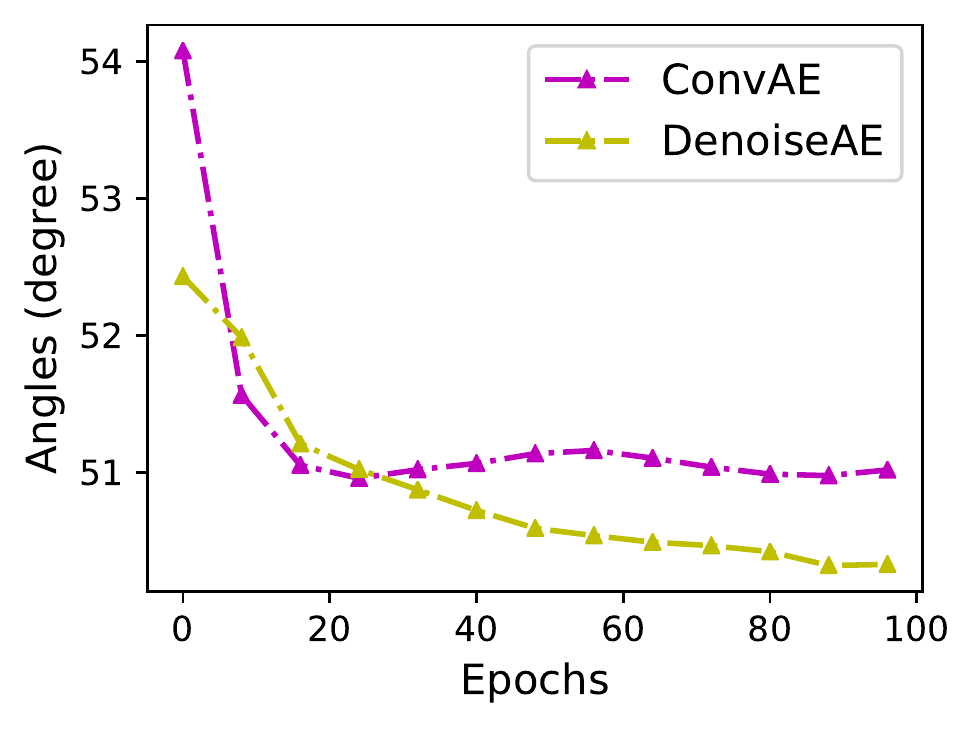}}
\subfloat[Unsupervised CIFAR-100]{\includegraphics[width=0.5\textwidth]{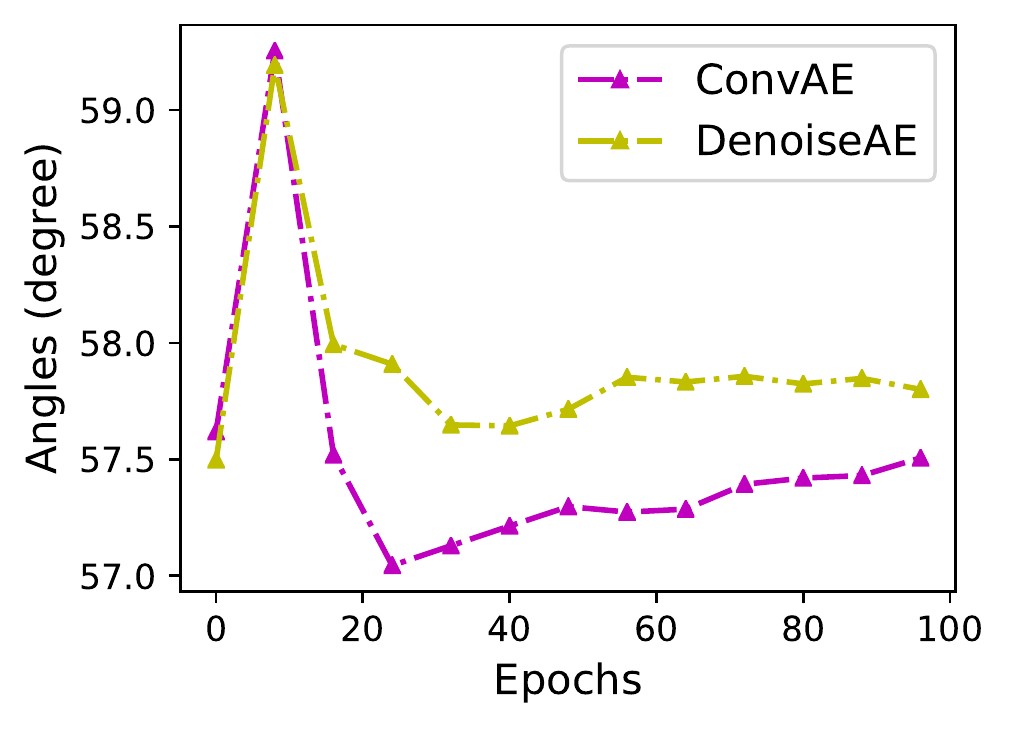}}\\
\subfloat[Self-Supervised CIFAR-10]{\includegraphics[width=0.5\textwidth]{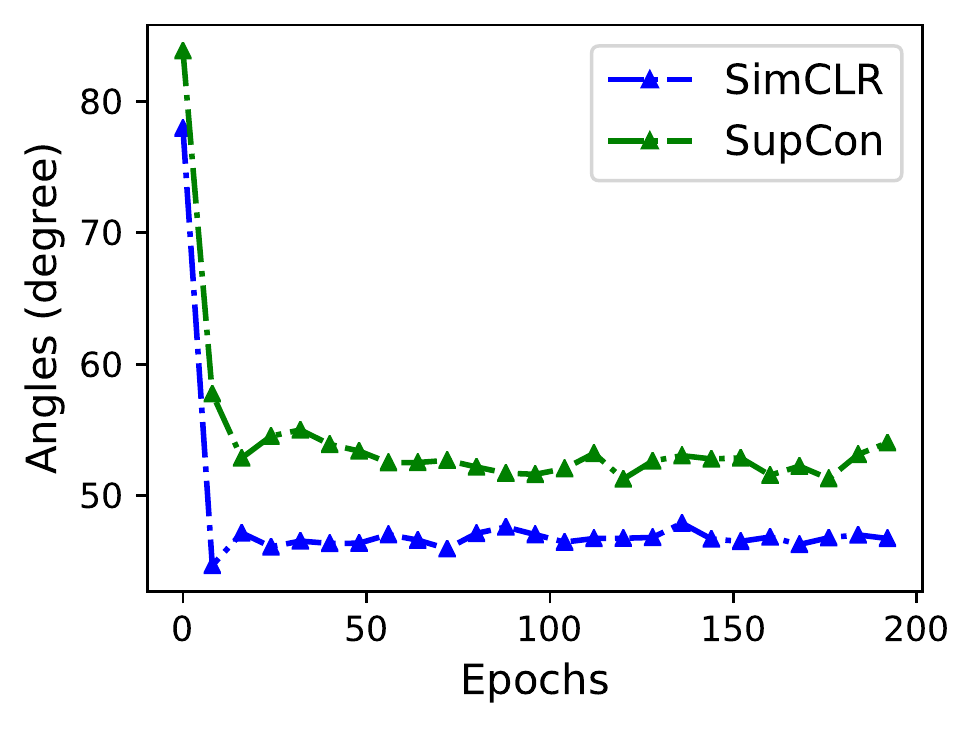}}
\subfloat[Self-Supervised CIFAR-100]{\includegraphics[width=0.5\textwidth]{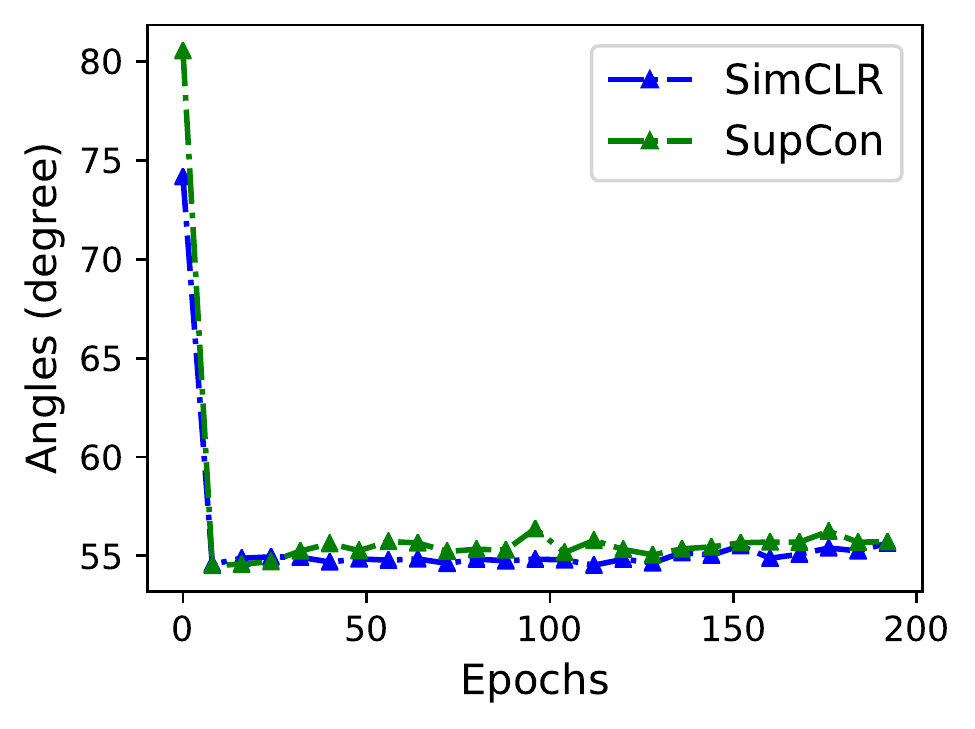}}
\caption{{Angles between principal subspaces of raw data and the checkpoint per training epoch, measured using model and data $\mathcal{P}$-vectors with CIFAR-10 (C10) and CIFAR-100 (C100).} }
\label{fig:claim3-1}
\end{figure}

Figure.~\ref{fig:claim3-1} shows the converging trend of angles between data and model $\mathcal{P}$-vectors over number of training epochs for supervised, unsupervised, and self-supervised learning models. The angles between the data and model $\mathcal{P}$-vectors start from about orthogonal and generally decrease and converge to about 50$^\circ$ degree to 60$^\circ$ degree. Note that the angle points corresponding to the first epoch on the curves are the largest $\mathcal{P}$-vector angles during the training procedure of the first epoch from the random scratch. The results indicate that the principal subspace of the well-trained models are more close to the principal subspace of the raw data, no matter which architectures are used for what kind of learning tasks or whether labels are used for training.


\begin{figure*}
\centering
\subfloat[VggNet16 (Train)]{\includegraphics[width=0.5\textwidth]{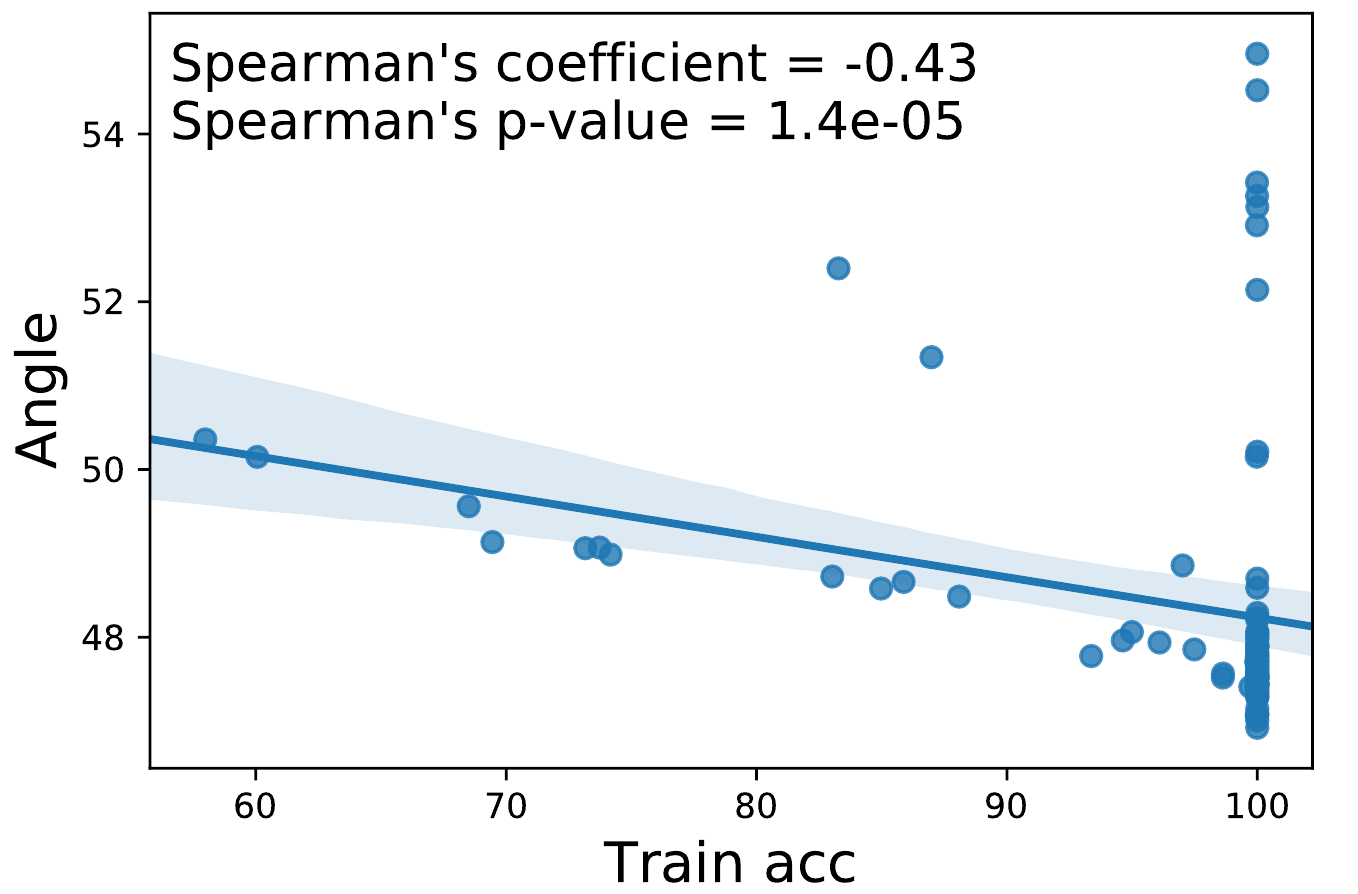}}
\subfloat[VggNet16 (Test)]{\includegraphics[width=0.5\textwidth]{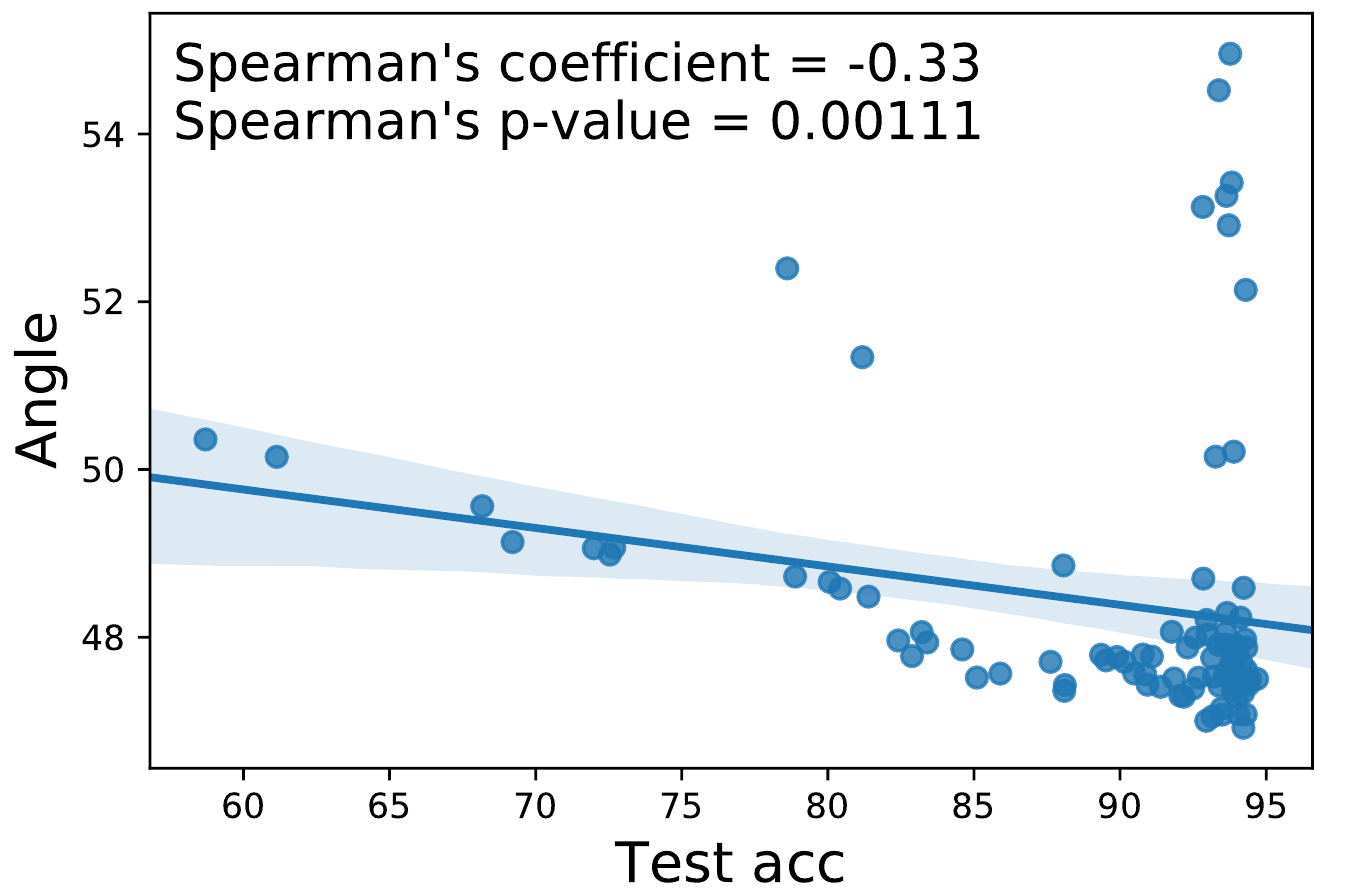}}\\
\subfloat[ResNet20 (Train)]{\includegraphics[width=0.5\textwidth]{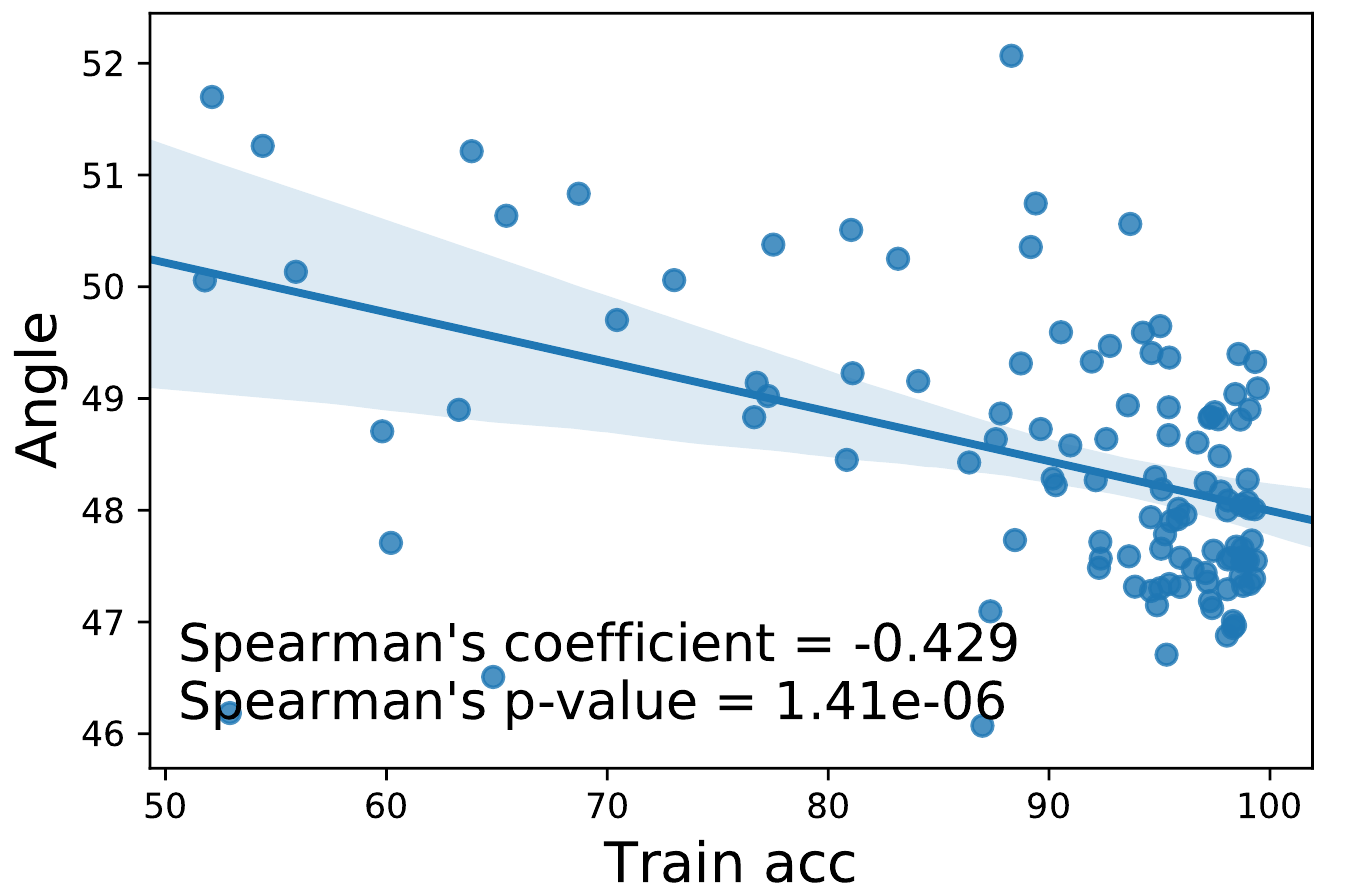}}
\subfloat[ResNet20 (Test)]{\includegraphics[width=0.5\textwidth]{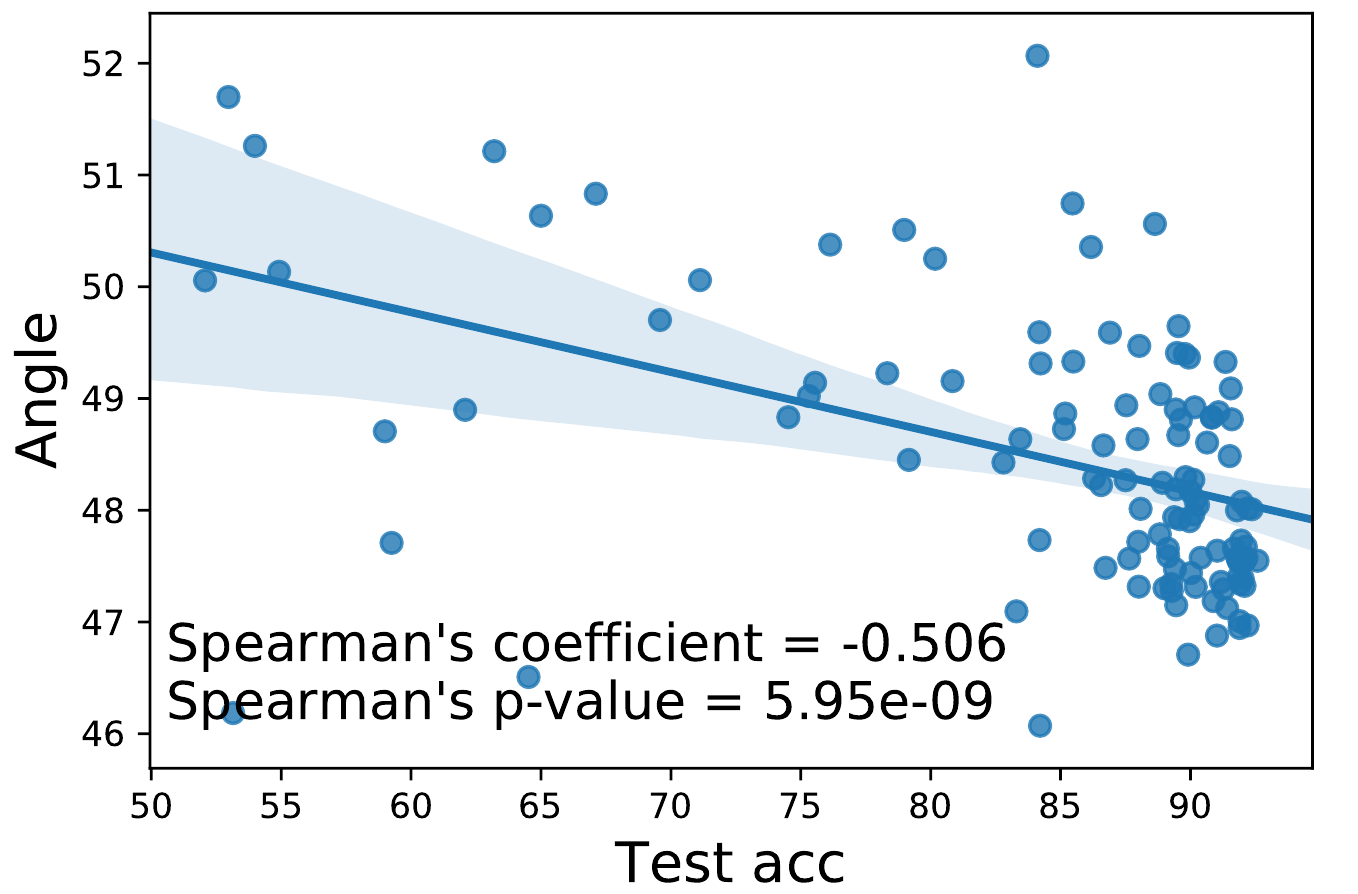}}\\
\subfloat[ResNet56 (Train)]{\includegraphics[width=0.5\textwidth]{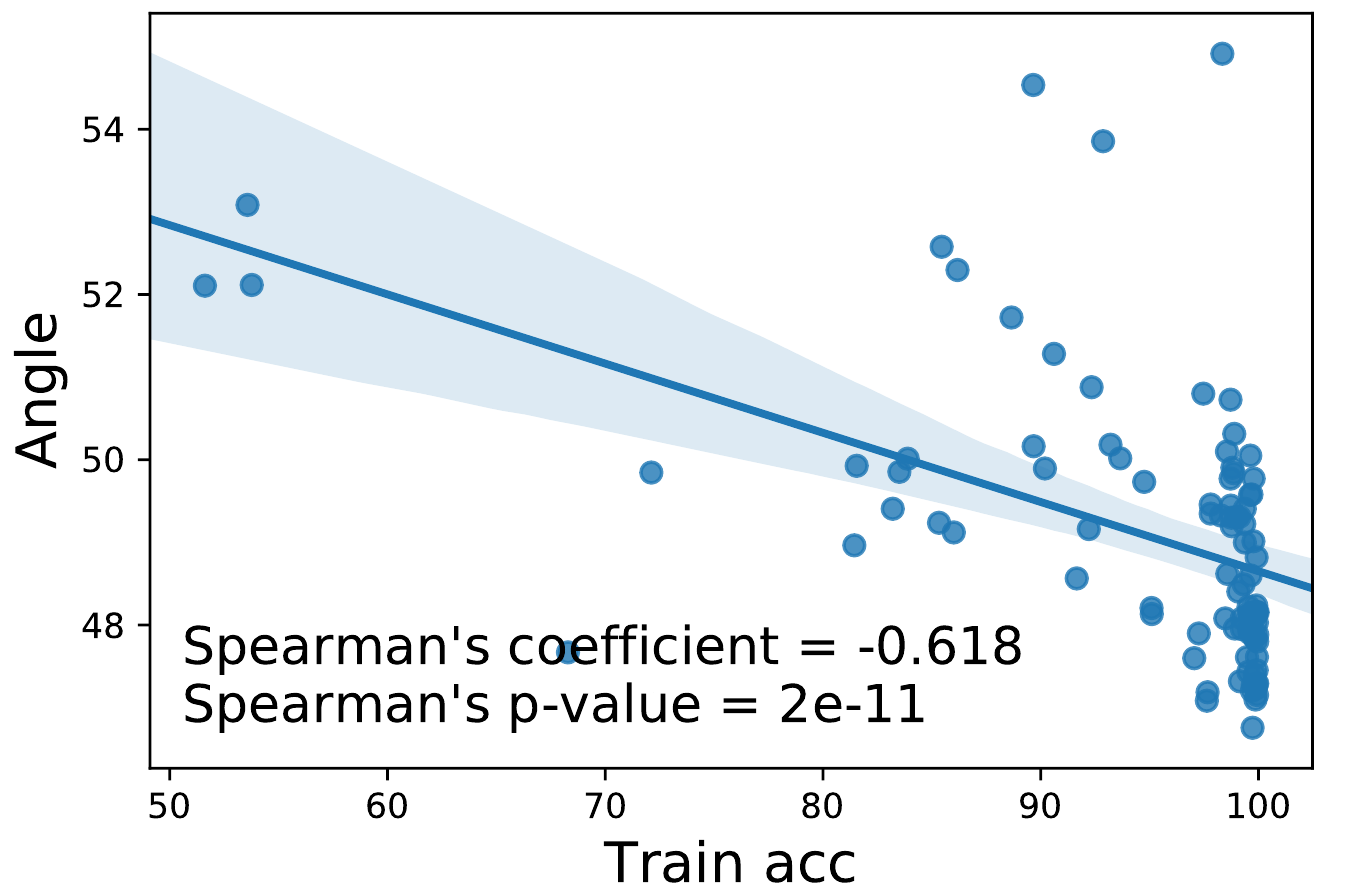}}
\subfloat[ResNet56 (Test)]{\includegraphics[width=0.5\textwidth]{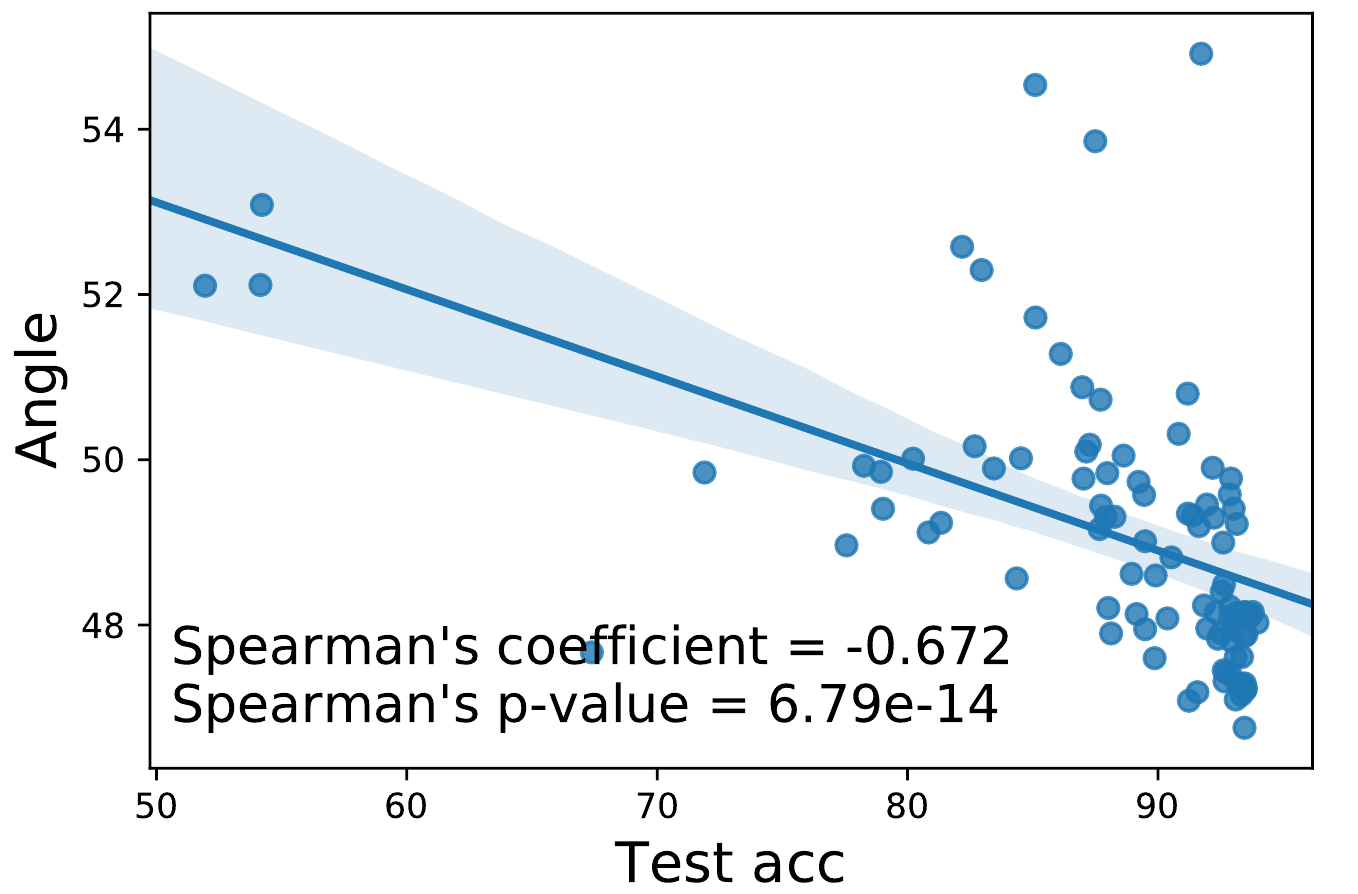}}
\caption{{Strong and consistent correlations between the model performance (training and testing accuracy) and the angels between principal subspaces of raw data and deep features  using $\mathcal{P}$-vectors and CIFAR-10 datasets.} }
\label{fig:claim3-2}
\end{figure*}

\begin{figure*}
\centering
\subfloat[ResNet20 (Train)]{\includegraphics[width=0.5\textwidth]{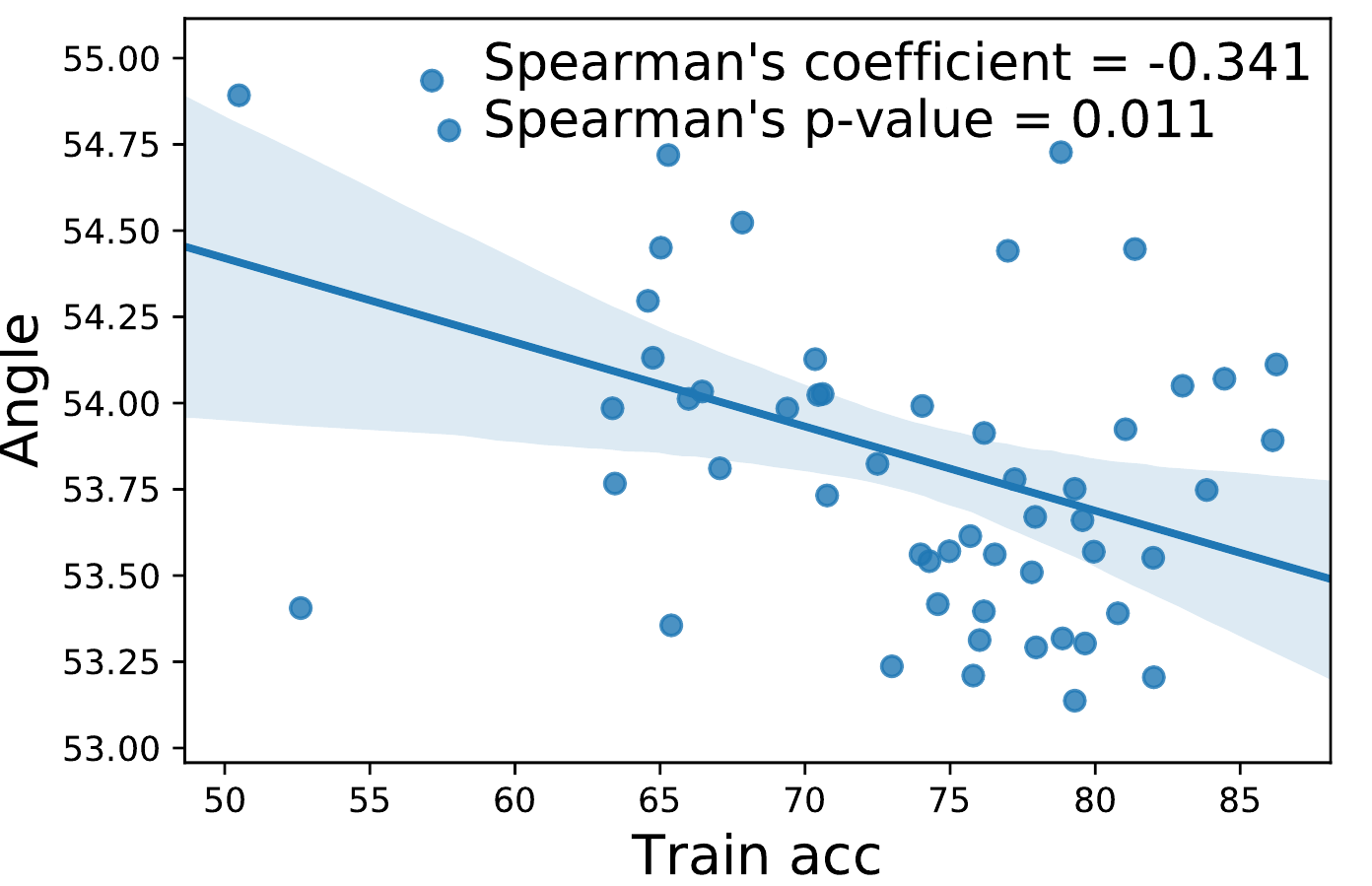}}
\subfloat[ResNet20 (Test)]{\includegraphics[width=0.5\textwidth]{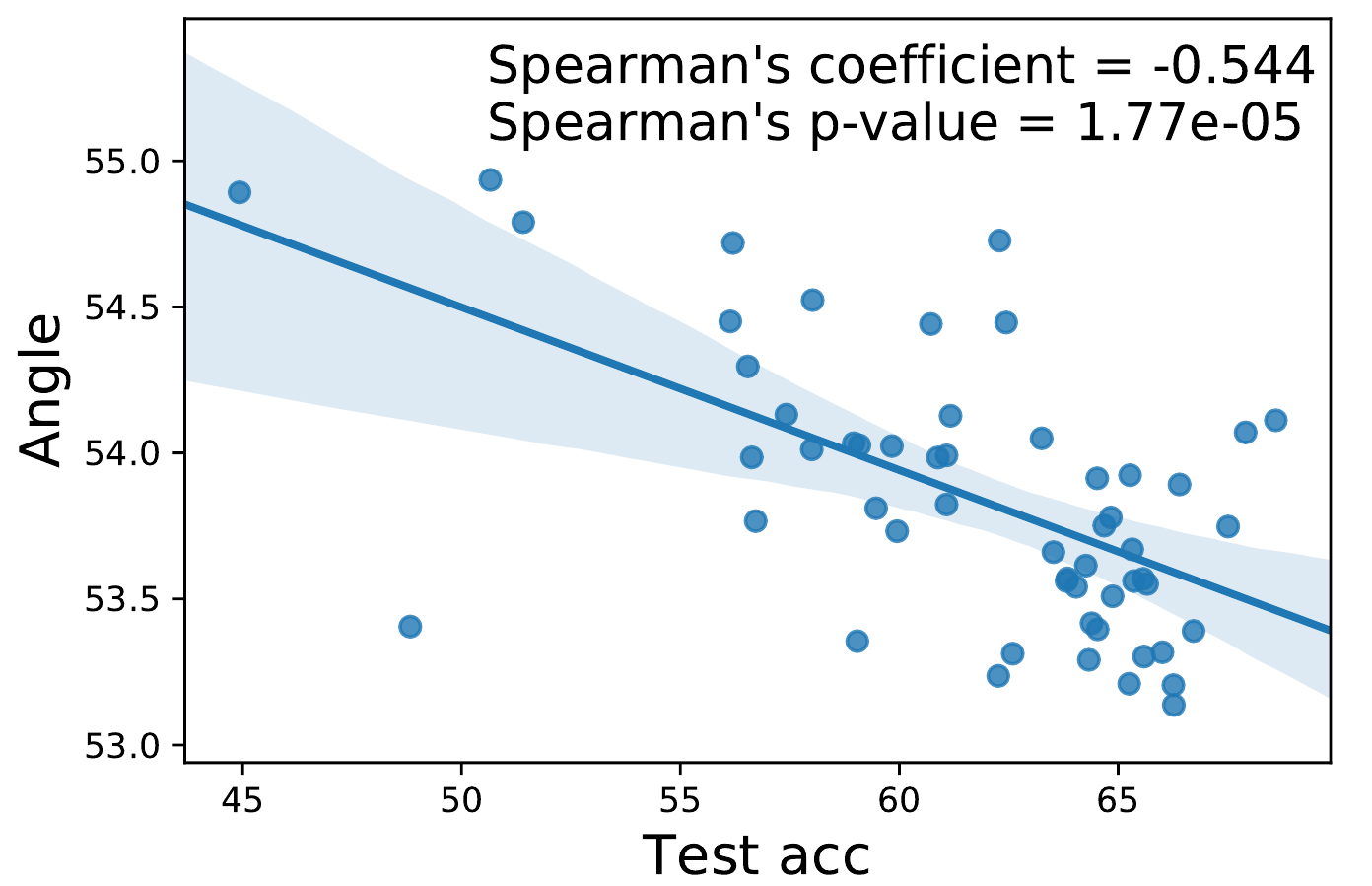}}\\
\subfloat[ResNet56 (Train)]{\includegraphics[width=0.5\textwidth]{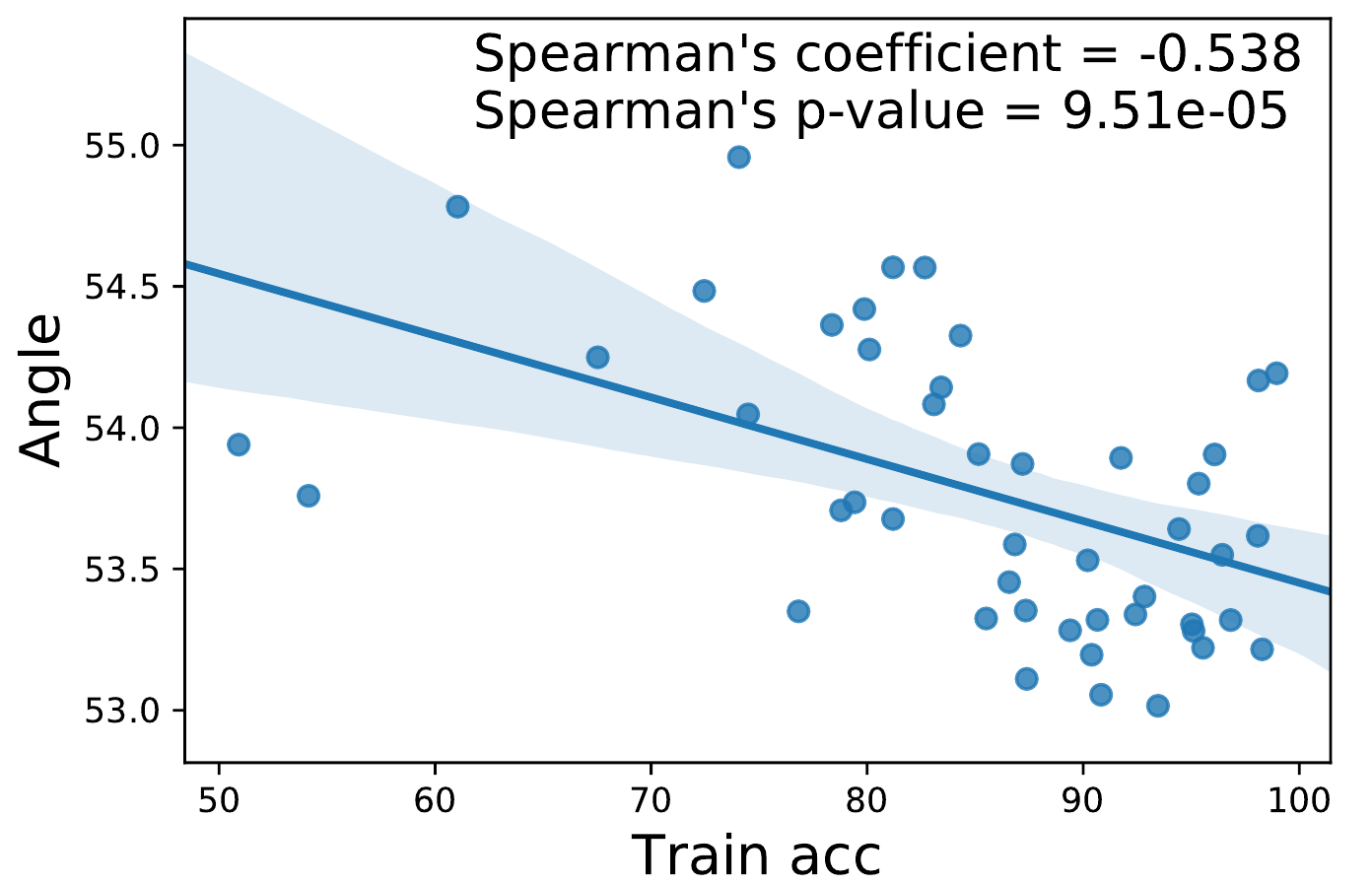}}
\subfloat[ResNet56 (Test)]{\includegraphics[width=0.5\textwidth]{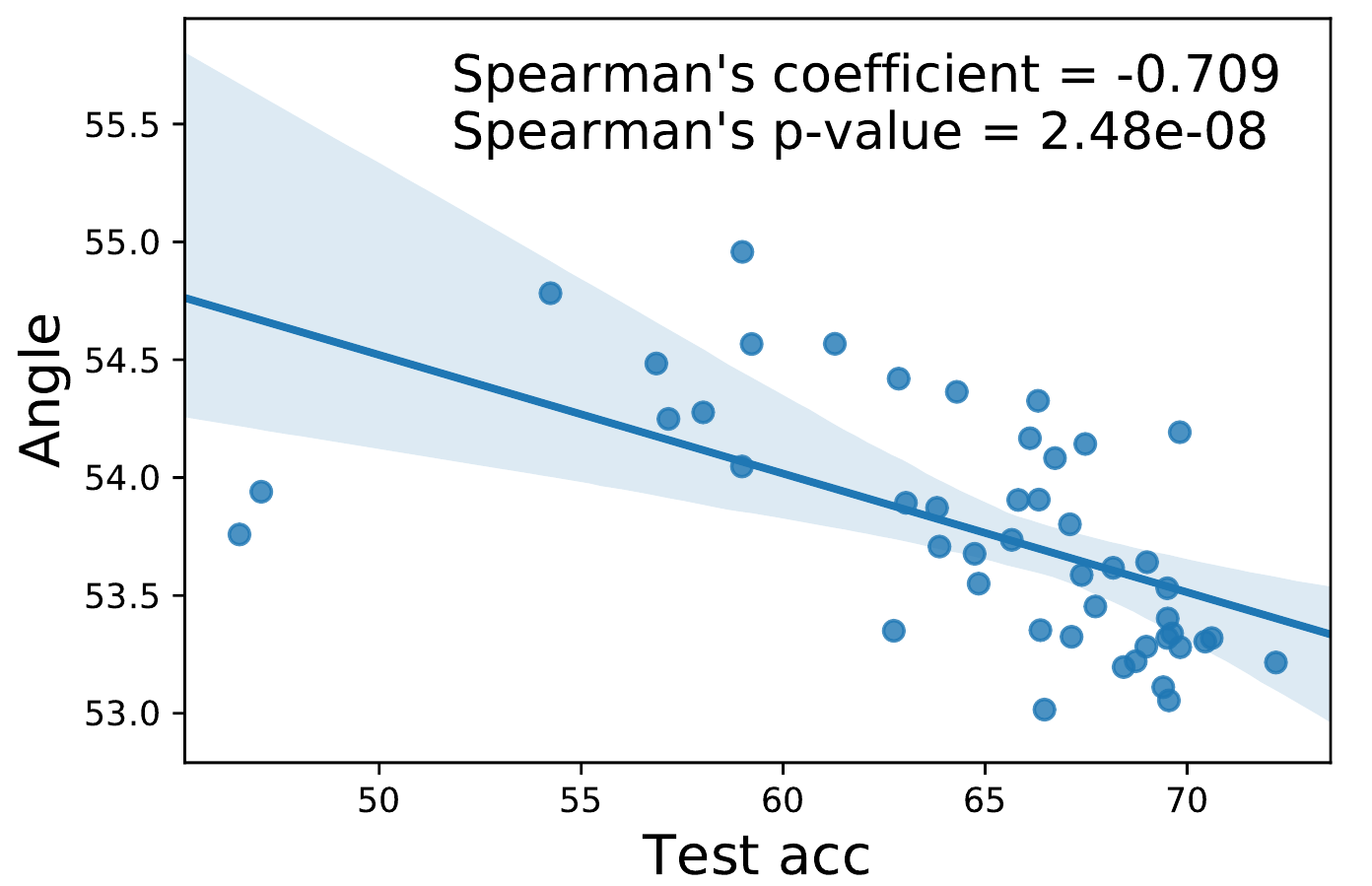}}\\
\subfloat[ResNet110 (Train)]{\includegraphics[width=0.5\textwidth]{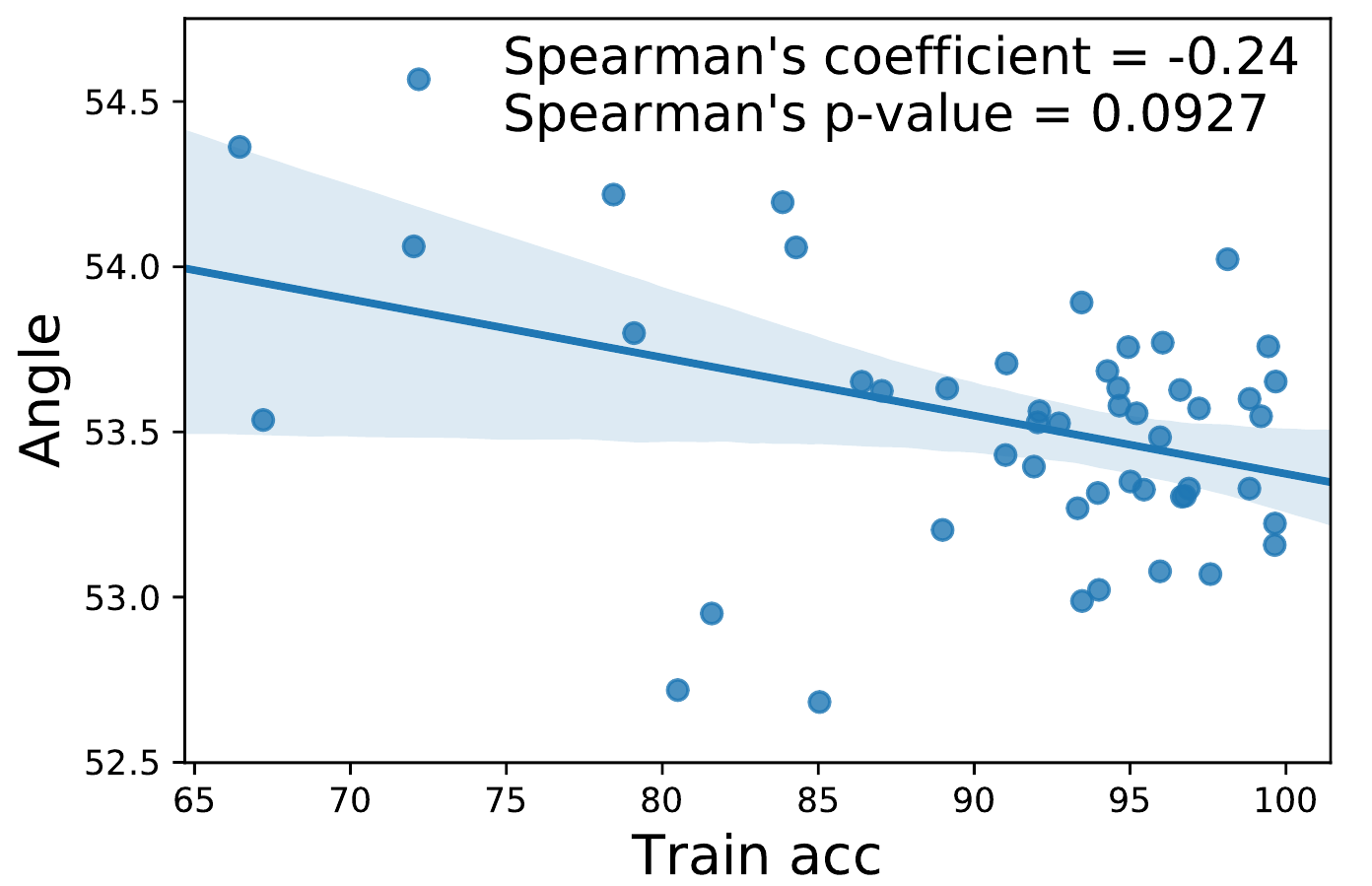}}
\subfloat[ResNet110 (Test)]{\includegraphics[width=0.5\textwidth]{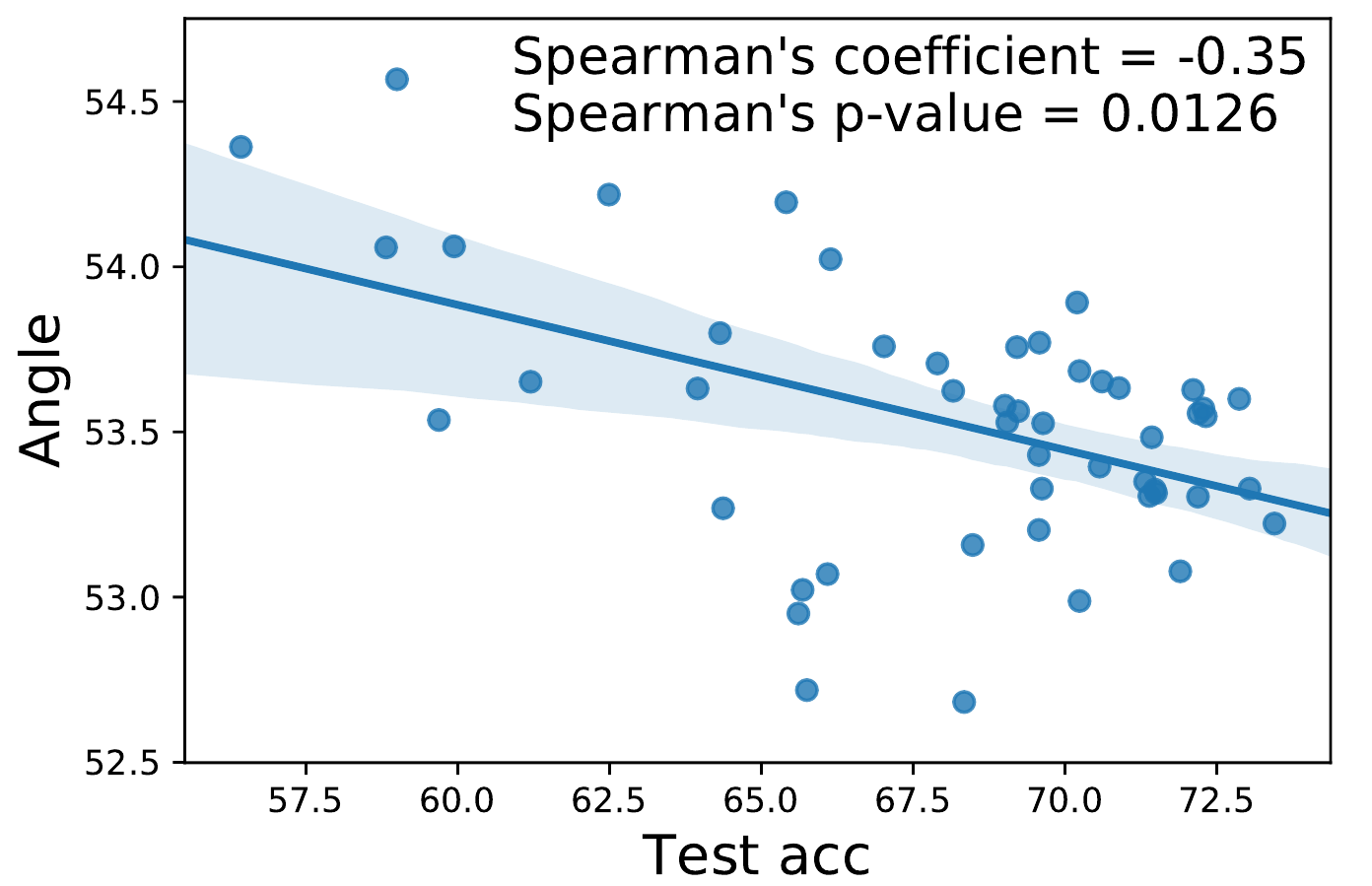}}
\caption{{Strong and consistent correlations between the model performance (training and testing accuracy) and the angels between principal subspaces of raw data and deep features  using $\mathcal{P}$-vectors and CIFAR-100 datasets.} }
\label{fig:claim3-2}
\end{figure*}

\subsection{Correlation between the Angles and Deep Learning Performance}
We further explore the relationship between the performance of the various models on different datasets and the angles between model and data $\mathcal{P}$-vectors for trained ones. Experiments are delivered using ResNet-20/56/110 and VGG-16 on CIFAR-10/100 datasets. As shown in Fig.\ref{fig:claim3-2}, there exists a strong and consistent correlation between the training/testing accuracy of models and the angles between the model and data $\mathcal{P}$-vectors. Note that in these experiments, we use the samples in the training dataset to estimate the model and data $\mathcal{P}$-vectors while \emph{avoiding the use of validation information}, so as the demonstrate the connection of $\mathcal{P}$-vectors all based on training samples to the generalization performance. For CIFAR-100 dataset, we use ResNets with pre-activation enabled.

To avoid the dominance of some outliers, we conduct correlation analysis between the rank of model performance and the rank of angles by the use of Spearman's coefficients and p-values. With $0.05$ as the threshold for significance, we find significance in the correlations between the angles between model and data $\mathcal{P}$-vectors and the training/testing accuracy for all above cases, except the correlation between the angles and training accuracy on ResNet-100 using CIFAR-100 dataset (p-value=$0.0927$). 
In Appendix (A.8), we also include an additional correlation analysis based on log-log plots, where we can further validate our observation. Further for all models on all datasets, the correlation between angles and the testing accuracy is slightly higher than the correlations between angles and training accuracy (and significantly lower p-value).

In this way, we could conclude that (1) both training and testing accuracy are correlated to the angles between the model and data $\mathcal{P}$-vectors, (2) the strong correlations between angles and the testing accuracy might not be caused by the correlations between the angles and training accuracy, as the earlier ones are even stronger, (3) the angles between the model and data $\mathcal{P}$-vectors would be a reasonable performance indicator, as they are strongly, consistently, and significantly correlated to the testing accuracy. This observation could be interpreted that \emph{a significant (locally) linear term exists in the well-trained model~\citep{Zhang2020Empirical}, which makes DNN feature principal subspace correlate to the data principal subspace}.


\subsection{Applications to Generalization Performance Prediction for Deep Models}
To further demonstrate the feasibility of using the $\mathcal{P}$-vector as a ``validation-free'' measure of generalization performance, we use the experiment settings of ``Predicting Generalization in Deep Learning Competition'' at NeurIPS 2020~\cite{jiang2020neurips} to evaluate the \emph{angle between principal subspaces of deep DNN features and raw data} on predicting the generalization performance of models using the training dataset. The competition offers a large number of deep models trained with various hyper-parameters and DNN architectures, while the official evaluator for the competition first predicts the generalization performance of every model using the proposed measure, then verify the prediction results through the mutual information (the higher the better) between the proposed measures and the (observable) ground truth of generalization gaps. 

In the experiments, we propose to use the angles between the model and data $\mathcal{P}$-vectors using the training dataset as the metrics of generalization performance. In the comparisons with the proposed $\mathcal{P}$-vectors, we include several baseline measures in generalization performance predictors, including VC Dimension~\citep{vapnik2013nature}, Jacobian norm w.r.t intermediate layers~\citep{pgdl2020}, Distance from the convergence point to initialization~\citep{nagarajan2019generalization}, and the Sharpness of convergence point~\citep{jiang2019fantastic}. In addition to these methods, we also propose ``Pseudo Validation Accuracy'' as a measure for comparisons, where this measure first uses random data augmentation apply to the original set of training data to generate a set of ``pseudo validation samples'', then tests the accuracy of the model using ``pseudo validation samples''. 

Table~\ref{tab:gengap_pred} presents the comparisons between the proposed measures and baselines. It shows that when the proposed measure -- angles between the model and data $\mathcal{P}$-vectors -- stands alone, the measures significantly outperform the baseline methods including Jacobian norm w.r.t intermediate layers and the VC dimensions. However, through complementing with other metrics, the metrics based on $\mathcal{P}$-vector angles could be further improved in predicting the generalization performance and finally outperform all baseline methods when combining with ``Pseudo validation accuracy''. Note that we combine the results of two metrics through weighted aggregation~(\cite{pihur2009rankaggreg}, with a constant weight 0.05) of two ranking lists that are sorted according to the two metrics respectively. 


\begin{table}
    \centering
    \begin{tabular}{|l|c|}
        \hline
        Methods	&	Prediction Score	\\
        \hline
        VC Dimension~\citep{vapnik2013nature}	&	0.020 	\\
        Jacobian norm w.r.t intermediate layers' features~\citep{pgdl2020}	&	2.061 	\\
        \textbf{$\mathcal{P}$-vector}	&	\textbf{3.325} 	\\
        Distance to the Initialization Point~\citep{nagarajan2019generalization}	&	4.921 	\\
        \textbf{Distance to Initialization Point + $\mathcal{P}$-vector}	&	\textbf{4.971} 	\\
        Sharpness of the Convergence Point~\citep{jiang2019fantastic}	&	10.667 	\\
        Pseudo Validation Accuracy	&	13.531 	\\
        \textbf{Pseudo Validation Accuracy + $\mathcal{P}$-vector}	&	\textbf{15.618} 	\\
        \hline
    \end{tabular}
    \caption{Scores of different method to predict the generalization gap using CIFAR-10 and SHVN.}
    \label{tab:gengap_pred}
\end{table}

\section{Conclusion}
In this work, we study principal subspaces of deep features learned in DNNs, we first propose the $\mathcal{P}$-vector (the top left singular vector of the \#samples$\times$\#feature matrix extracted from DNN) as the proxy measurement of the principal subspace and use the angles between $\mathcal{P}$-vectors to compare principal subspaces.
%
%
We observe that, no matter which architectures or whether the labels have been used to train the models, the angle between principal subspaces of different DNNs (measured using $\mathcal{P}$-vectors)
%
would decrease to smaller ones (e.g., around $10^\circ$ for most models in this study with cosine similarity higher than 0.985), when the models are trained using the same dataset. We thus conclude that these DNNs would share a common principal subspace in deep feature space.
Furthermore, during the training procedure from the random scratch, the principal subspace of deep features would
%
slowly approach to the principal subspace of data vectors from an almost-orthogonal status (e.g., 80$^\circ$--90$^\circ$) to smaller angles (e.g., 50$^\circ$--60$^\circ$).~
%

Finally, we find that angles between principal subspaces of deep features and data vectors are strongly correlated to the performance of models while they are capable of predicting generalization performance, even when $\mathcal{P}$-vectors are all estimated using training dataset only. As was discussed, we believe the empirical observations obtained here are partially due to the local linearity of DNN models~\citep{Zhang2020Empirical}, while we have validate the significance of $\mathcal{P}$-vectors to measure the principal subspace of deep features in Section  3.2. Our future work may focus on the theoretical understanding to these phenomena. 
\section{Declarations}
Not applicable.
\bibliographystyle{unsrt}
\bibliography{main}

\clearpage
\appendix
\section{Appendix}
This appendix includes results of additional experiments.
\subsection*{A.0 Comparison of angles between  $\mathcal{P}$-vector extracted from well-trained models Using Different Architectures on ImageNet}
\begin{figure}
\centering
\subfloat[ImageNet (Train)]{\includegraphics[width=0.75\textwidth]{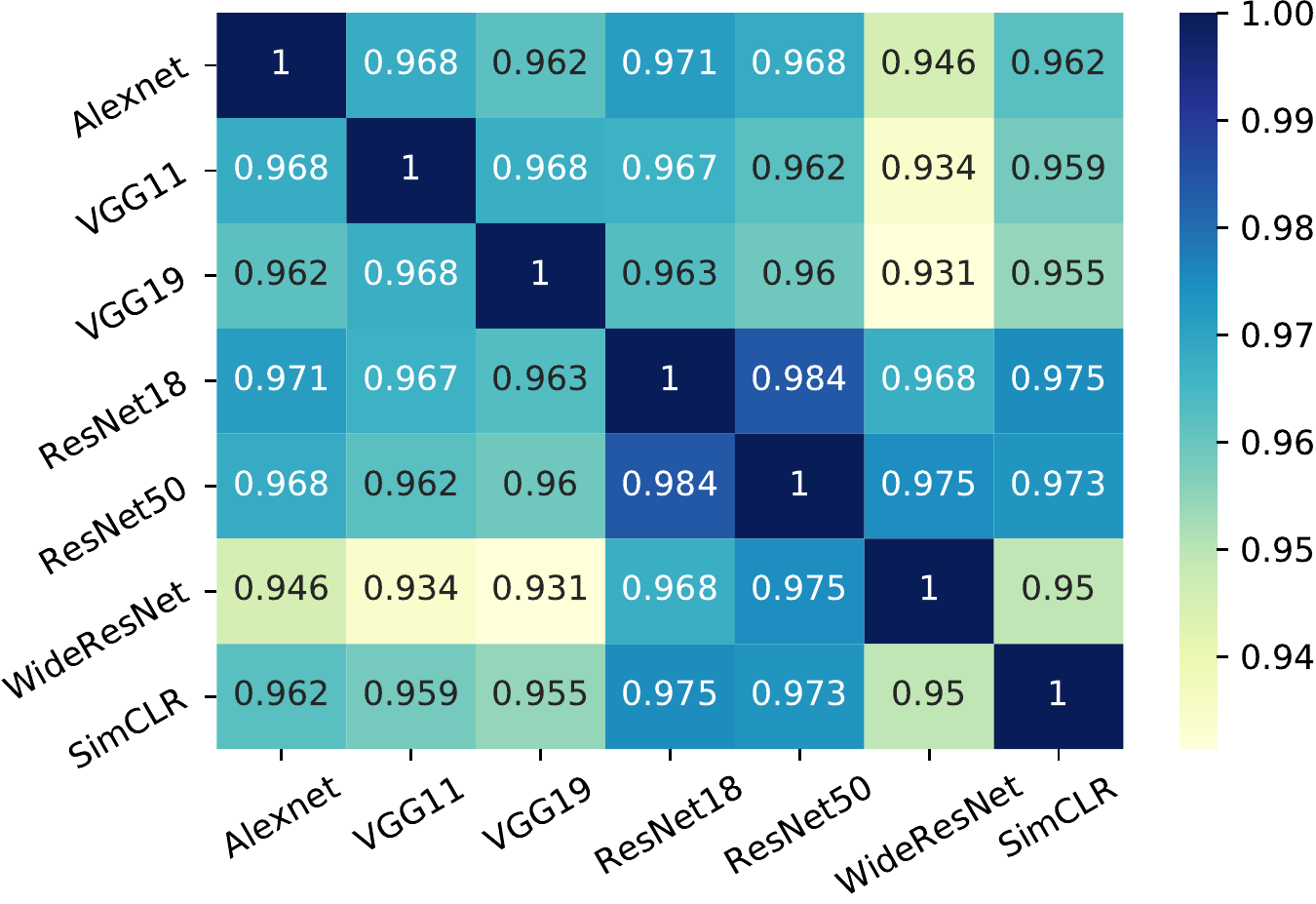}}\\
\subfloat[ImageNet (Test)]{\includegraphics[width=0.75\textwidth]{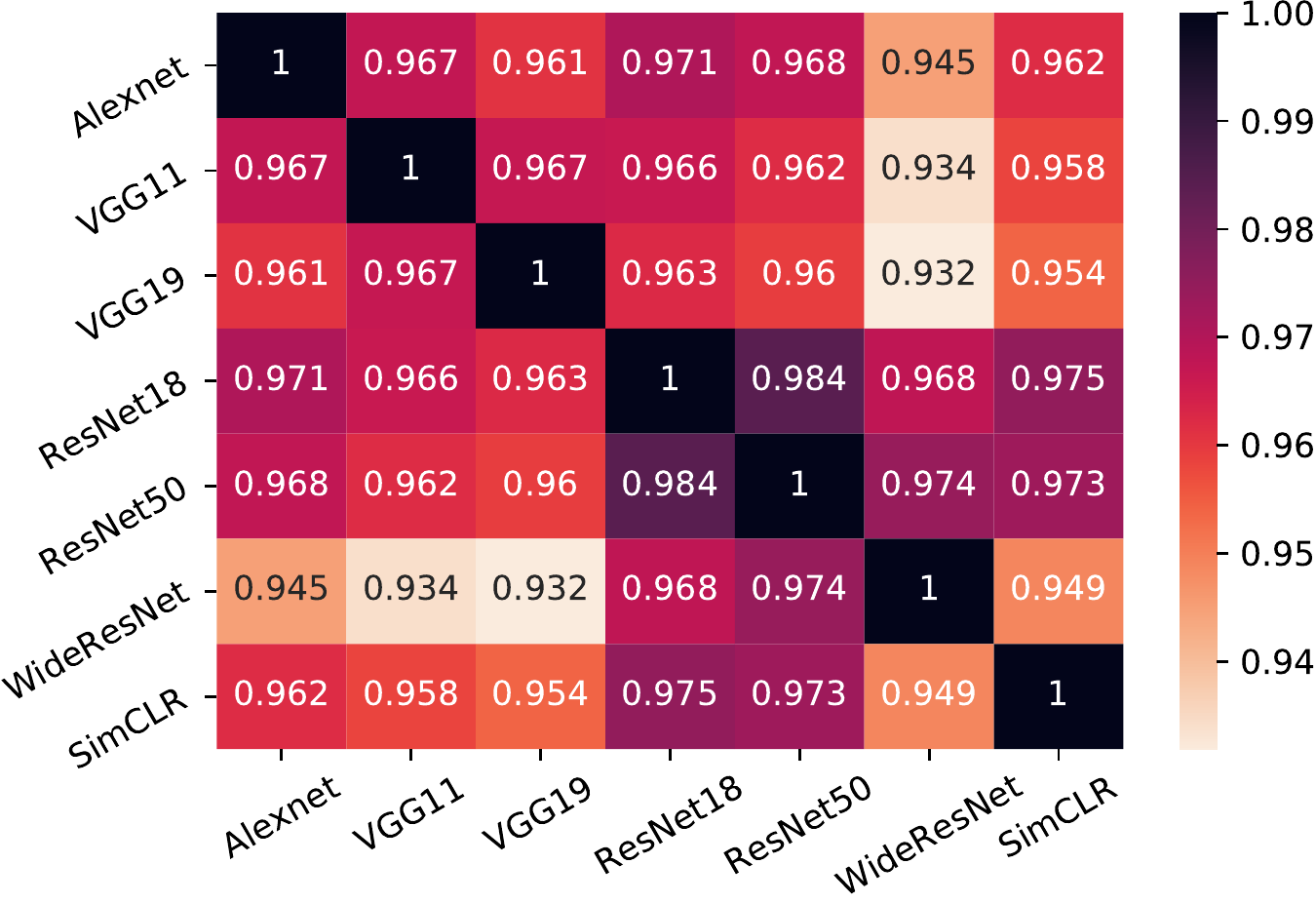}}
\caption{Cosine of angles between principal subspaces of deep features, measured using $\mathcal{P}$-vectors, for models trained under default settings.}
\label{fig:claim1-app}
\end{figure}

\subsection{Comparison of angles between checkpoints and well-trained $\mathcal{P}$-vector Using Different Architectures on CIFAR-100}
\begin{figure*}[hbtp]
\centering
\subfloat[Supervised (Train)]{\includegraphics[width=0.5\textwidth]{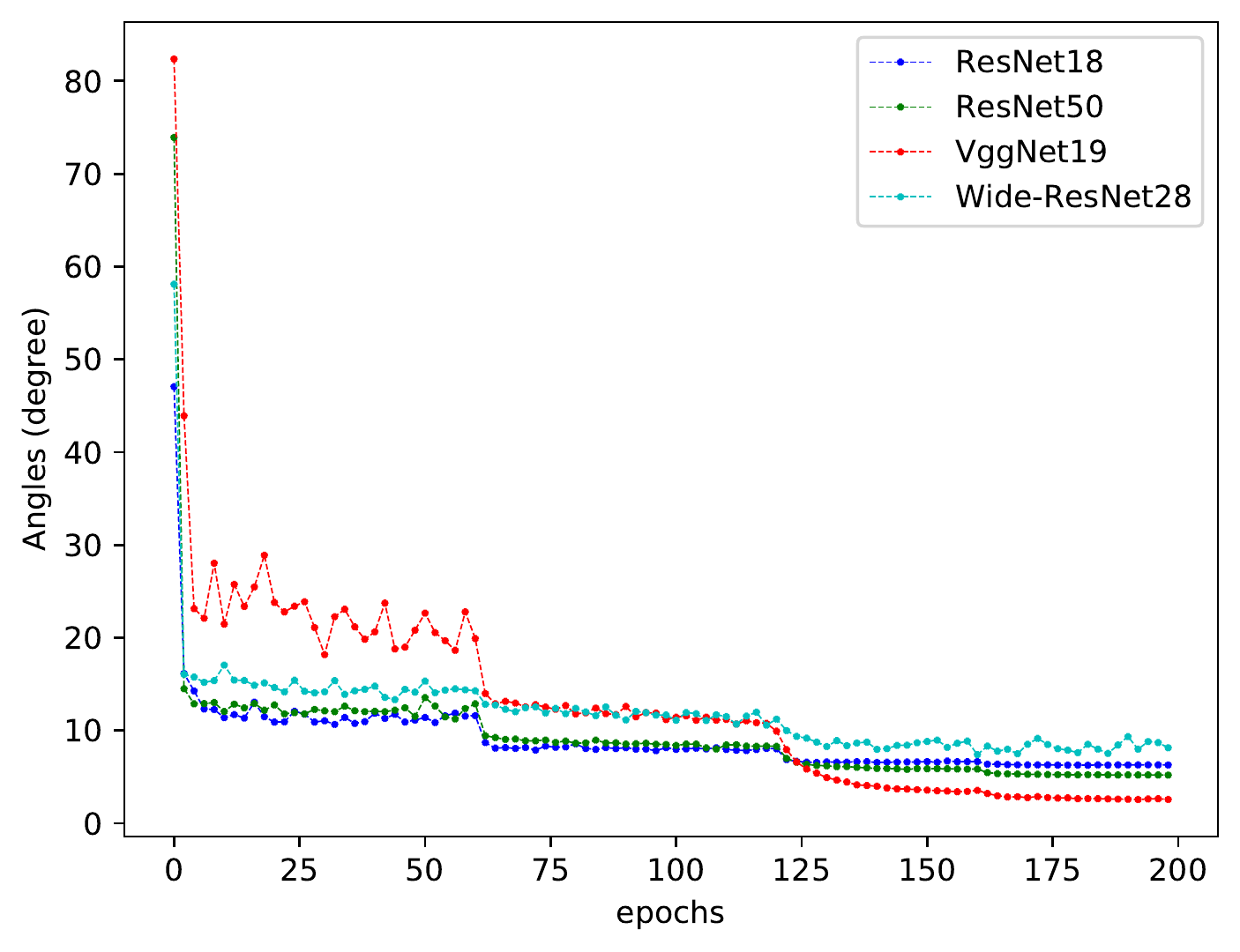}}
\subfloat[Supervised (Test)]{\includegraphics[width=0.5\textwidth]{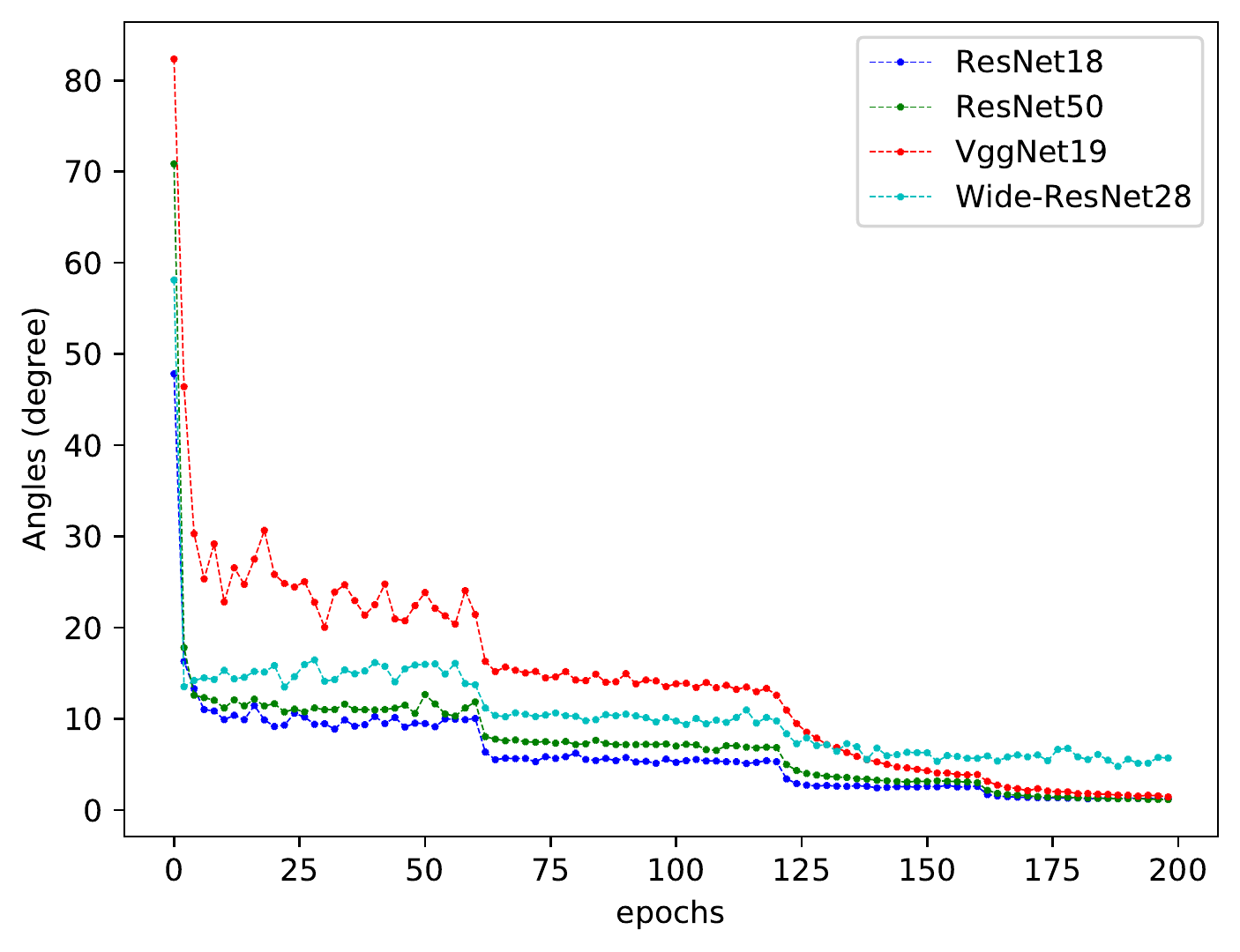}}\\
\subfloat[Unsupervised (Train)]{\includegraphics[width=0.5\textwidth]{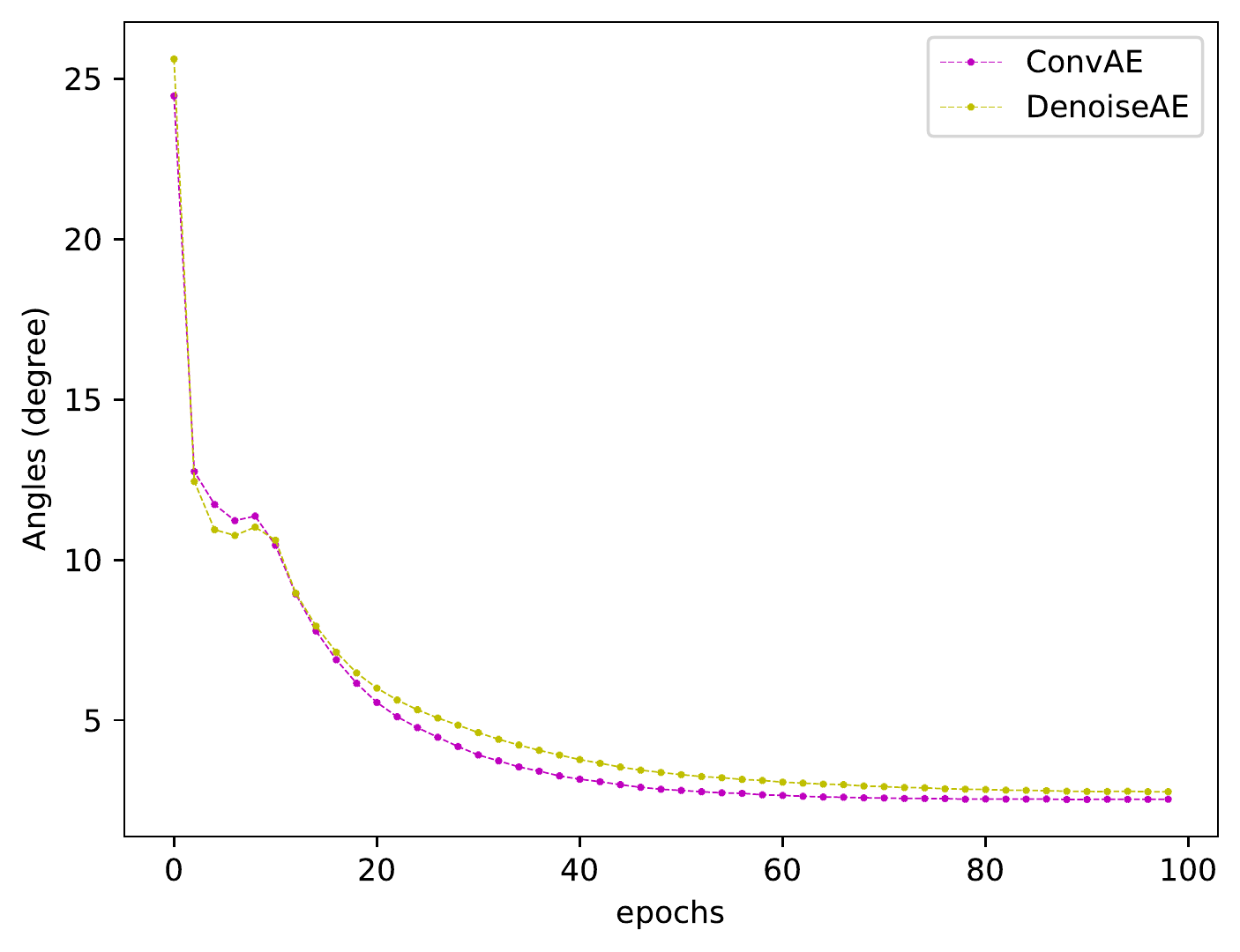}}
\subfloat[Unsupervised (Test)]{\includegraphics[width=0.5\textwidth]{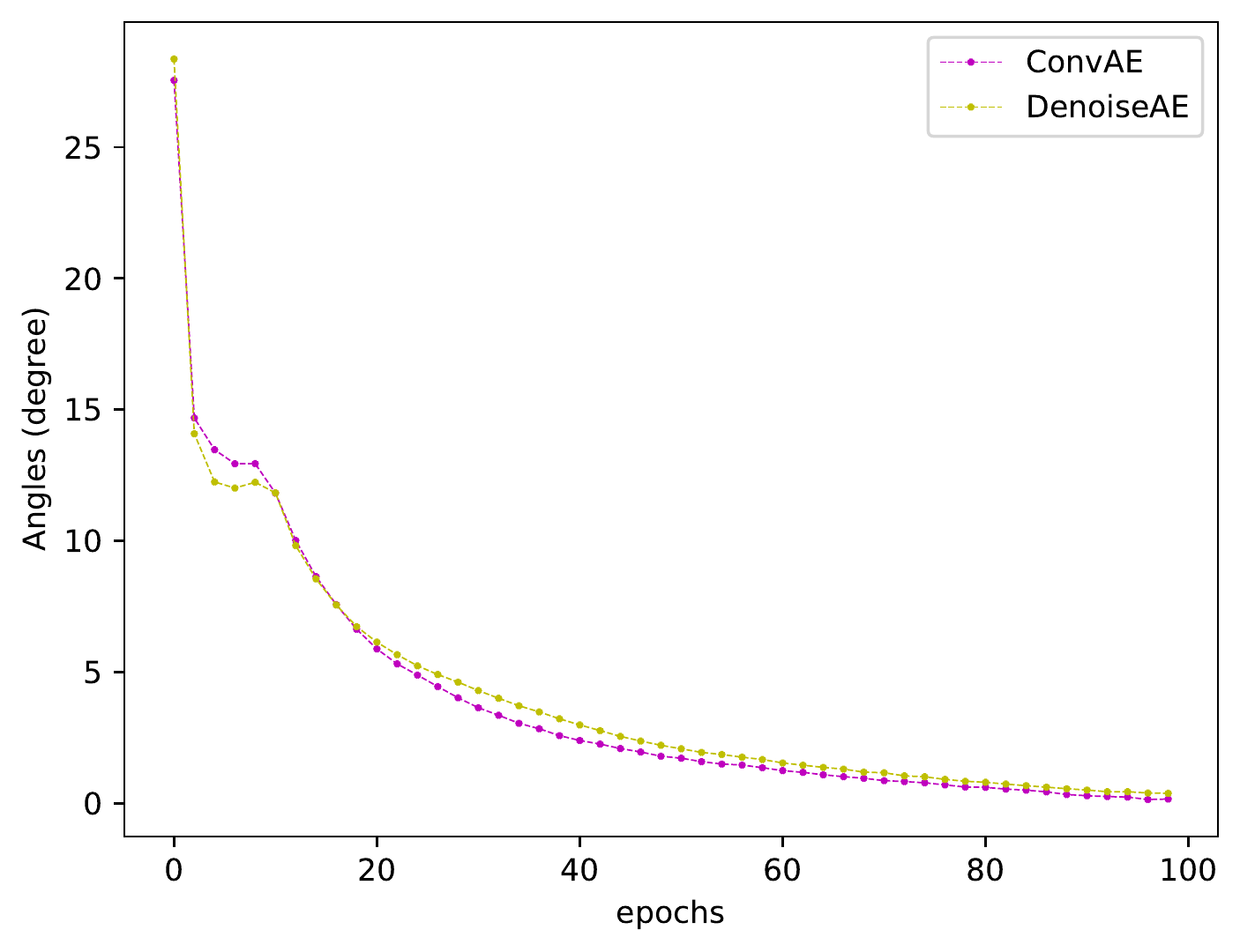}}\\
\subfloat[Self-Sup.
(Train)]{\includegraphics[width=0.5\textwidth]{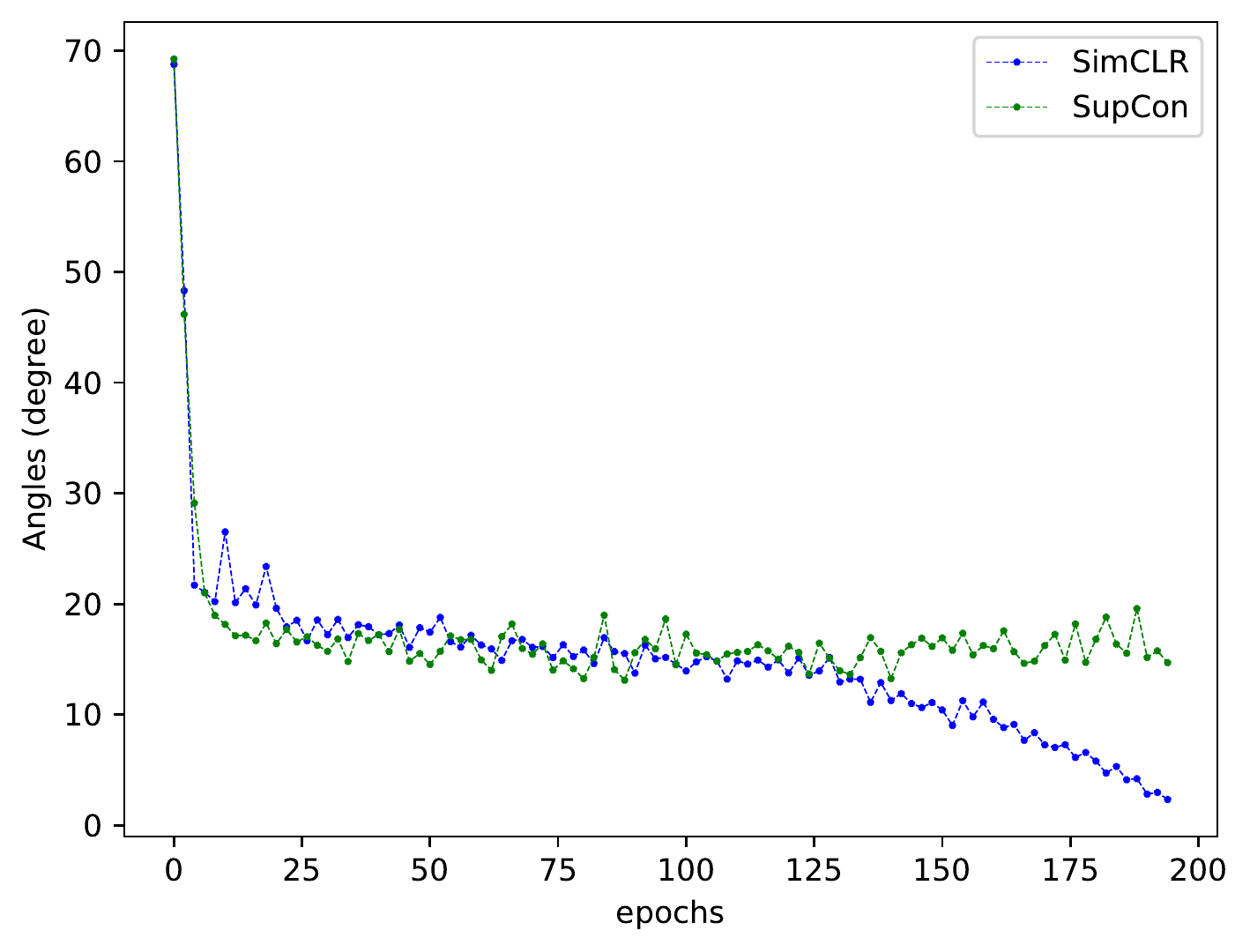}}
\subfloat[Self-Sup. (Test)]{\includegraphics[width=0.5\textwidth]{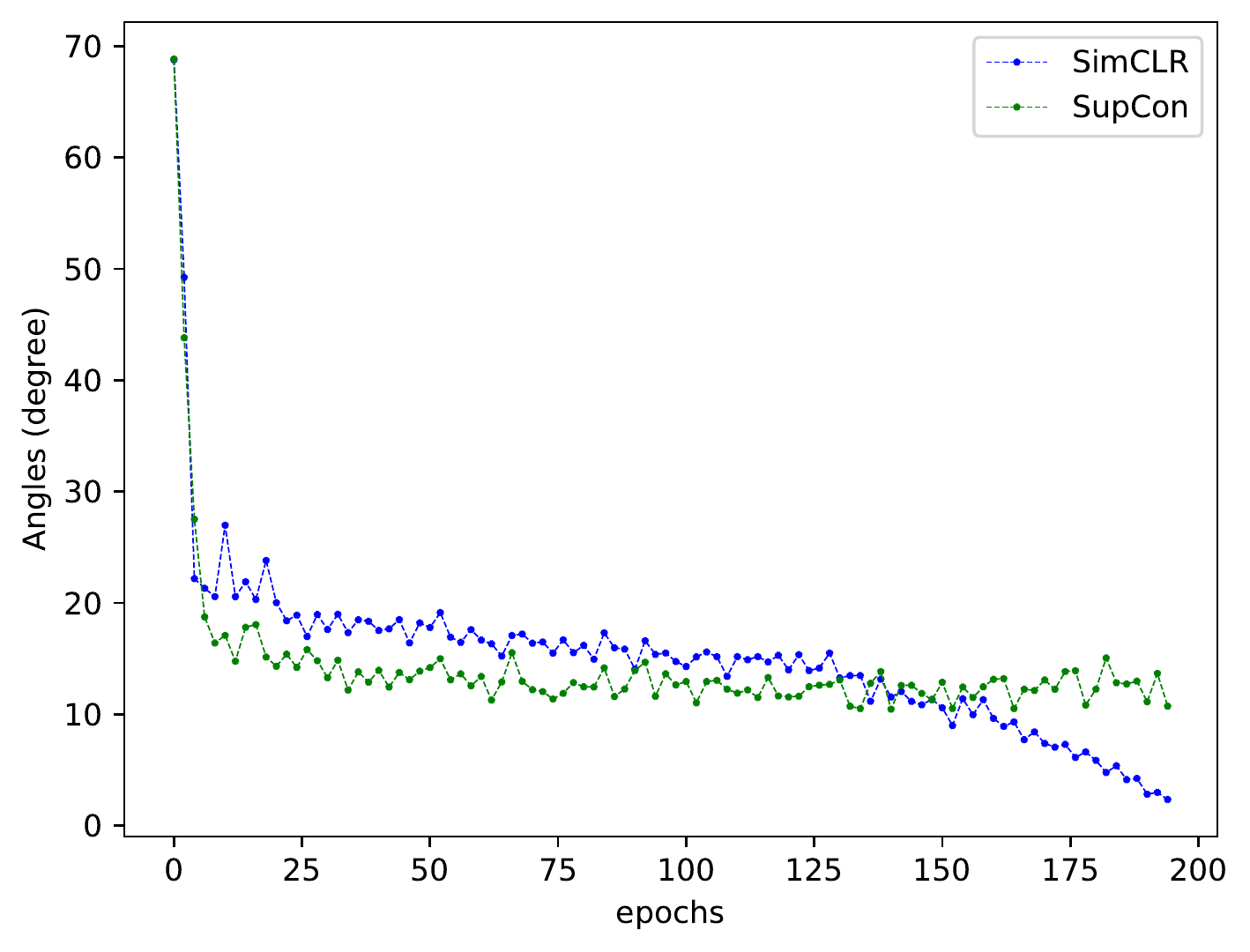}}
\caption{Convergence to the Common Feature Subspace with CIFAR-100.  Curves of angles of $\mathcal{P}$-vectors between the well-trained model and its checkpoint per training epoch of three learning supervisory manners. The trends of convergence for the angles can be observed in all models. }
\label{fig:claim2_1_cifar100}
\end{figure*}

In the main text, we presented the result on CIFAR-10 dataset. To generalize the observations, we repeated the experiments on CIFAR-100 dataset to validate our hypothesis of the convergence of the angles between model checkpoints and well-trained model $\mathcal{P}$-vectors. We investigate the change of angles over the $\mathcal{P}$-vectors of training model checkpoints per epoch with comparison to the $\mathcal{P}$-vectors of well-trained models (model of epoch 200 in our case).  As shown in Fig.\ref{fig:claim2_1_cifar100}, a gradually decreasing manner of the curves for the angle between $\mathcal{P}$-vectors and all angles between $\mathcal{P}$-vectors cross models with different supervisory manners generally converge to a value that smaller than 10$^\circ$ degree. We can conclude that the hypothesis of the existence of common subspace during the learning procedure also stands on the experiments with CIFAR-100 dataset. 

\subsection{The Non-monotonic Trend within the First Epoch}
\begin{figure}
\centering
\subfloat[Supervised CIFAR-10]{\includegraphics[width=0.5\textwidth]{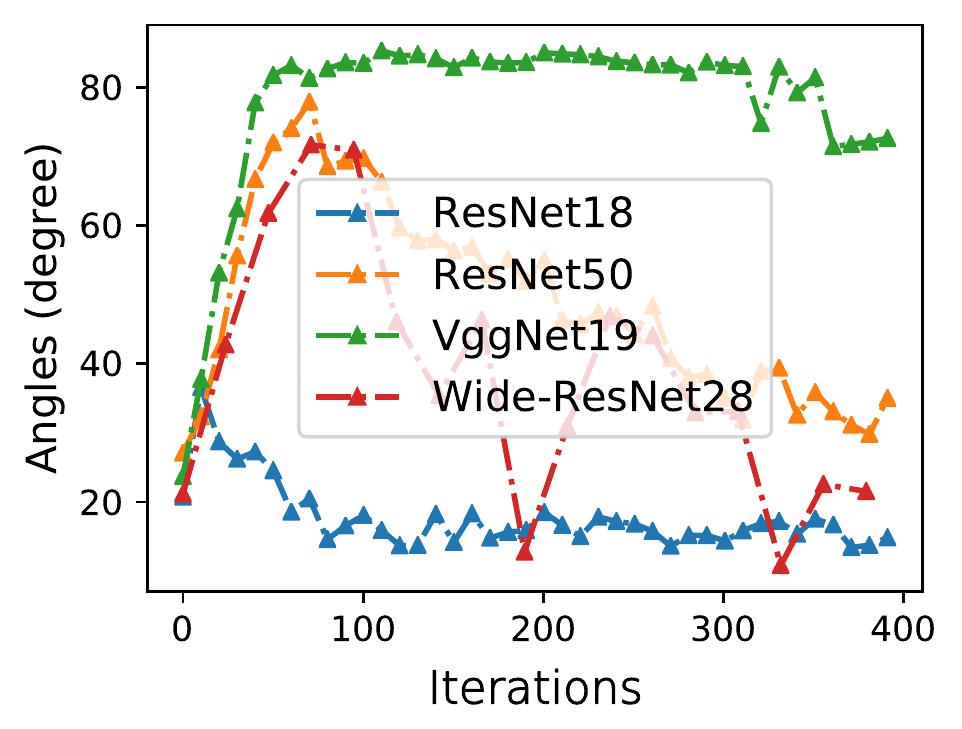}}
\subfloat[Supervised CIFAR-100]{\includegraphics[width=0.5\textwidth]{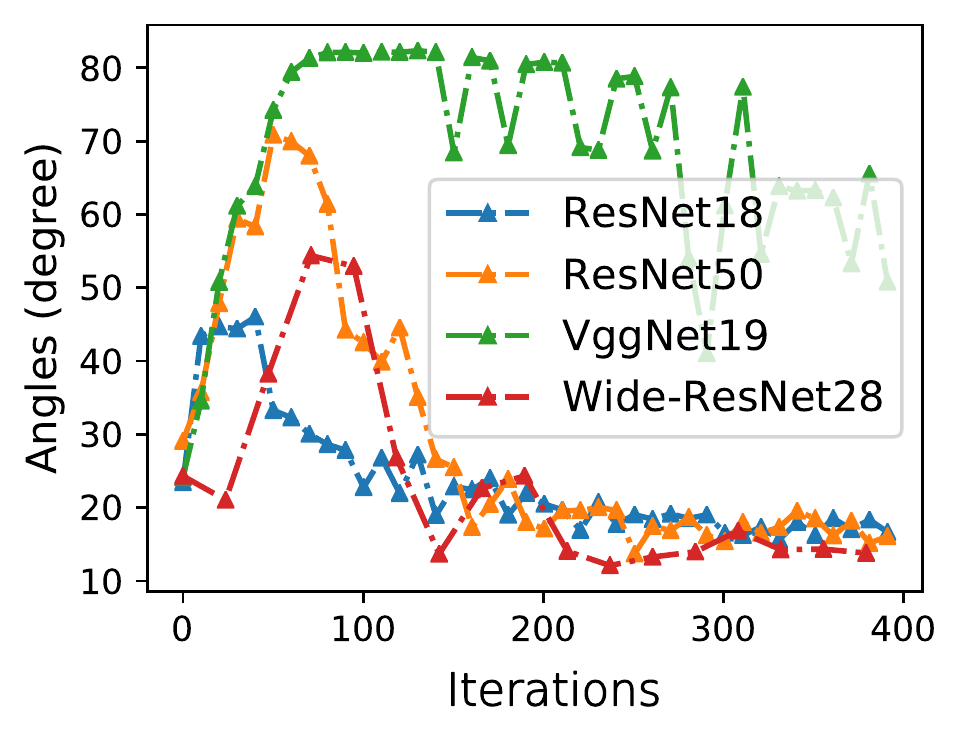}}\\
\subfloat[Unsupervised CIFAR-10]{\includegraphics[width=0.5\textwidth]{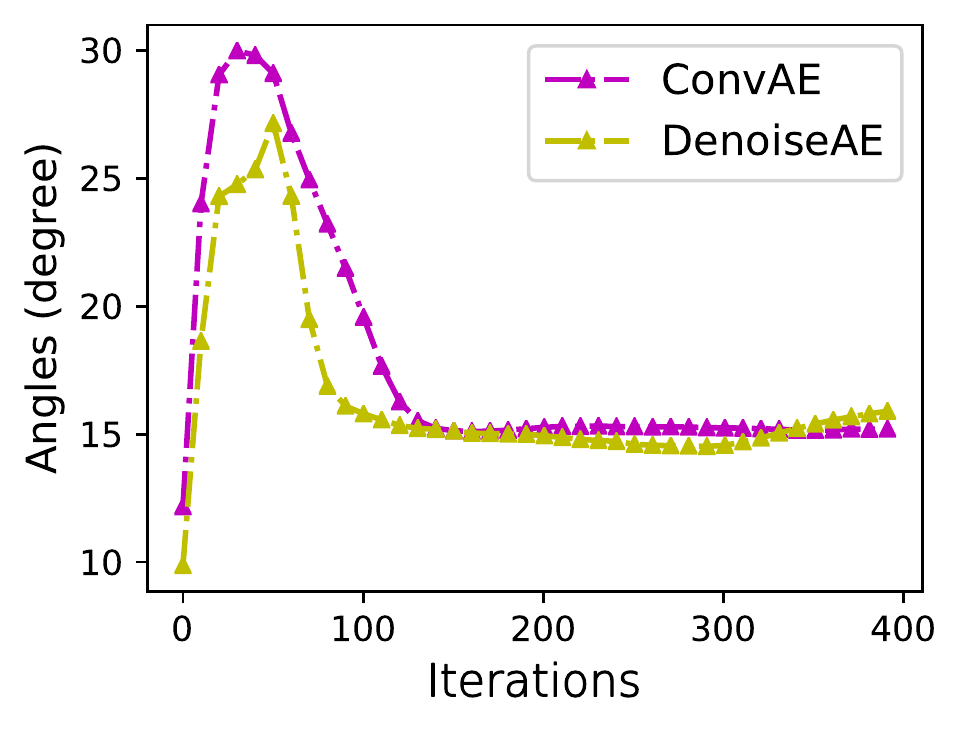}}
\subfloat[Unsupervised CIFAR-100]{\includegraphics[width=0.5\textwidth]{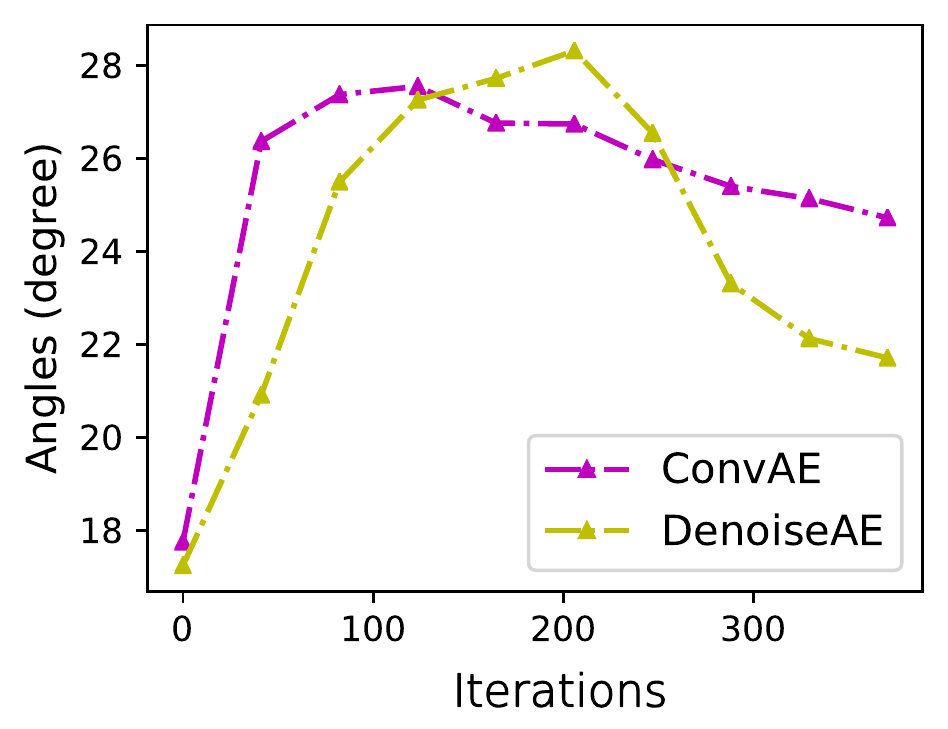}}\\
\subfloat[Self-Supervised CIFAR-10]{\includegraphics[width=0.5\textwidth]{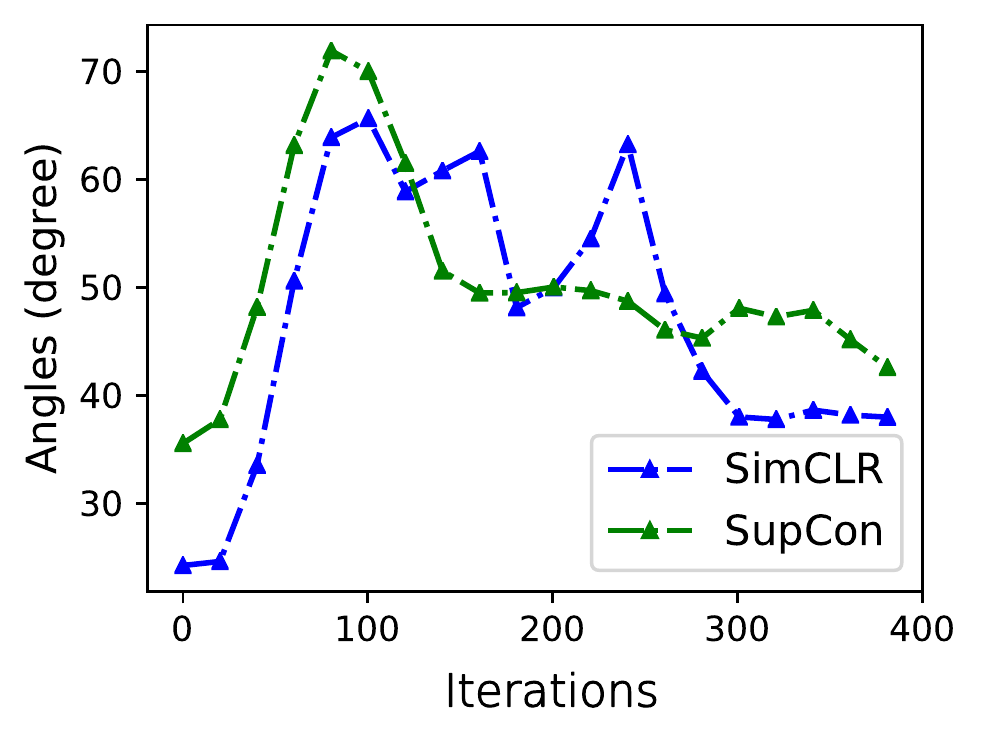}}
\subfloat[Self-Supervised CIFAR-100]{\includegraphics[width=0.5\textwidth]{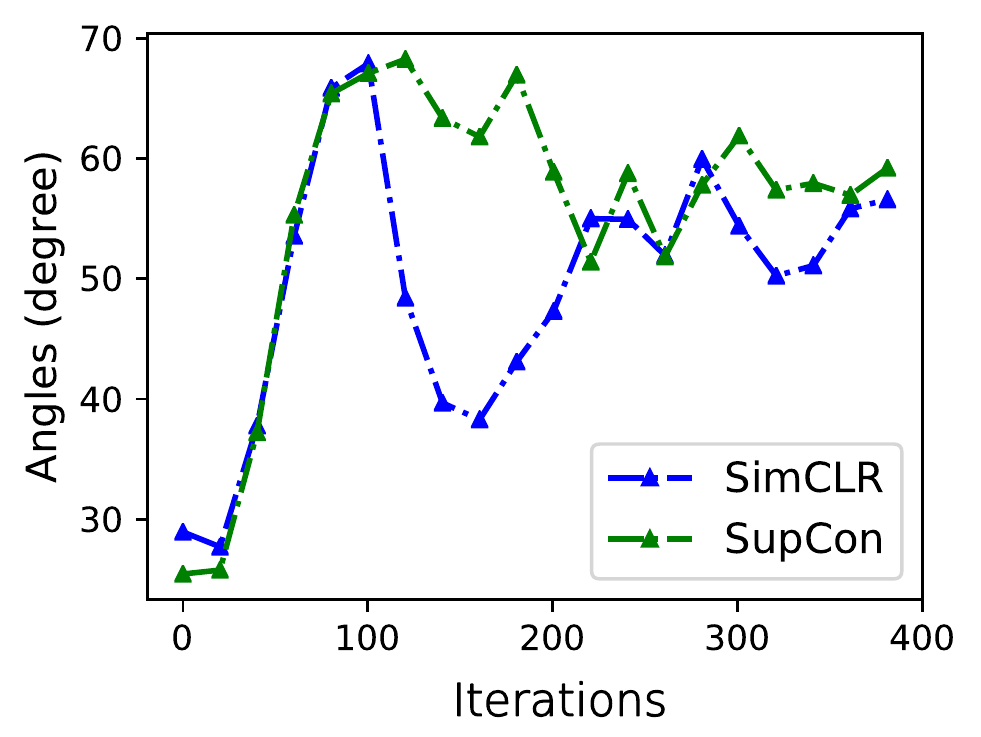}}
\caption{{Angles between principal subspaces, measured using $\mathcal{P}$-vectors based on the Testing Sets of CIFAR-10 and CIFAR-100, between well-trained models and checkpoints per training iteration in the first epoch.} }
\label{fig:epoch0-test}
\end{figure}

The variation of angles between $\mathcal{P}$-vectors for the well-trained model and its checkpoint per training epoch of each iteration in the first epoch. The non-monotonic trends within the first epoch also incorporate in the experiments using the testing sets of CIFAR-10 and CIFAR-100 datasets. Fig.\ref{fig:epoch0-test} shows the curves indicating the variation of angles between the training model $\mathcal{P}$-vectors and the well-trained model $\mathcal{P}$-vectors in the iterations in the first epoch. As we use 128 as the batch size in training procedure, the number of iterations for updates is 391 per epoch. We obtain the observation of a non-monotonic trend that the angle first rises with the random initialization and drop down. And in the rest of training process, the angles keeps the approximately monotonically decreasing and converging to small values. The experiments shows consistent result and conclusion on the testing set of CIFAR-10 and CIFAR-100 with the discussion in section 3.

\subsection{Model-to-model Common Subspace}
\begin{figure*}[hbtp]
\centering
\subfloat[Sup. vs. Sup.]{\includegraphics[width=0.33\textwidth]{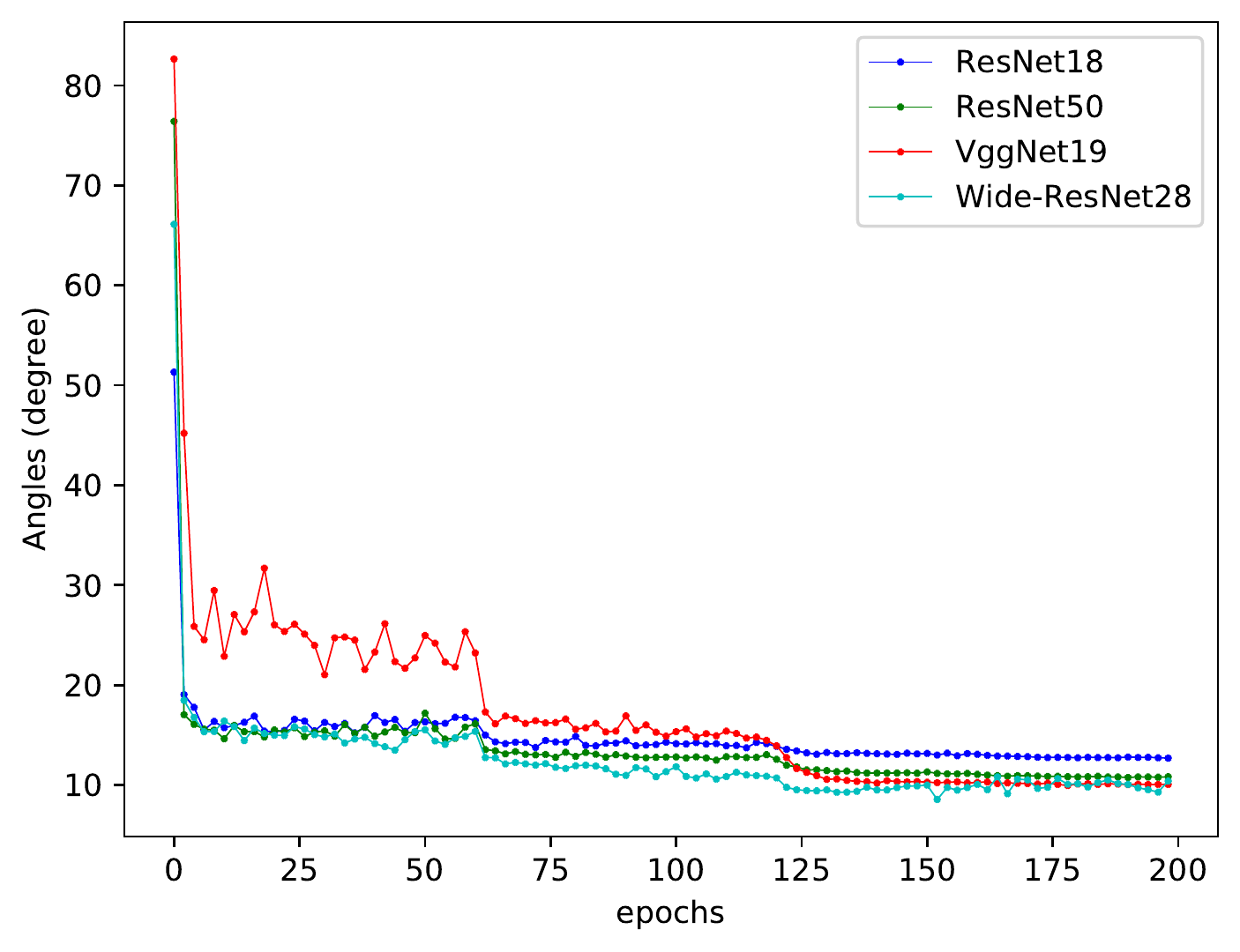}}
\subfloat[Sup. vs. Unsup.]{\includegraphics[width=0.33\textwidth]{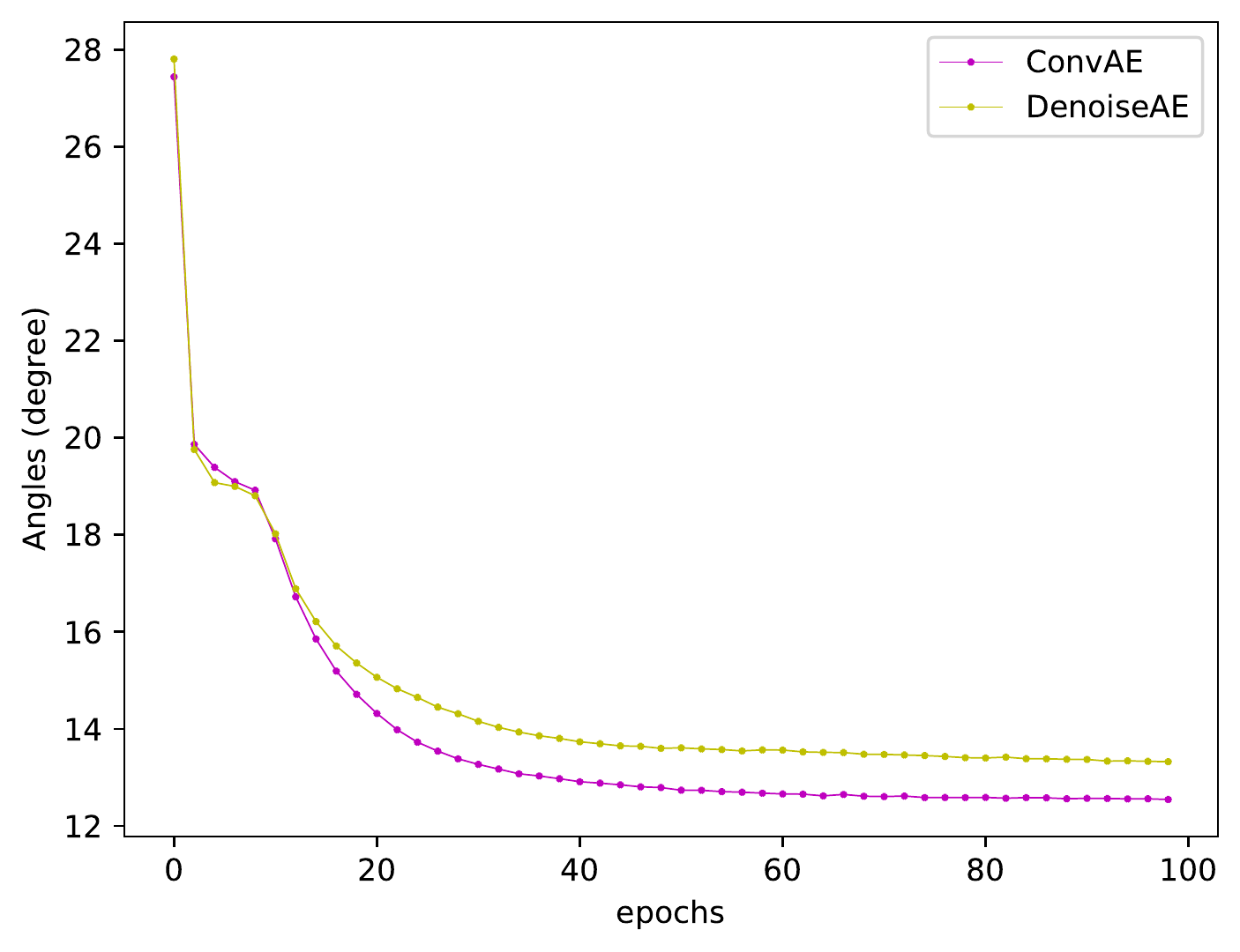}}
\subfloat[Sup. vs. Self-Sup.]{\includegraphics[width=0.33\textwidth]{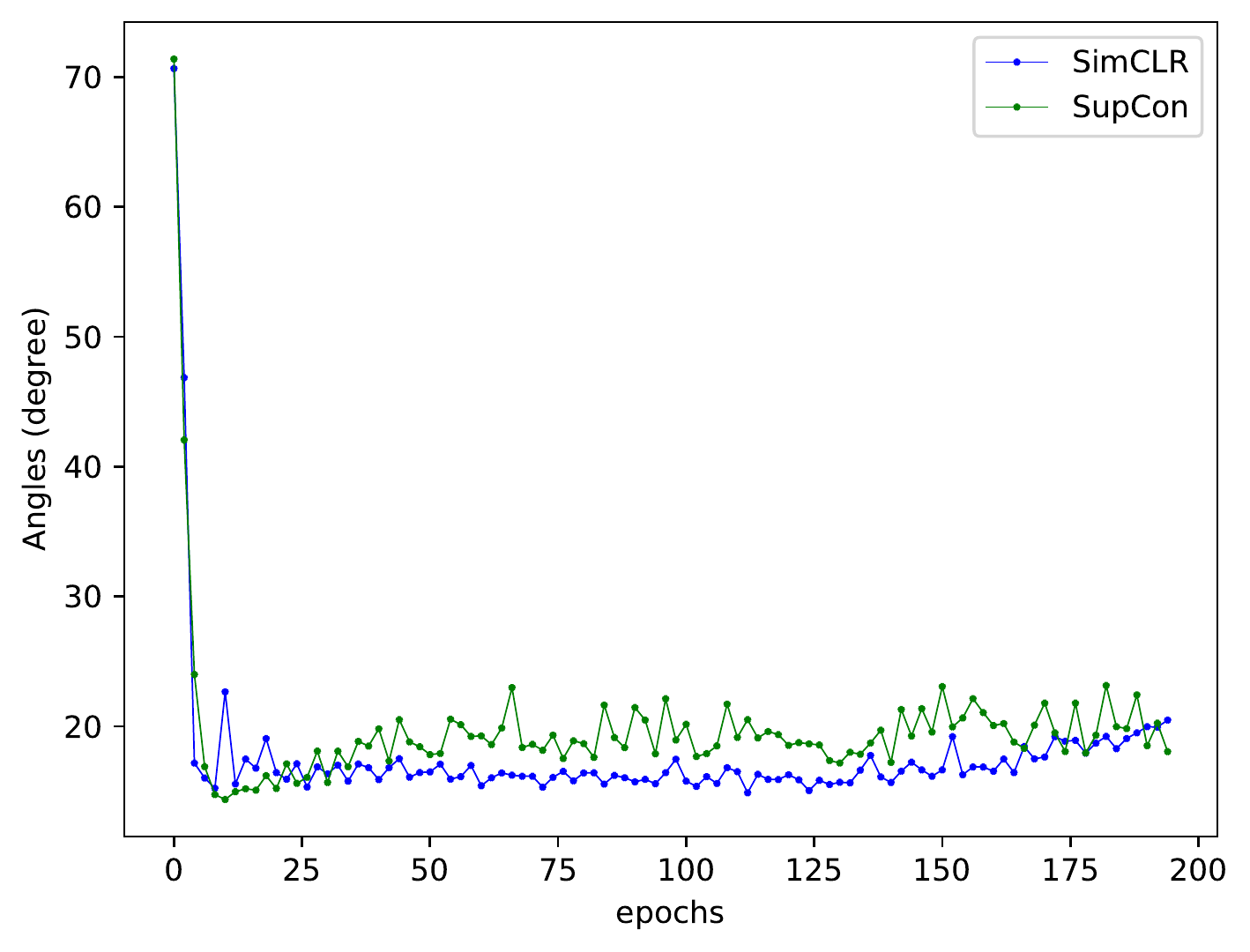}}\\
\subfloat[Unsup. vs. Sup.]{\includegraphics[width=0.33\textwidth]{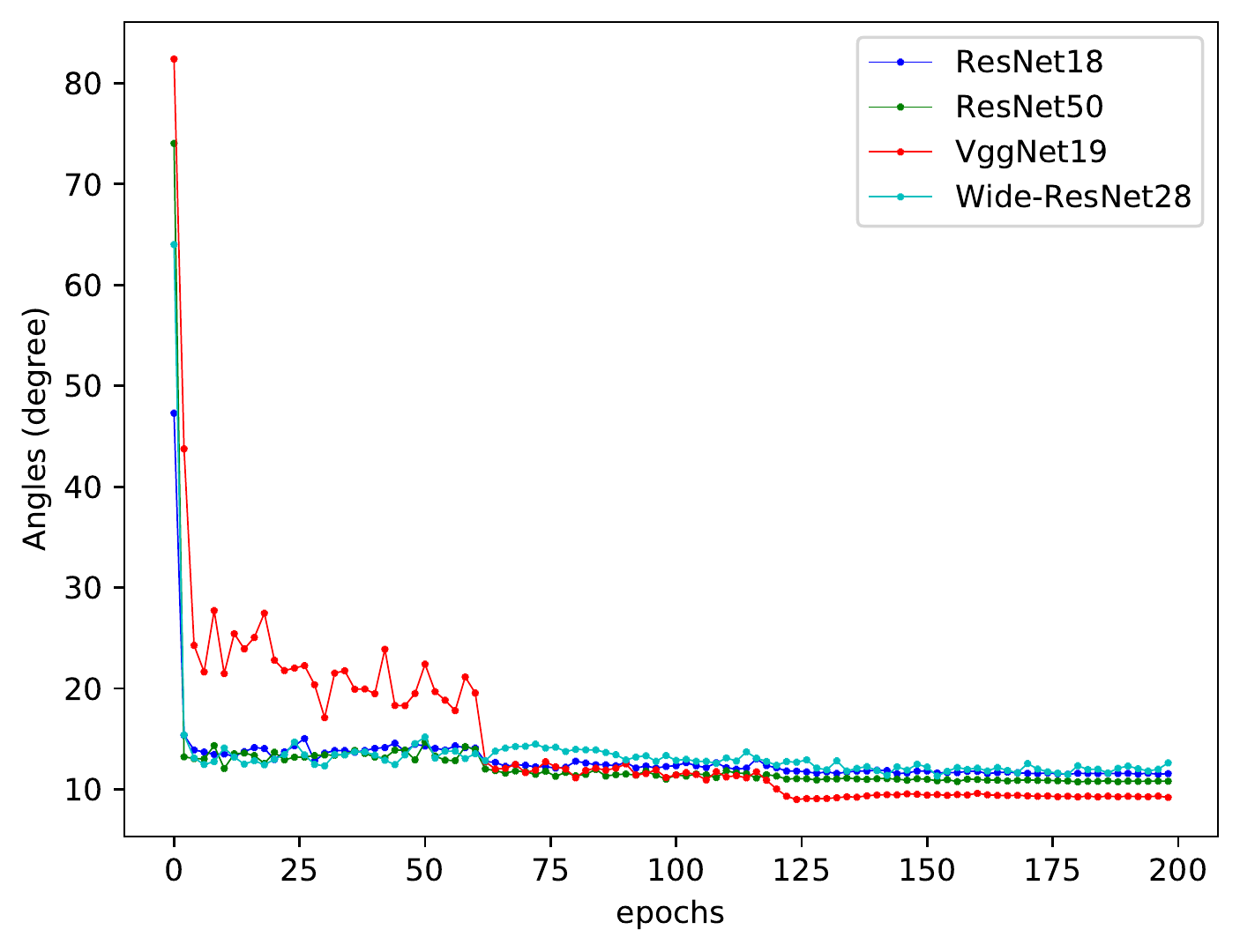}}
\subfloat[Unsup. vs. Unsup.]{\includegraphics[width=0.33\textwidth]{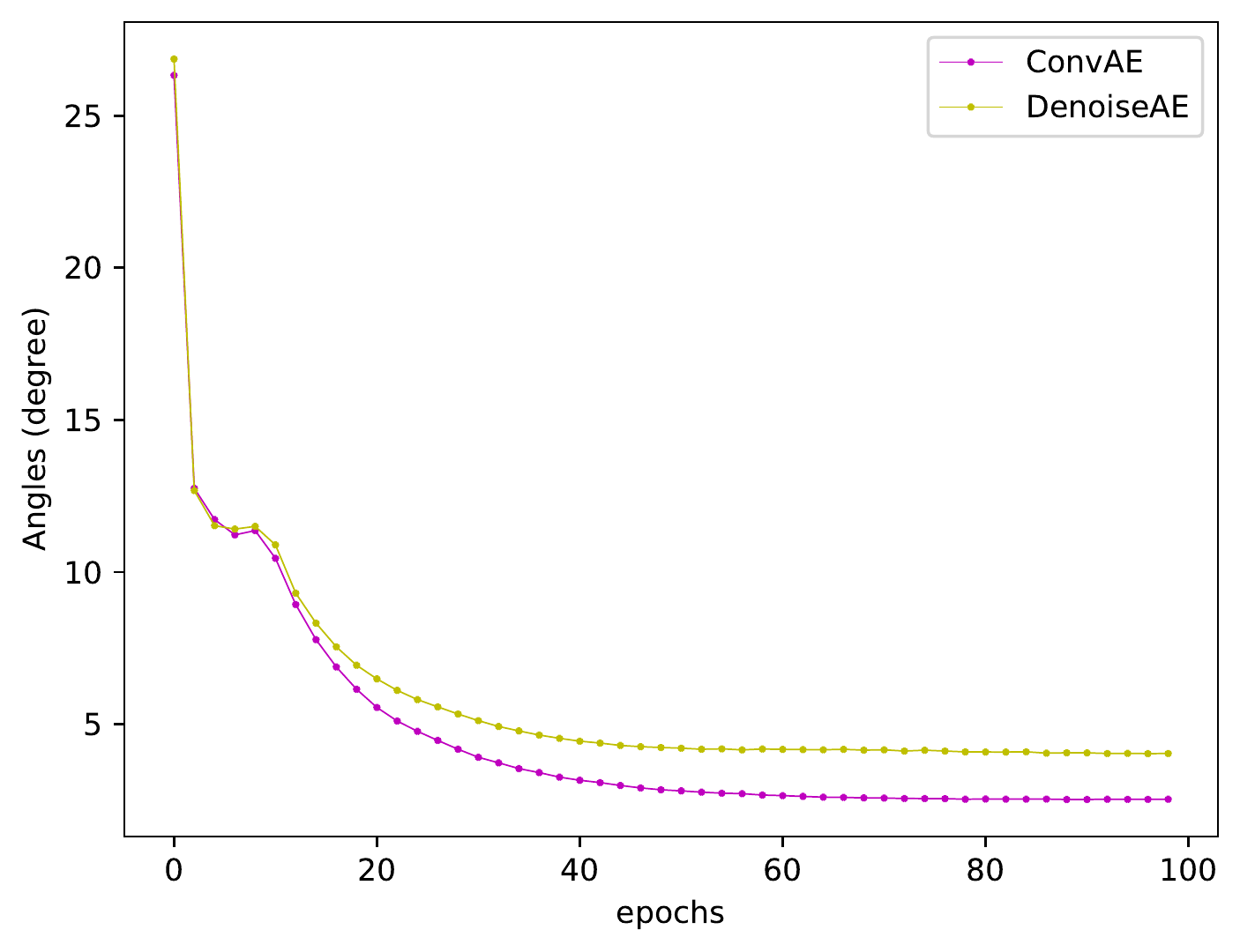}}
\subfloat[Unsup vs Self-Sup]{\includegraphics[width=0.33\textwidth]{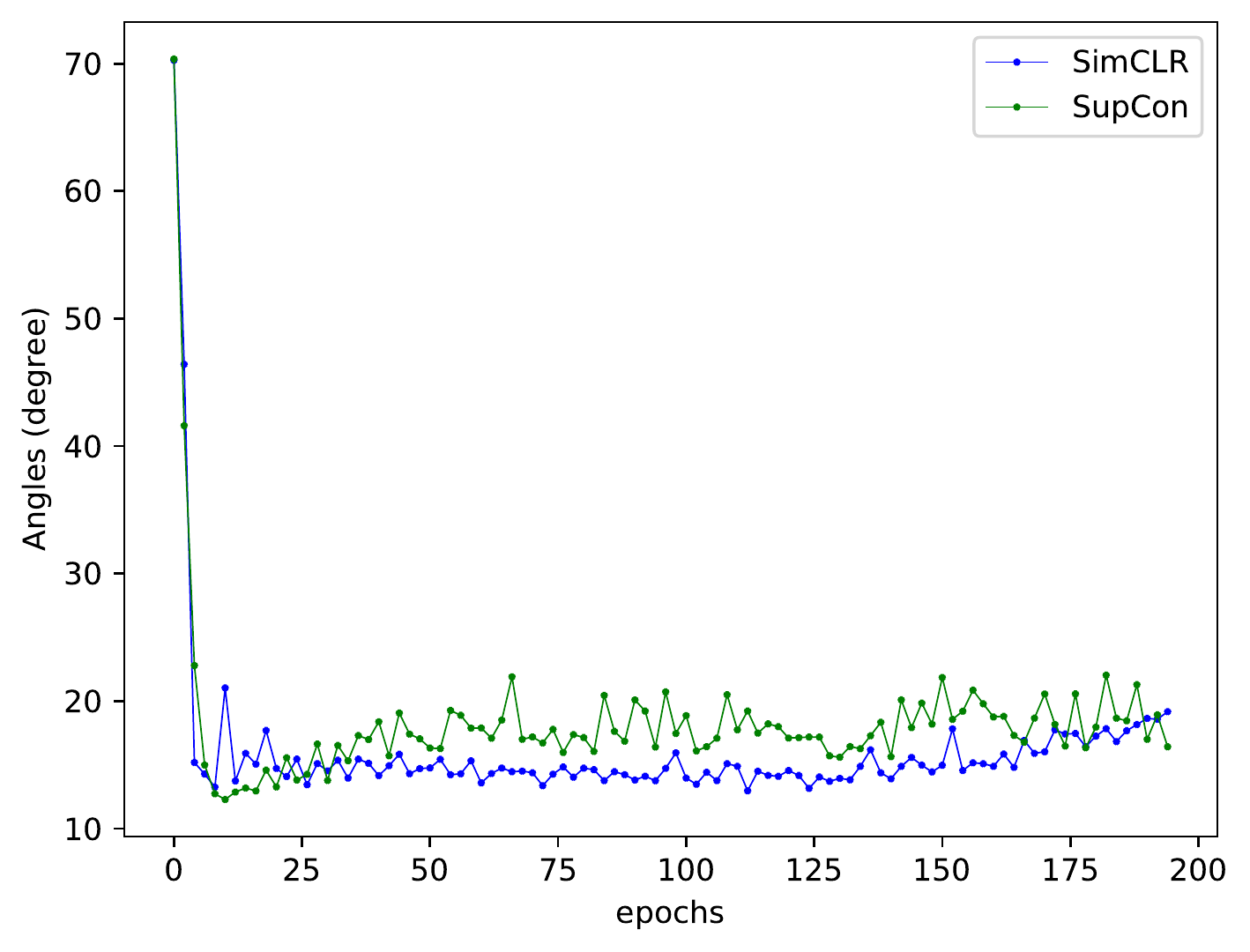}}\\
\subfloat[Self. vs Sup.]{\includegraphics[width=0.33\textwidth]{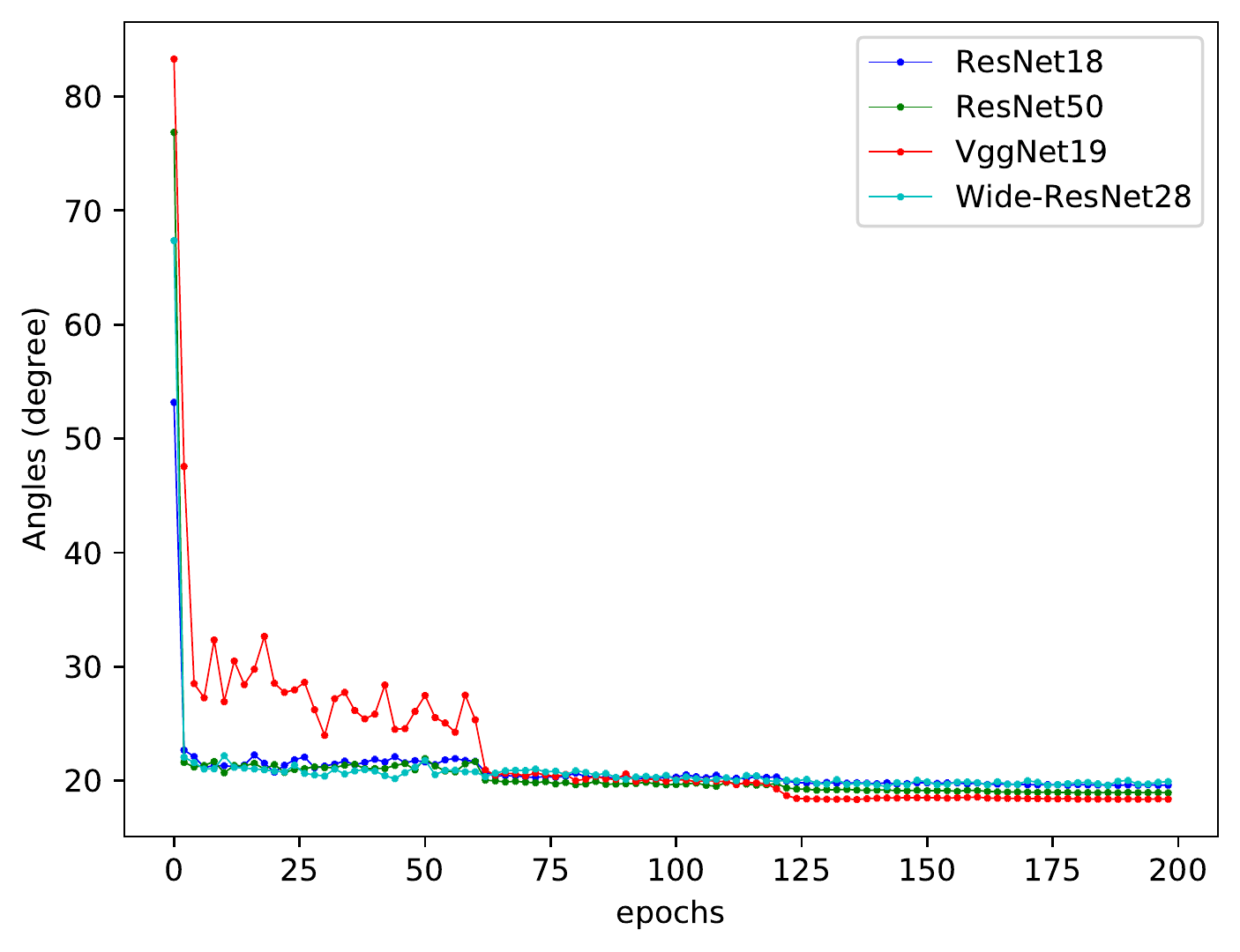}}
\subfloat[Self. vs Unsup.]{\includegraphics[width=0.33\textwidth]{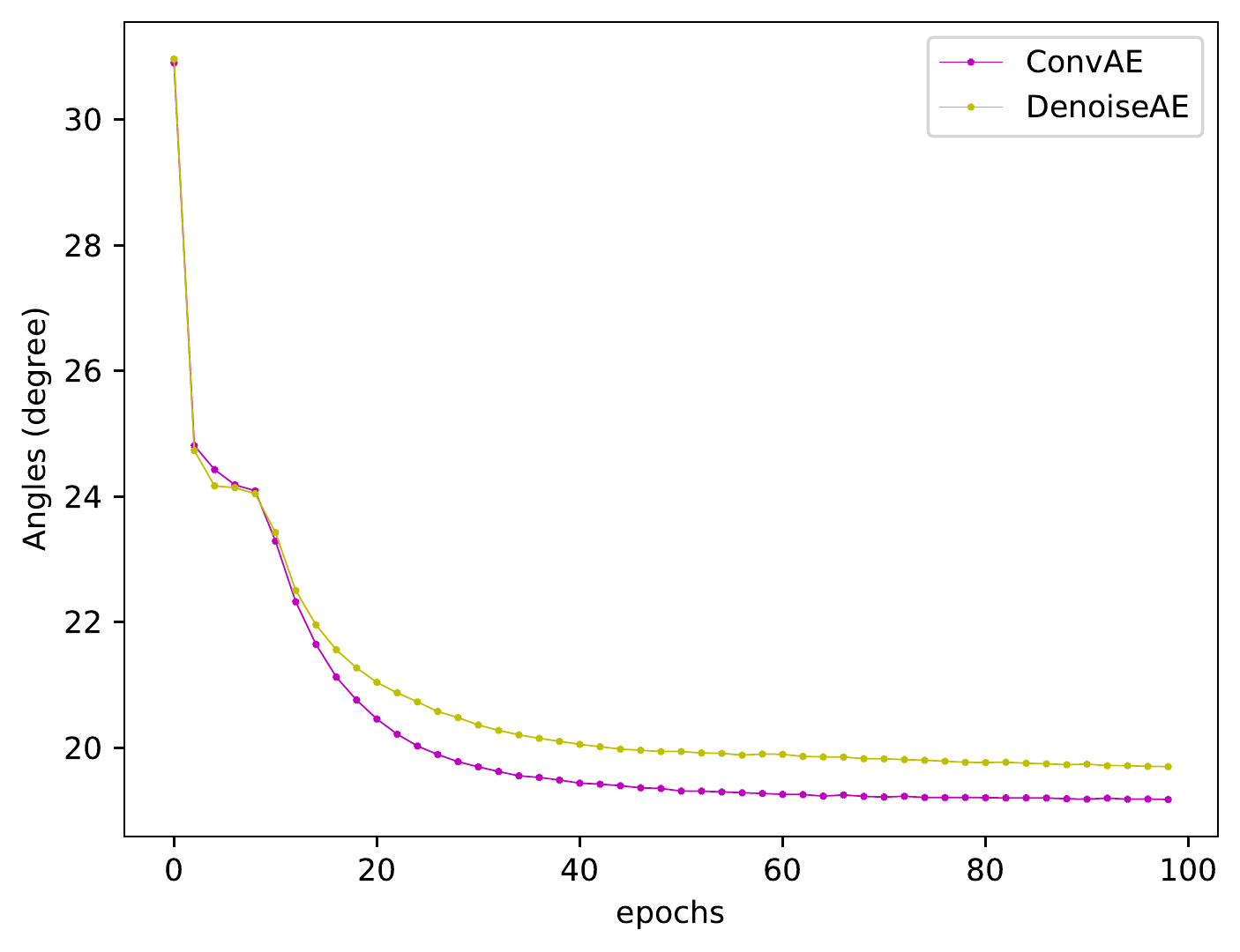}}
\subfloat[Self. vs Self-Sup.]{\includegraphics[width=0.33\textwidth]{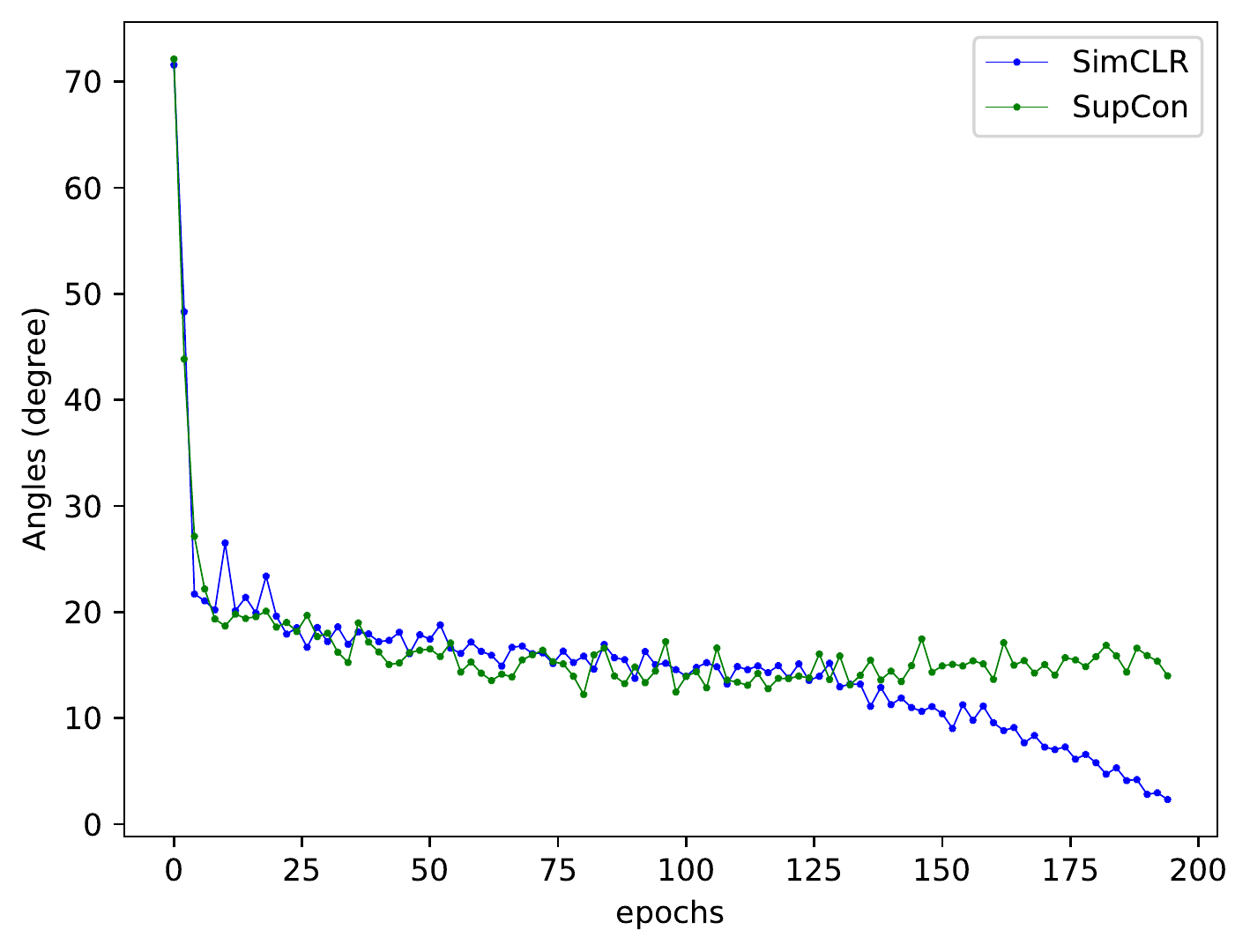}}
\caption{{Convergence of $\mathcal{P}$-vector angles between checkpoint per epoch and well-trained models using CIFAR-100.} }
\label{fig:claim2_2_cifar100}
\end{figure*}

We also test and verify the model-to-model common subspace shared by models trained with different supervisory manners on CIFAR-100 dataset. Experiments carried out to evaluate the angles between $\mathcal{P}$-vectors for checkpoints of all models and $\mathcal{P}$-vectors for well-trained supervised, unsupervised and self-supervised models, where we use the well-trained Wide-ResNet28/Convolution Auto-encoder and SimCLR model (trained with 200 epochs under suggest settings) as the reference of supervised, unsupervised and self-supervised models, respectively. As shown in Fig.\ref{fig:claim2_2_cifar100}, a consistent convergence for the curves of the angles can be observed and support our hypothesis that the dynamics learning procedure construct the common subspace gradually. 

\subsection{The Non-monotonic Trend in the First Epoch of Comparison of Angles between Model and Raw Data $\mathcal{P}$-vectors}
\begin{figure*}[hbtp]
\centering
\subfloat[Supervised CIFAR-10]{\includegraphics[width=0.5\textwidth]{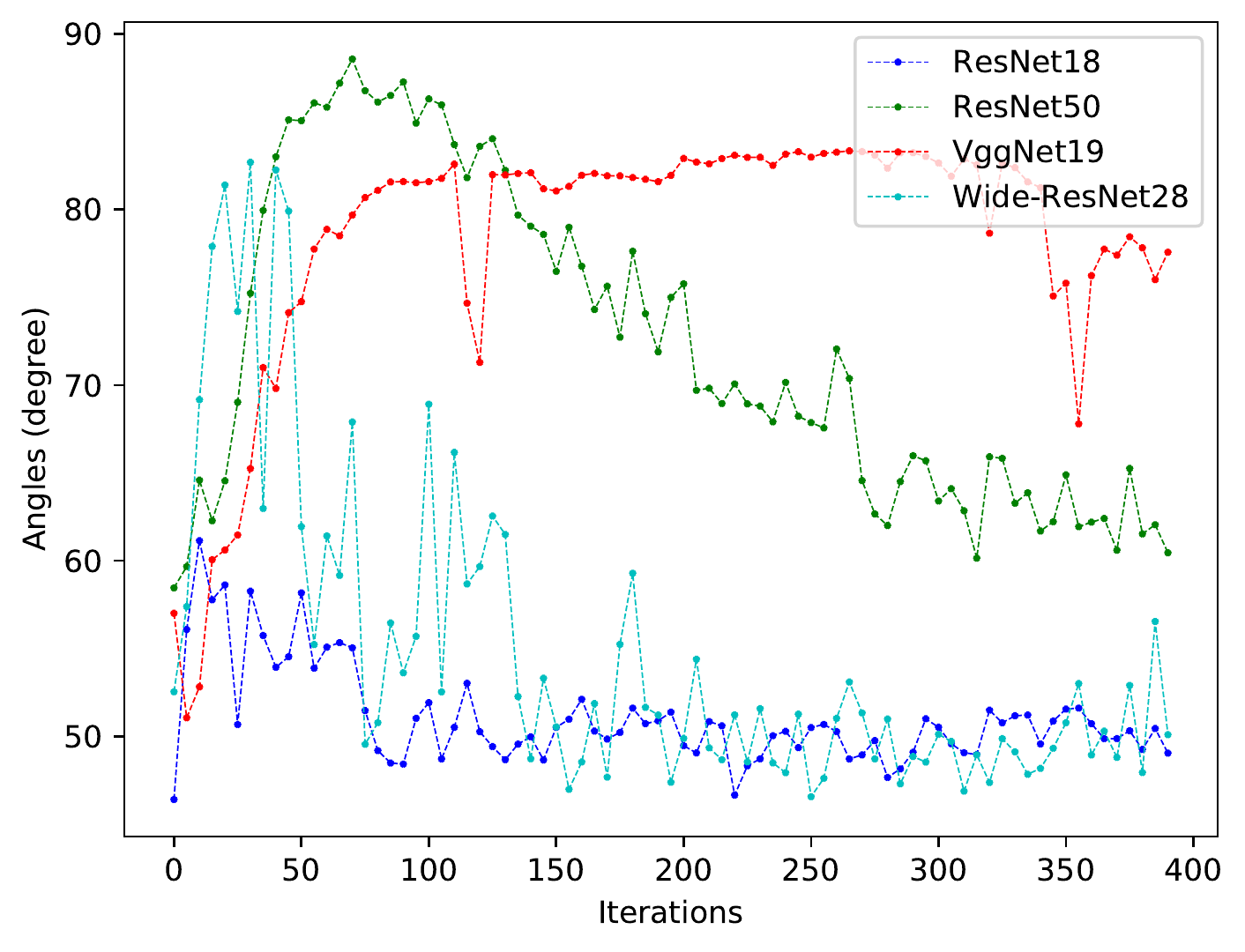}}
\subfloat[Supervised CIFAR-100]{\includegraphics[width=0.5\textwidth]{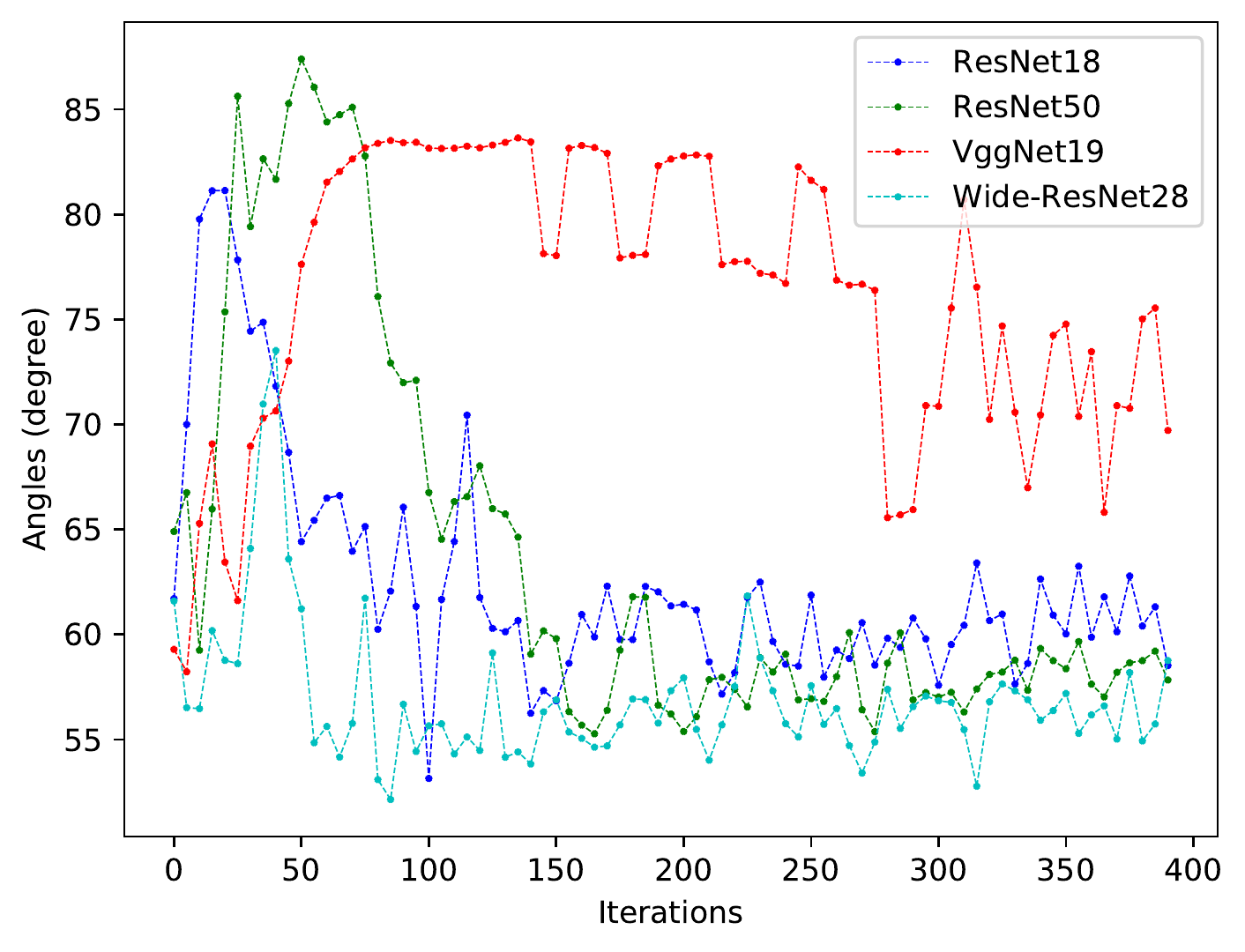}}\\

\subfloat[Unsupervised CIFAR-10]{\includegraphics[width=0.5\textwidth]{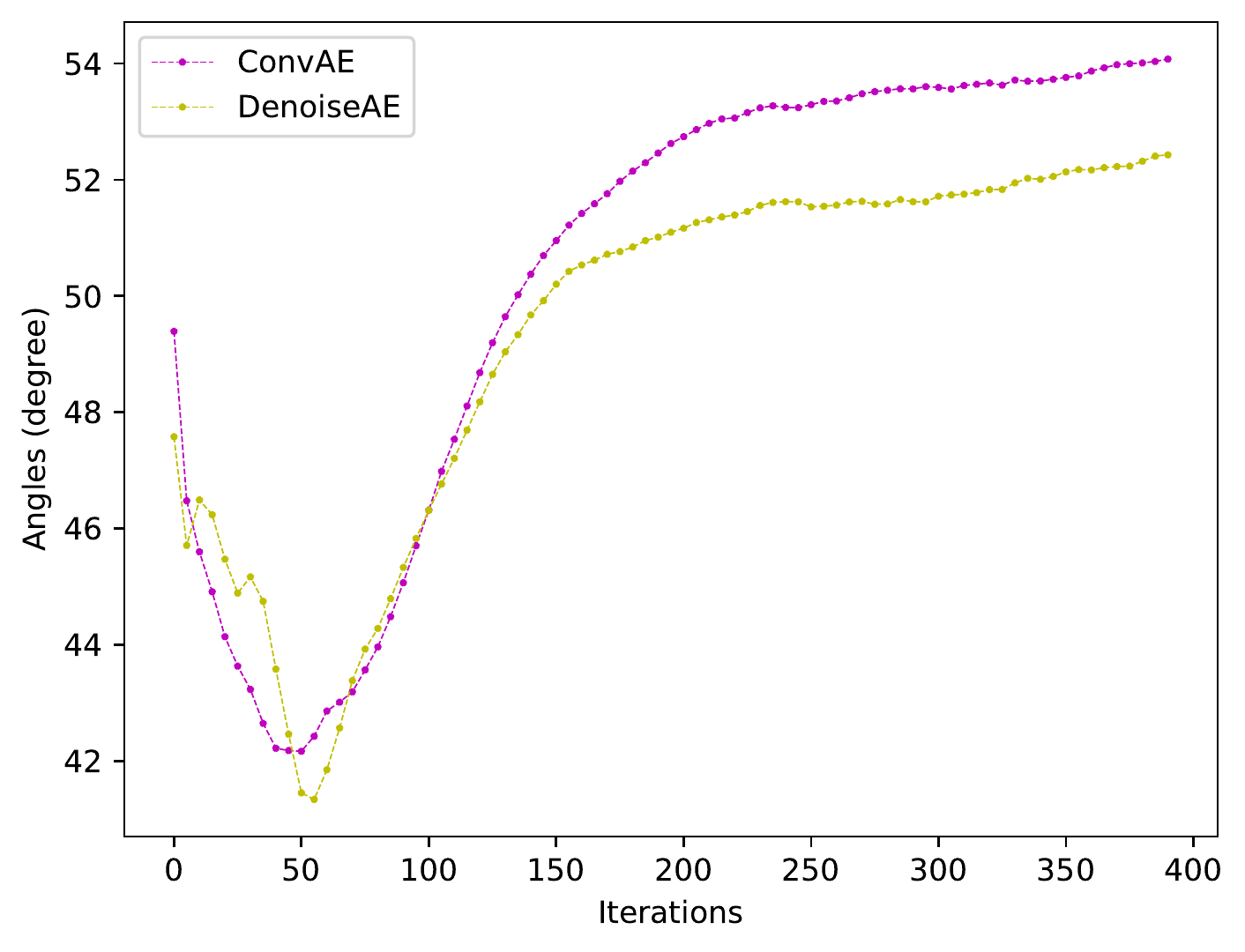}}
\subfloat[Unsupervised CIFAR-100]{\includegraphics[width=0.5\textwidth]{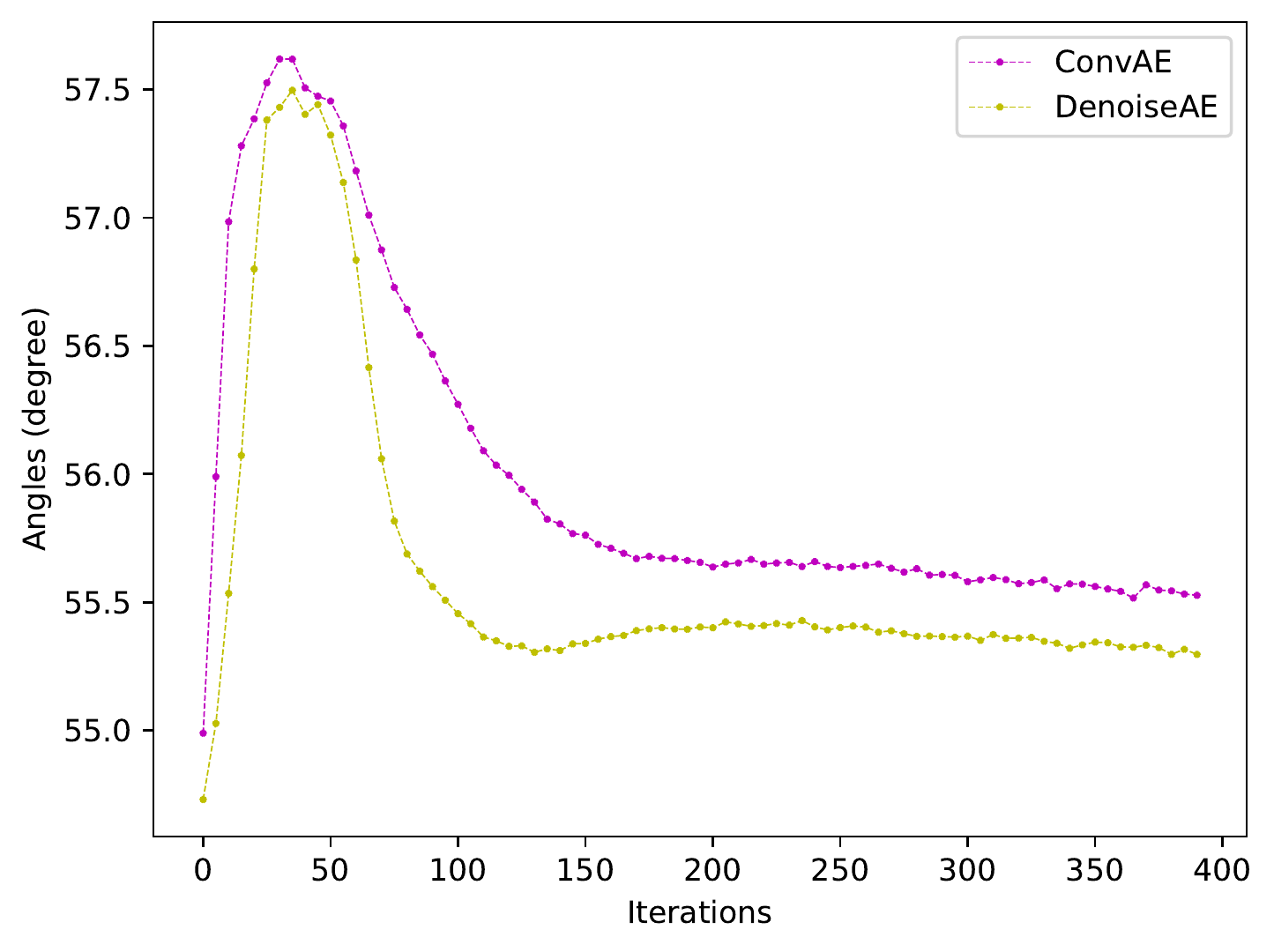}}\\

\subfloat[Self-Sup CIFAR-10.]{\includegraphics[width=0.5\textwidth]{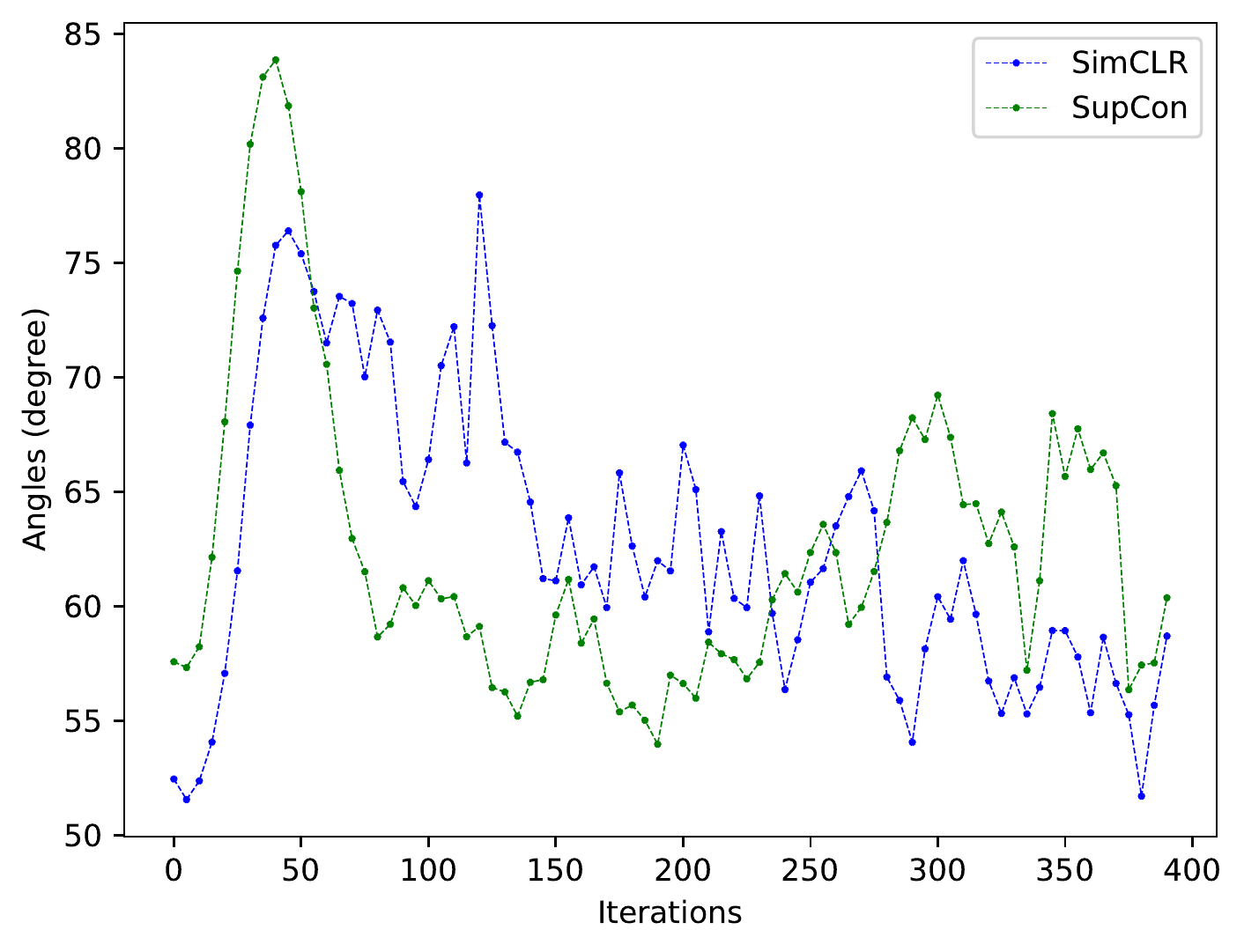}}
\subfloat[Self-Sup CIFAR-100.]{\includegraphics[width=0.5\textwidth]{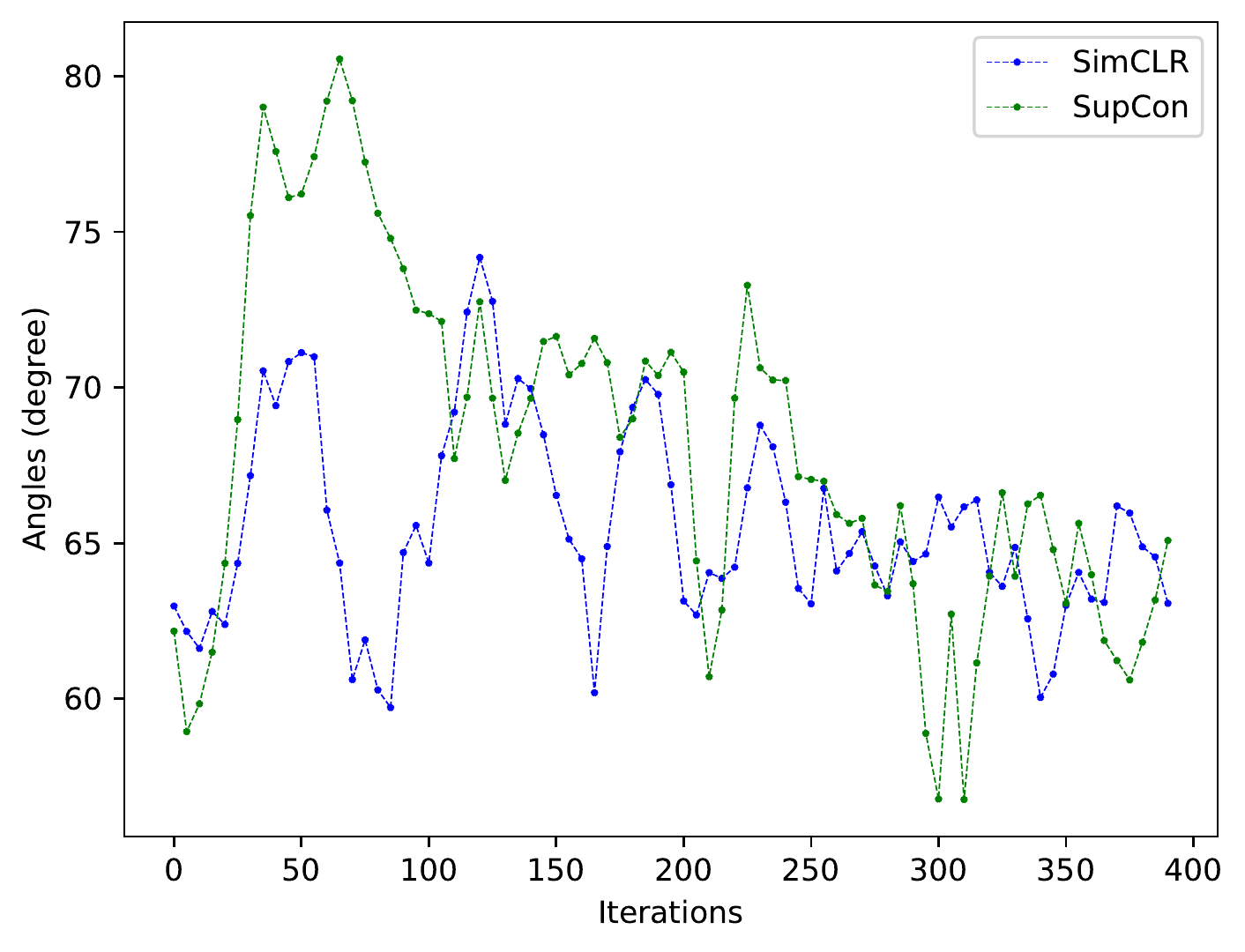}}
\caption{{Angles between the $\mathcal{P}$-vectors of the training model and the raw datasets over the number of iterations in the first epoch using CIFAR-10 /CIFAR-100.} }
\label{fig:epoch0_cifar10100}
\end{figure*}

We also explore the construction procedure for the common subspace share between feature vectors and the raw data during the training process.  Experiments are carried out to compare the space of models and raw data $\mathcal{P}$-vectors on the training dataset. As shown in Fig.\ref{fig:epoch0_cifar10100}, we observe a non-monotonic trend that the angle first rises with the random initialization and drop down. The angles keeps the approximately monotonically decreasing and converging to small values in following training epochs. The experiments shows consistent result on both CIFAR-10 and 100 dataset. Note that we follow the default random data augmentation policy to pre-process the training dataset.


\subsection{Case Studies on the Angle Variety Between Model and Raw Data for Each Layer}
\begin{figure*}
\centering
\subfloat[Epoch 5.]{\includegraphics[width=0.193\textwidth]{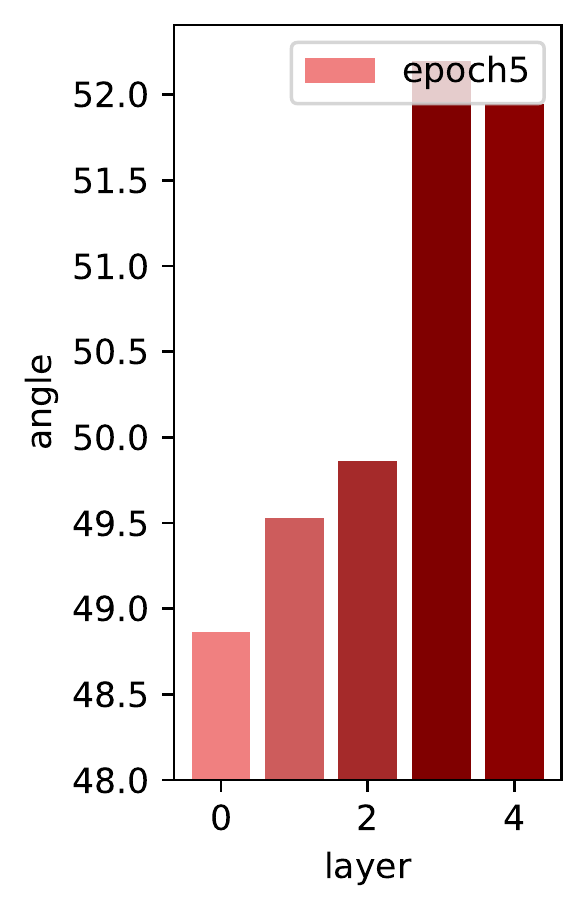}}
\subfloat[Epoch45]{\includegraphics[width=0.2\textwidth]{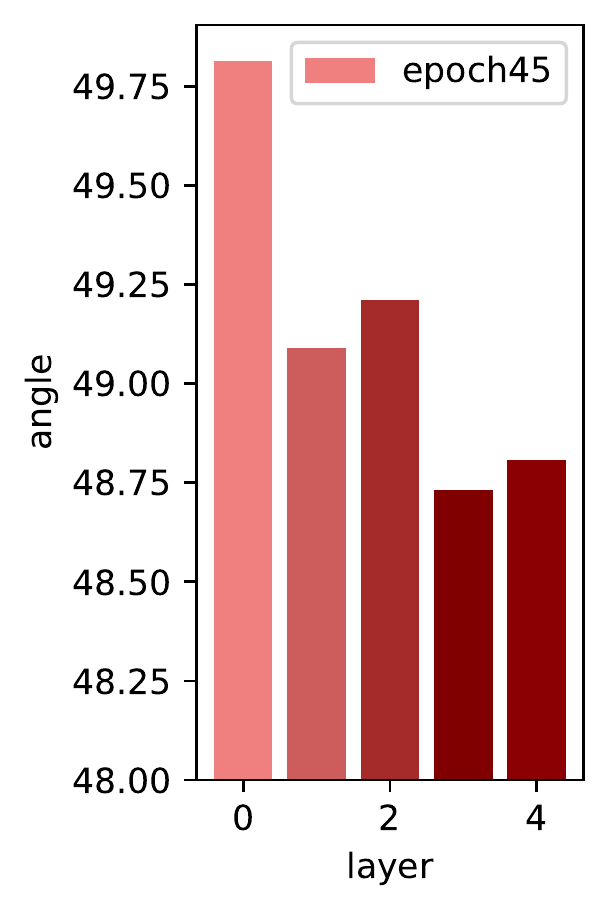}}
\subfloat[Epoch85.]{\includegraphics[width=0.193\textwidth]{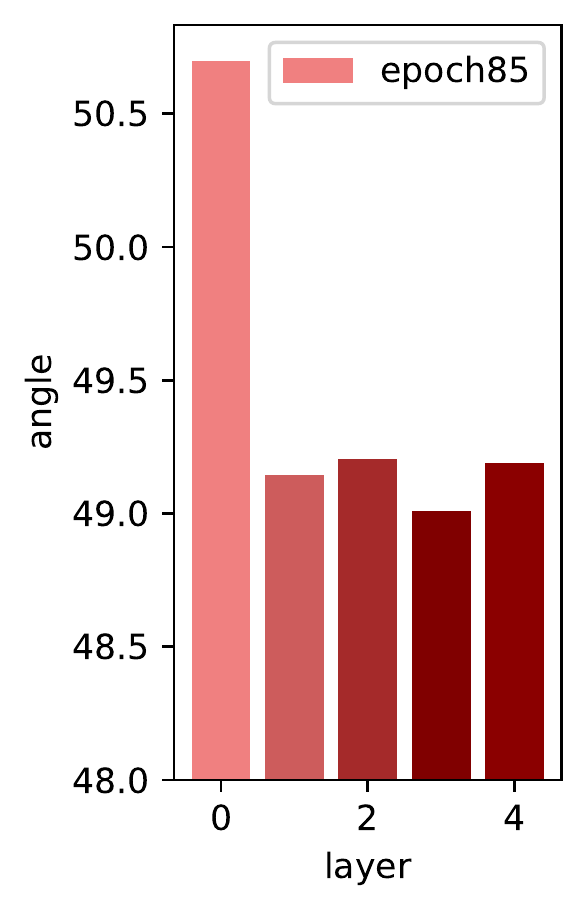}}
\subfloat[Epoch125.]{\includegraphics[width=0.2\textwidth]{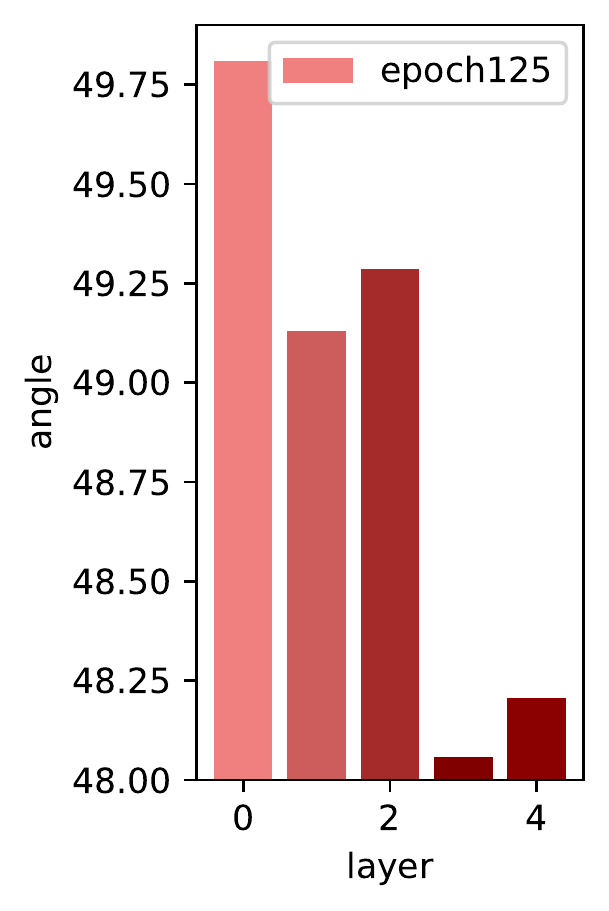}}
\subfloat[Epoch165.]{\includegraphics[width=0.2\textwidth]{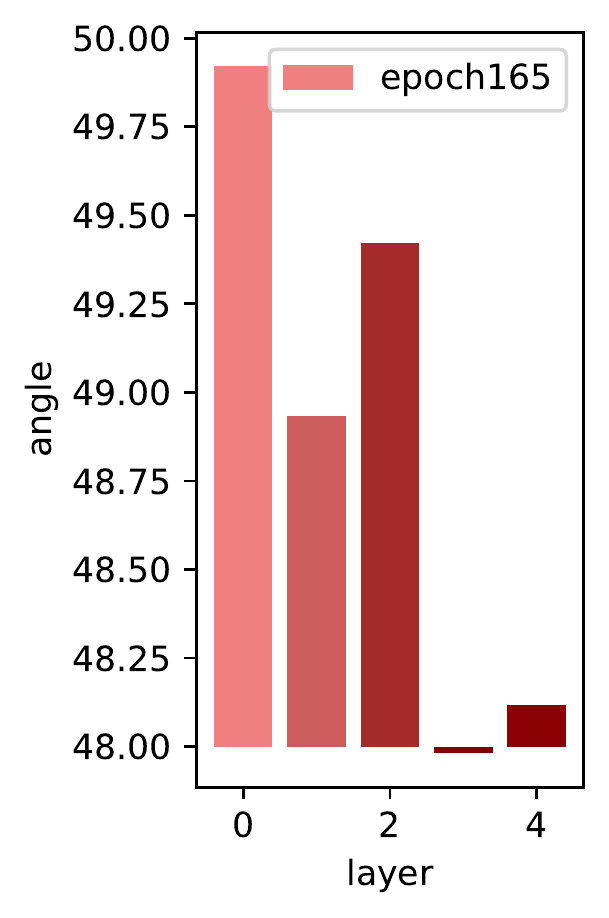}}
\caption{{Angles changes through layer between the $\mathcal{P}$-vectors of the training and raw data over the number of iterations in the first epoch using CIFAR-10.} }
\label{fig:perlayer_cifar10}
\end{figure*}

To explore the dynamic variation process by zooming in to every layer variation in each epoch, we perform the case study using Resnet-18 structure on CIFAR-10 dataset. As shown in Fig.\ref{fig:perlayer_cifar10}, we give the angles according to layers of 5 different epochs, where the x-axis indicating the indices of residual blocks in the network structure and y-axis refers to the angles between the model checkpoint and raw data $\mathcal{P}$-vectors. We observed that in early training stage, the angles between the model checkpoint and raw data $\mathcal{P}$-vectors keeps an increasing manner when the features passing through layers and turn into a decrease trend towards the stacked layers in the late training stage. This set of experiments further support our hypothesis that the dynamics learning procedure construct the common subspace gradually through training process.

\subsection{Distribution of Values in the $\mathcal{P}$-vector}
\begin{figure*}
\subfloat[CIFAR 10 Epoch 0]{\includegraphics[width=0.33\textwidth]{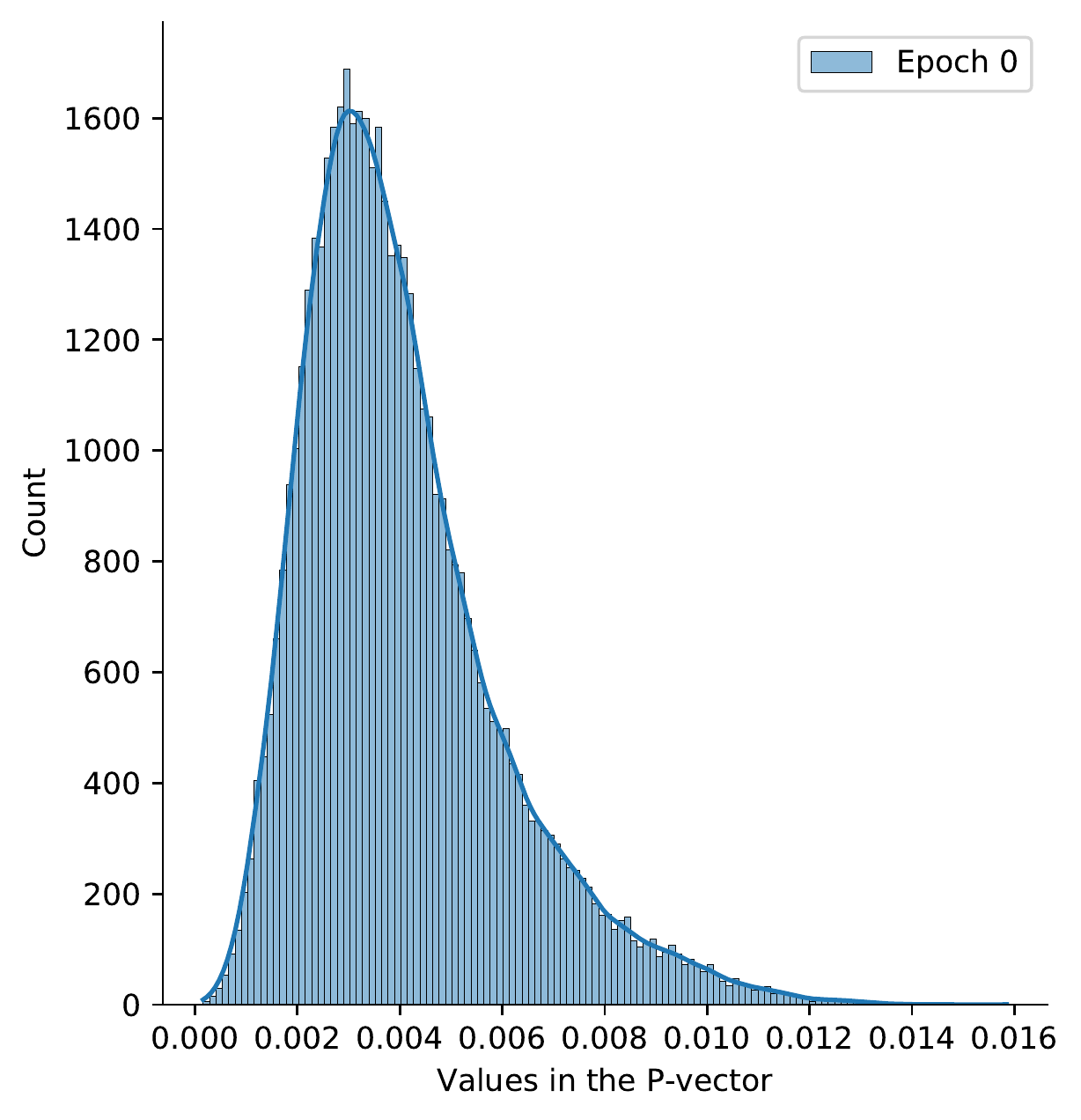}}
\subfloat[CIFAR 10 Epoch 60]{\includegraphics[width=0.33\textwidth]{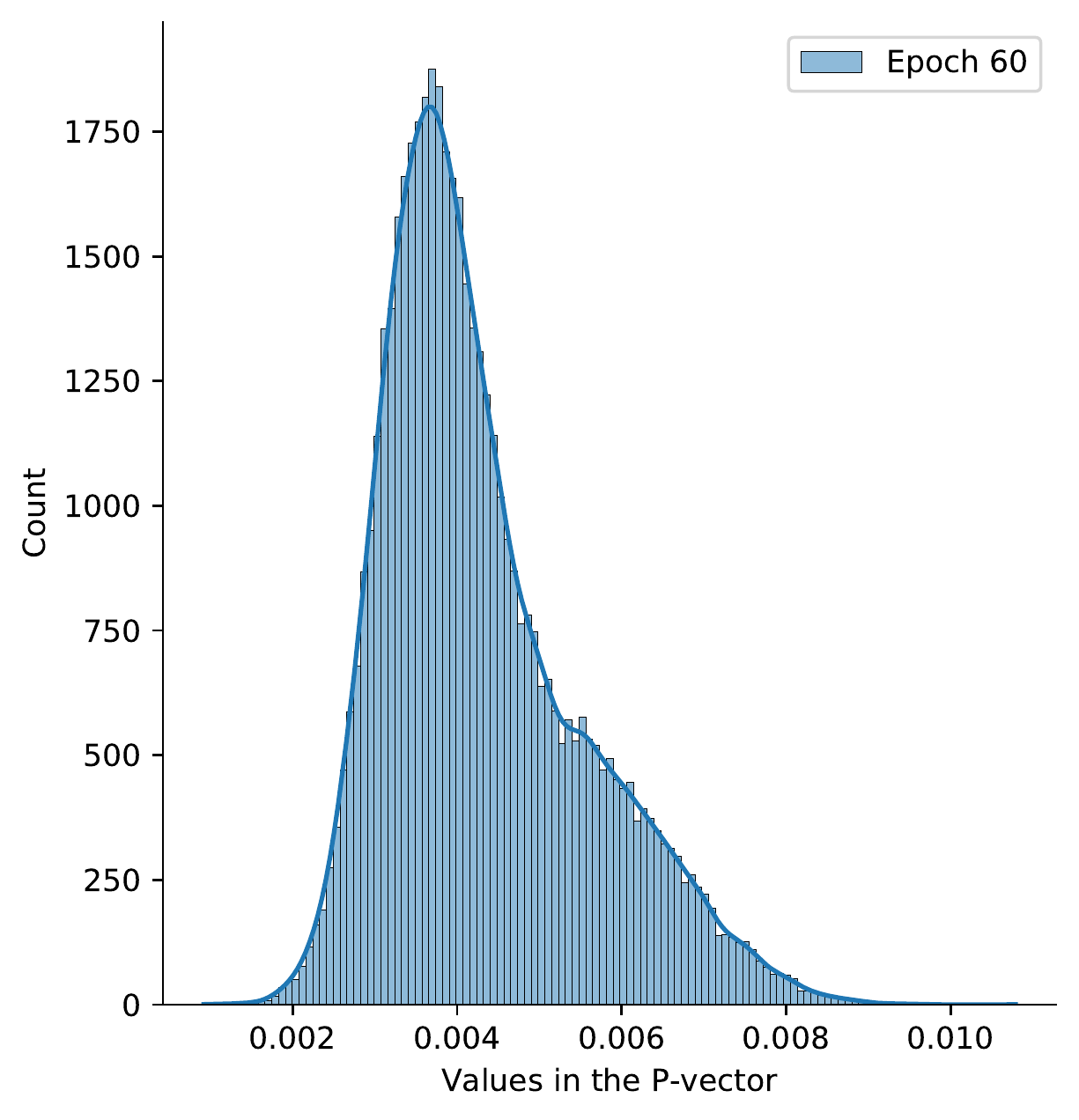}}
\subfloat[CIFAR 10 Epoch 120]{\includegraphics[width=0.33\textwidth]{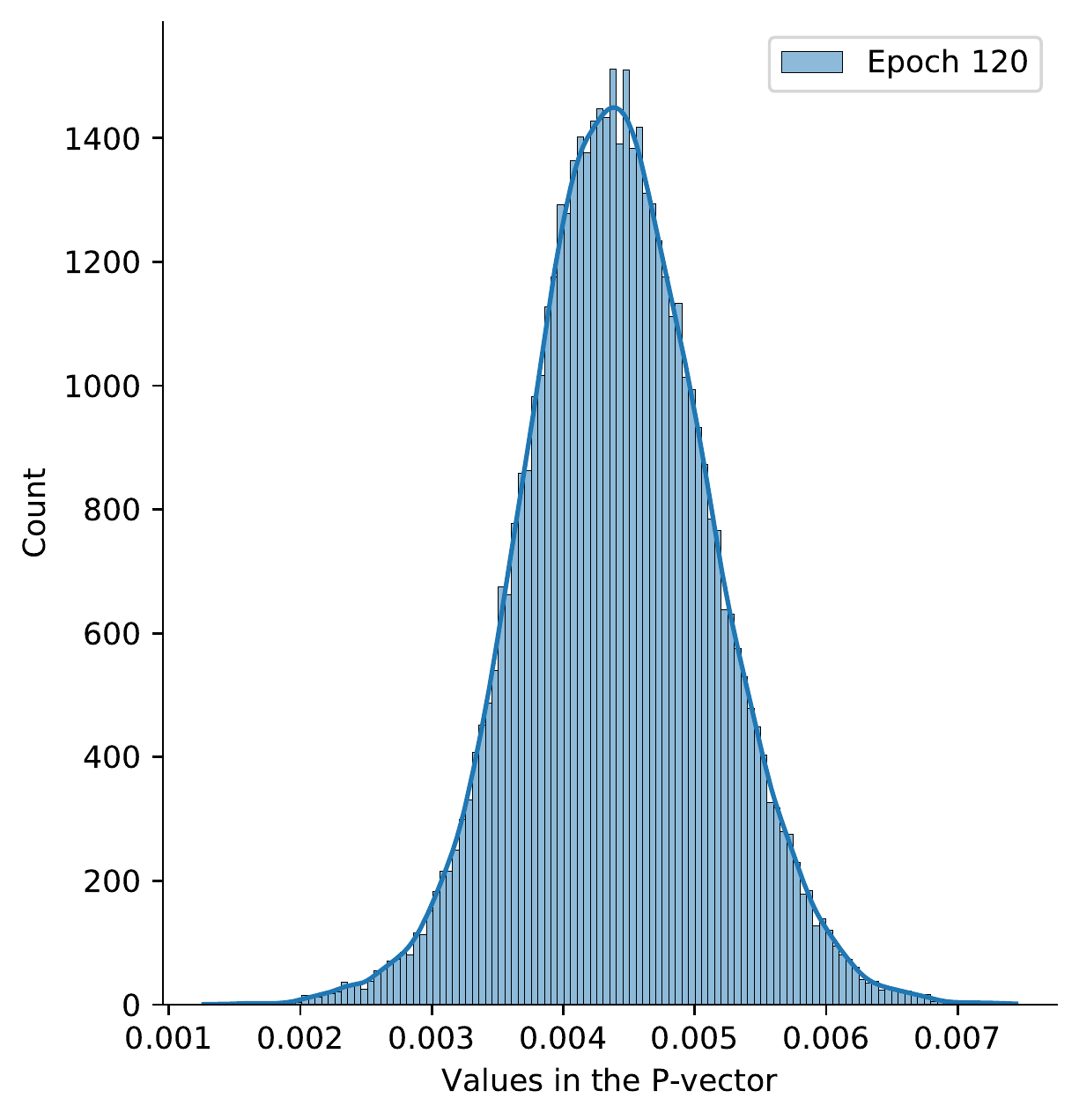}}\\
\subfloat[CIFAR 10 Epoch 160]{\includegraphics[width=0.33\textwidth]{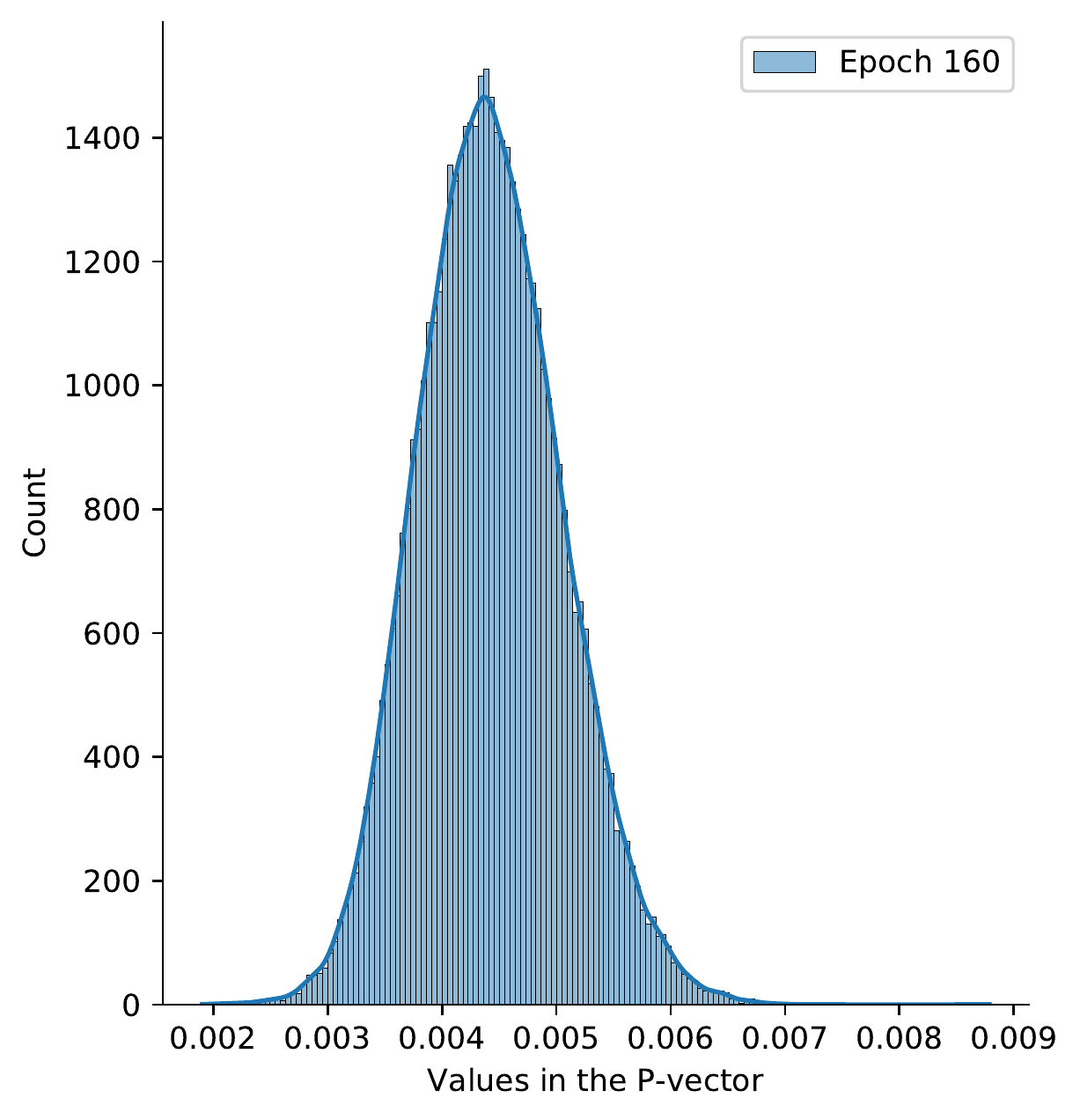}}
\subfloat[CIFAR 10 Epoch 200]{\includegraphics[width=0.33\textwidth]{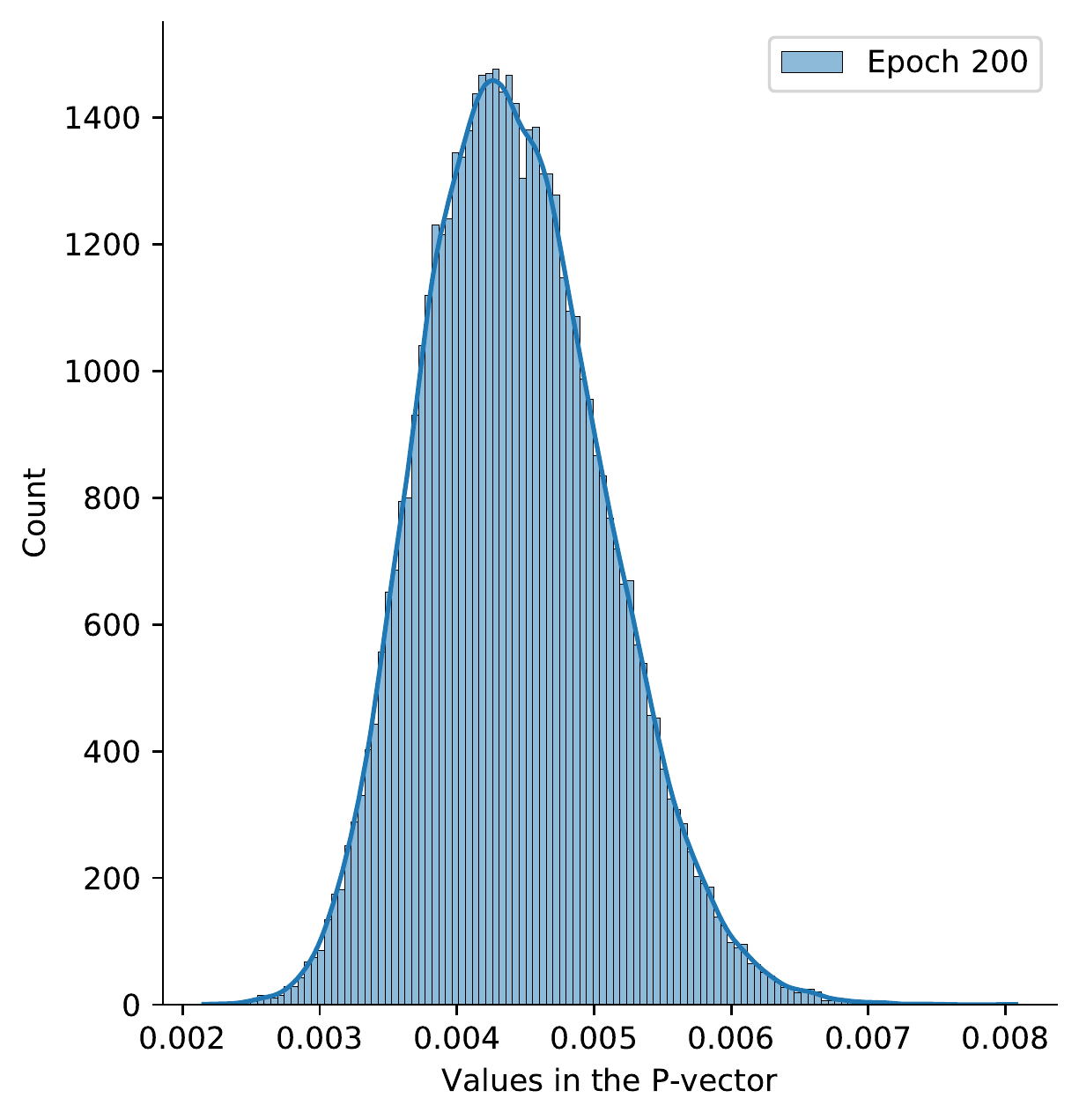}}\\
\subfloat[CIFAR 100 Epoch 0]{\includegraphics[width=0.33\textwidth]{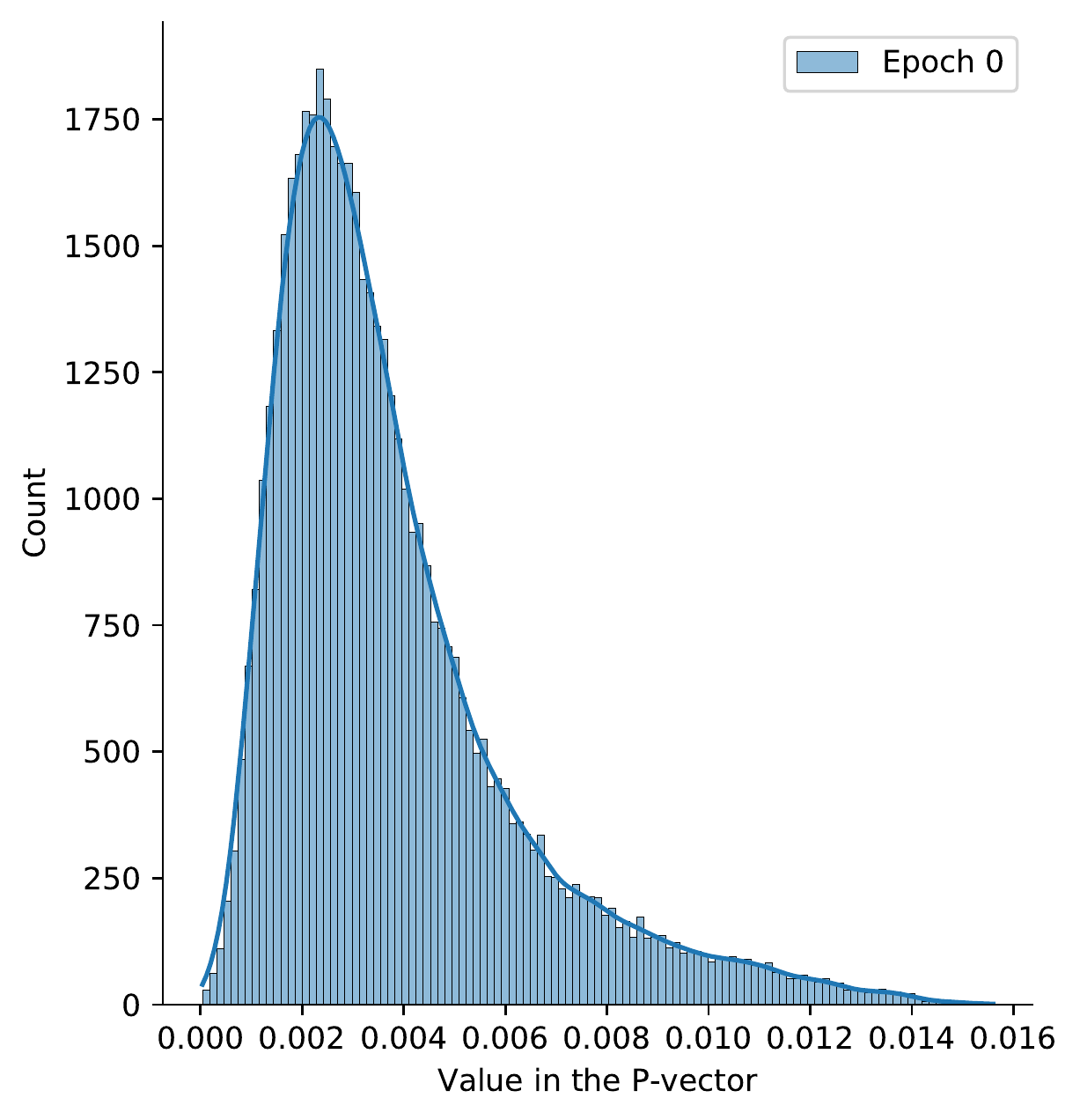}}
\subfloat[CIFAR 100 Epoch 60]{\includegraphics[width=0.33\textwidth]{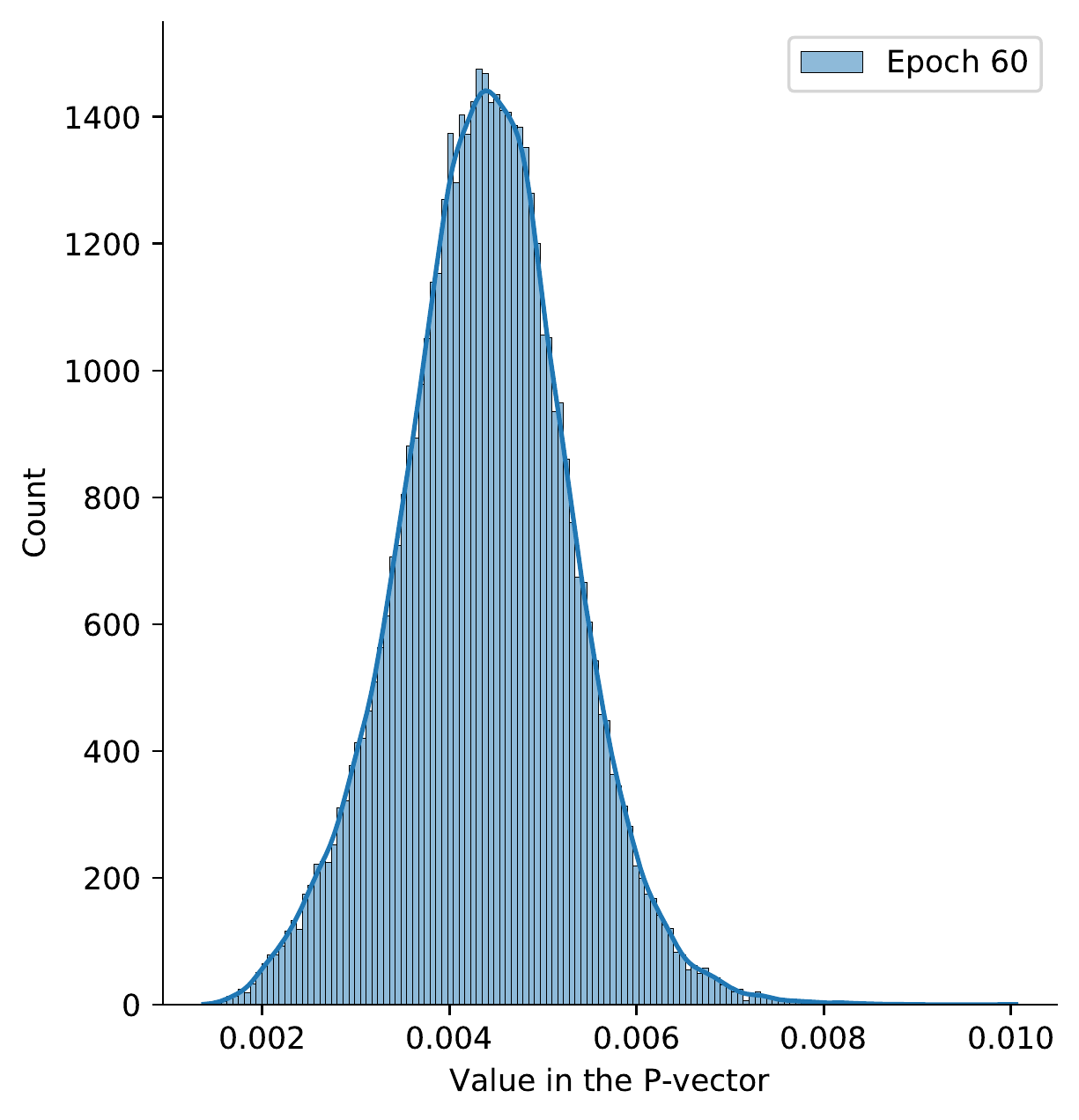}}
\subfloat[CIFAR 100 Epoch 120]{\includegraphics[width=0.33\textwidth]{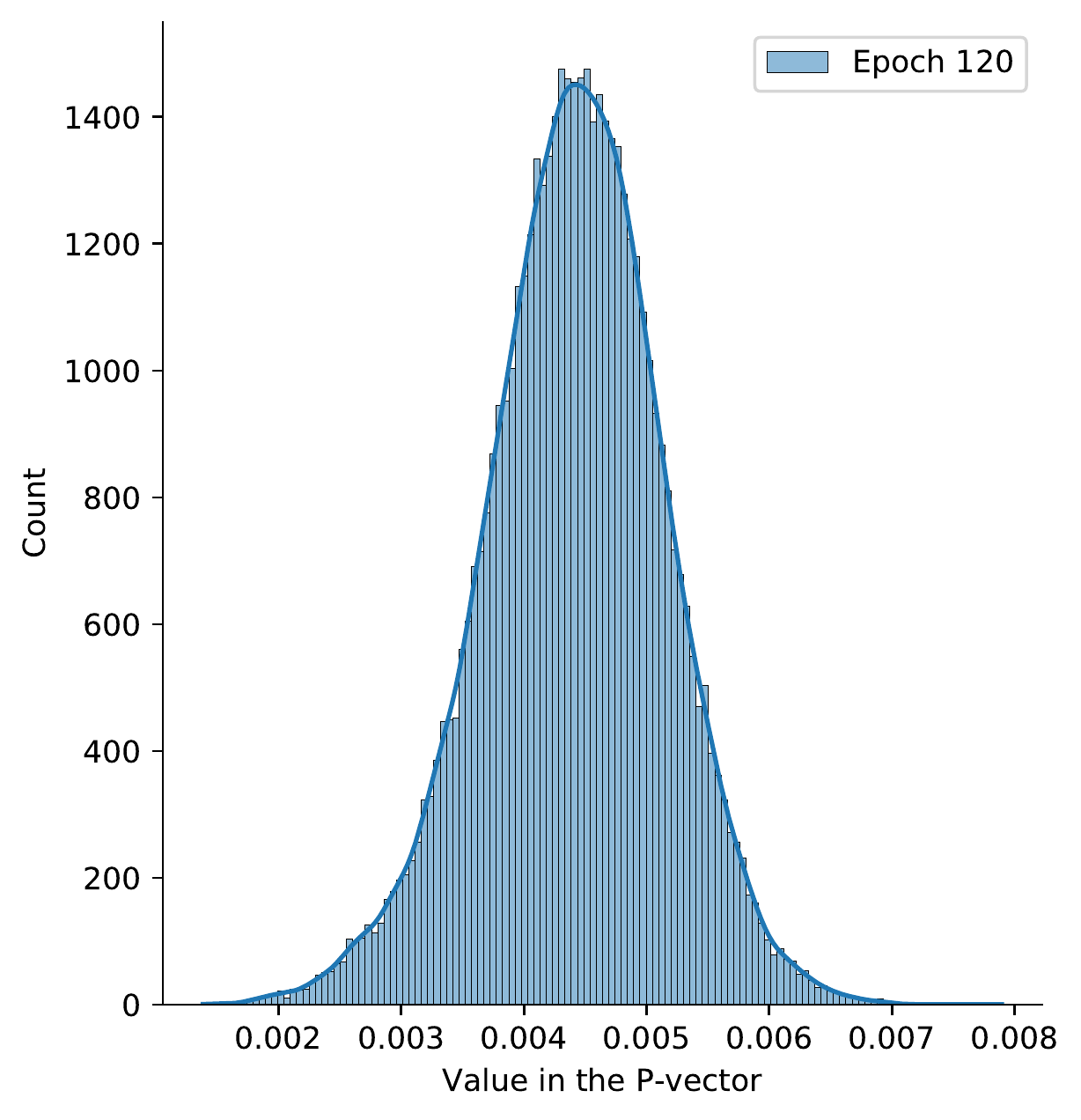}}\\
\subfloat[CIFAR 100 Epoch 160]{\includegraphics[width=0.33\textwidth]{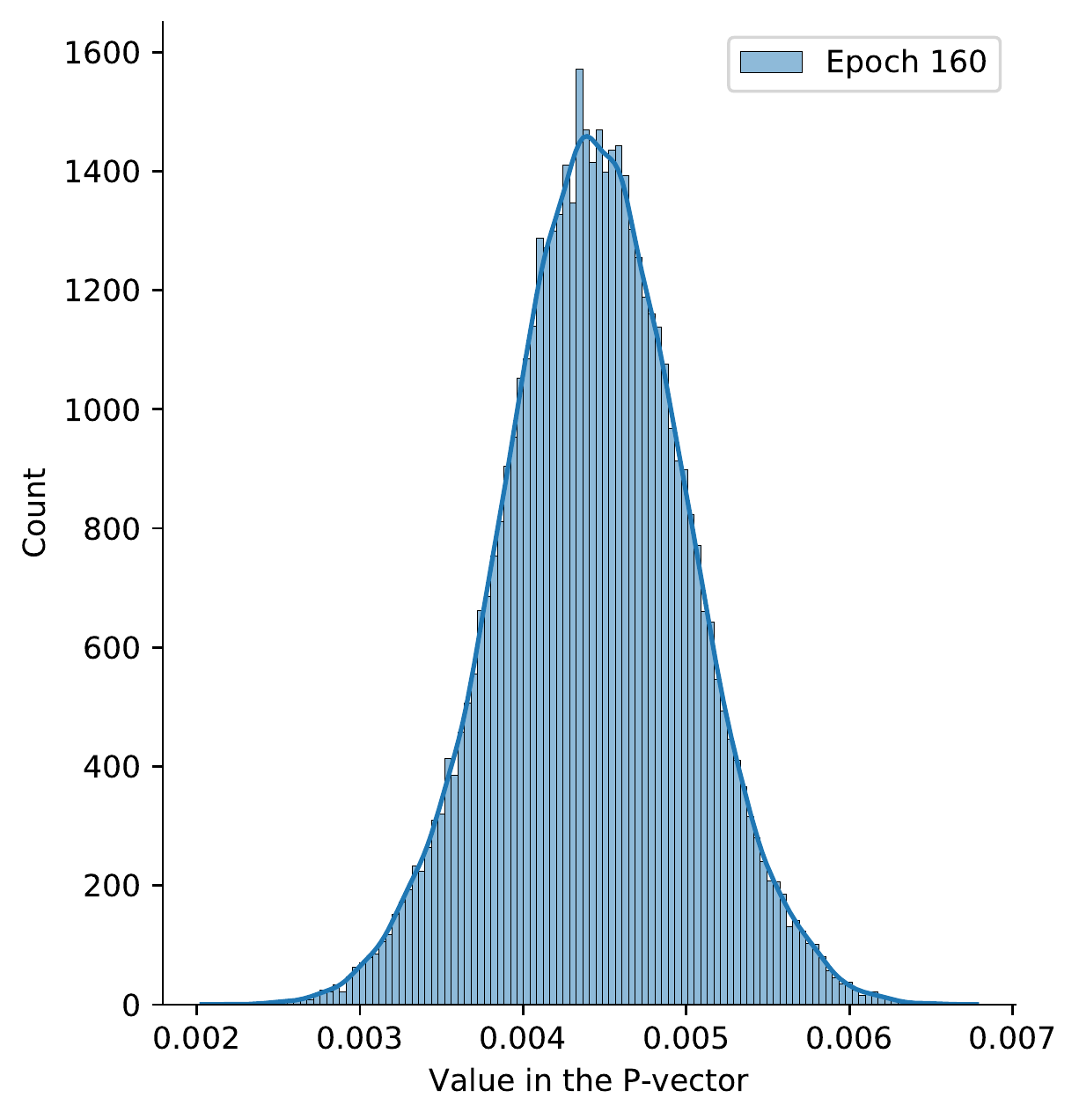}}
\subfloat[CIFAR 100 Epoch 200]{\includegraphics[width=0.33\textwidth]{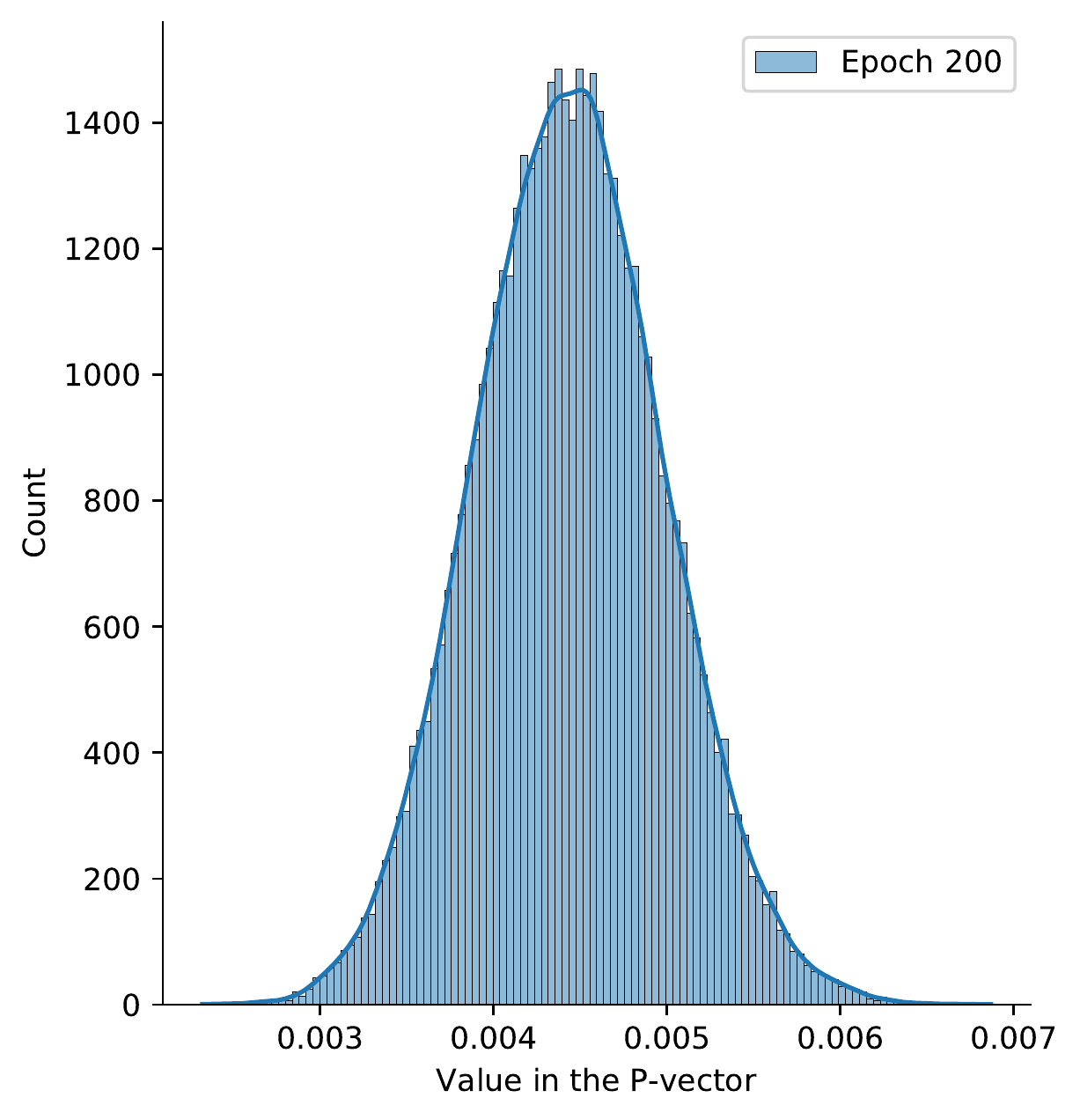}}
\caption{{Frequency of Values appeared in the P-vector versus training epochs.} }
\label{fig:Frequency}
\end{figure*}

\begin{figure*}
\subfloat[CIFAR 100 Epoch 0,1]{\includegraphics[width=0.9\textwidth]{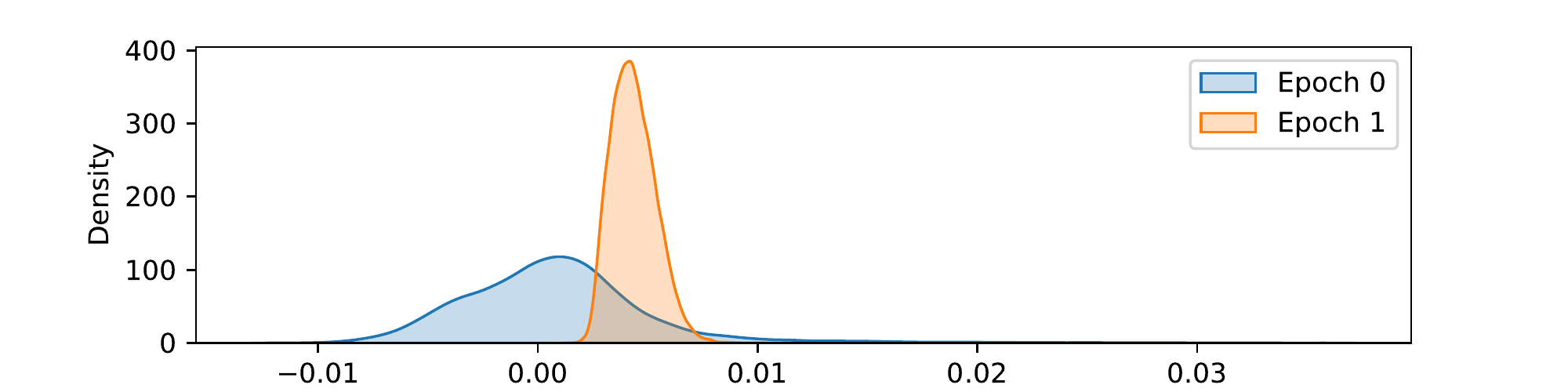}}\\
\subfloat[CIFAR 100 Epoch 1,60,120,160,200]{\includegraphics[width=0.9\textwidth]{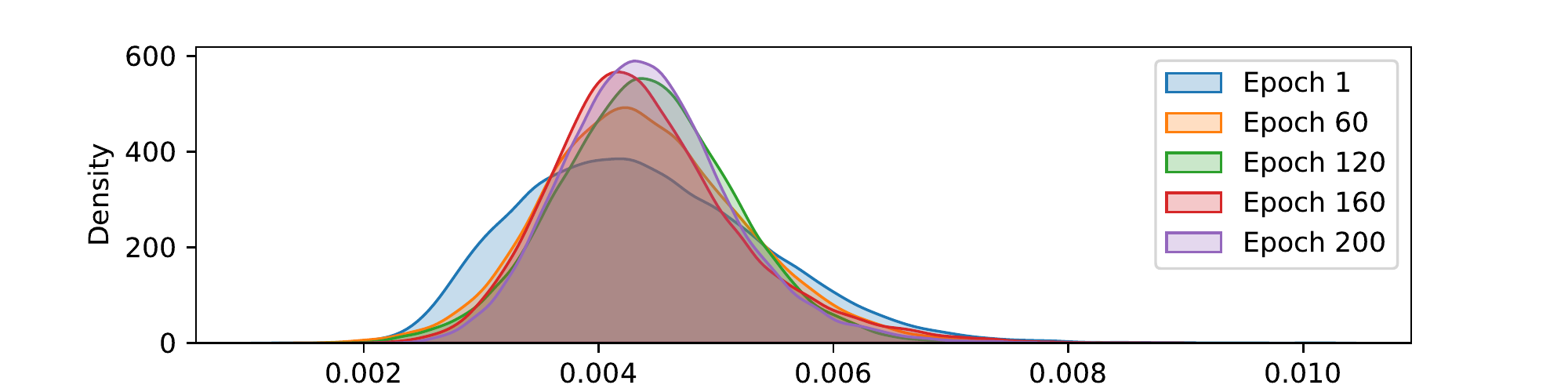}}
\caption{{KDE Smoothed Frequency of Values appeared in the P-vector versus training epochs.} }
\label{fig:KDE_Smoothed_Frequency}
\end{figure*}

Please refer to Figure~\ref{fig:Frequency} for the results of experiments carried out on CIFAR-10 and CIFAR-100 datasets using ResNet-50. To have a better view of the distribution drift, we further provide the KDE-smoothed frequency map of the $\mathcal{P}$-vector for wide-resnet training using CIFAR-100 dataset, as shown in Figure~\ref{fig:KDE_Smoothed_Frequency}.

\subsection{No Convergence found in comparisons between the Top Singular Vectors Other Than the $\mathcal{P}$-vectors}
\begin{figure*}
\centering
\subfloat[Top 1 singular vector.] {\includegraphics[width=0.48\textwidth]{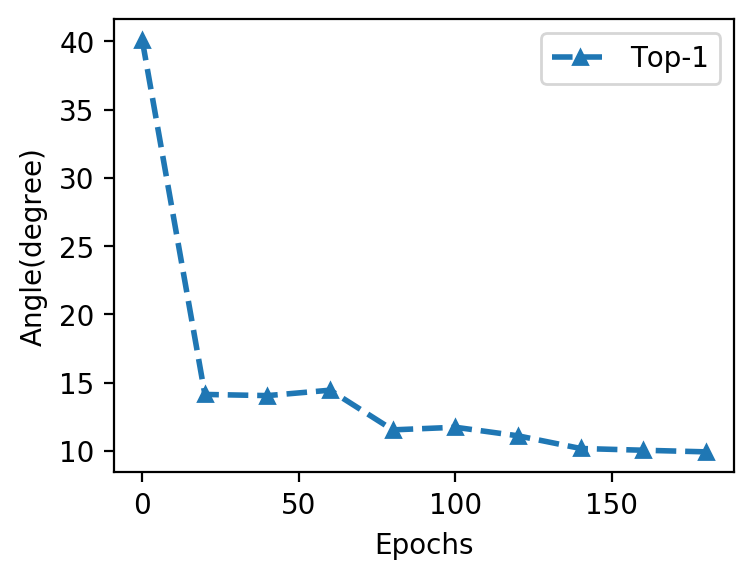}}
\subfloat[Top 2 singular vector.] {\includegraphics[width=0.5\textwidth]{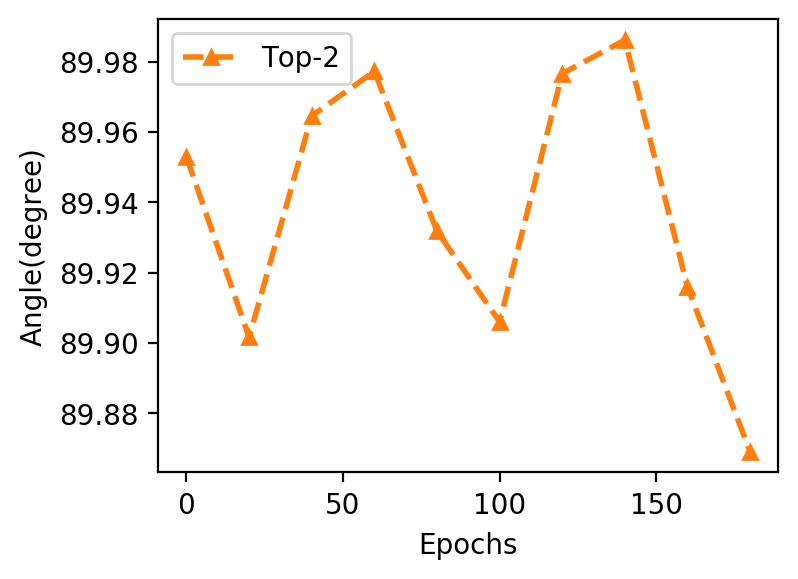}}\\
\subfloat[Top 3 singular vector.]
{\includegraphics[width=0.5\textwidth]{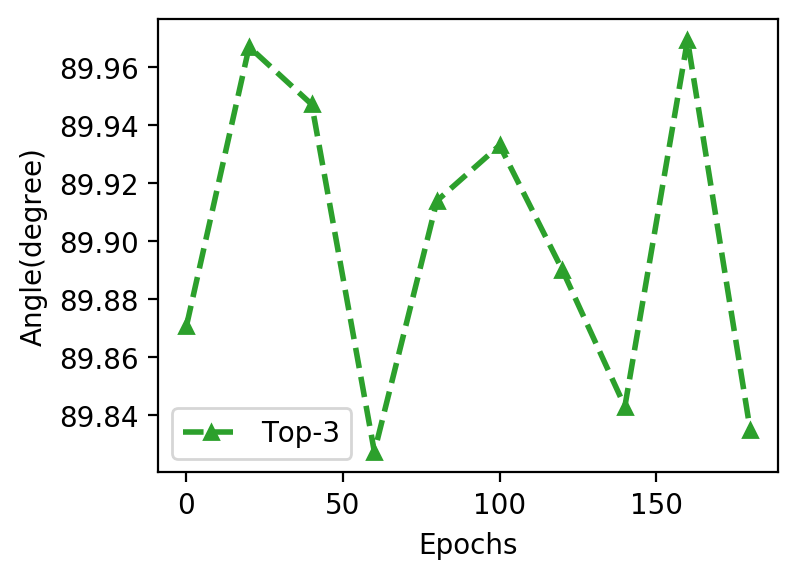}}
\subfloat[Top 4 singular vector.]
{\includegraphics[width=0.5\textwidth]{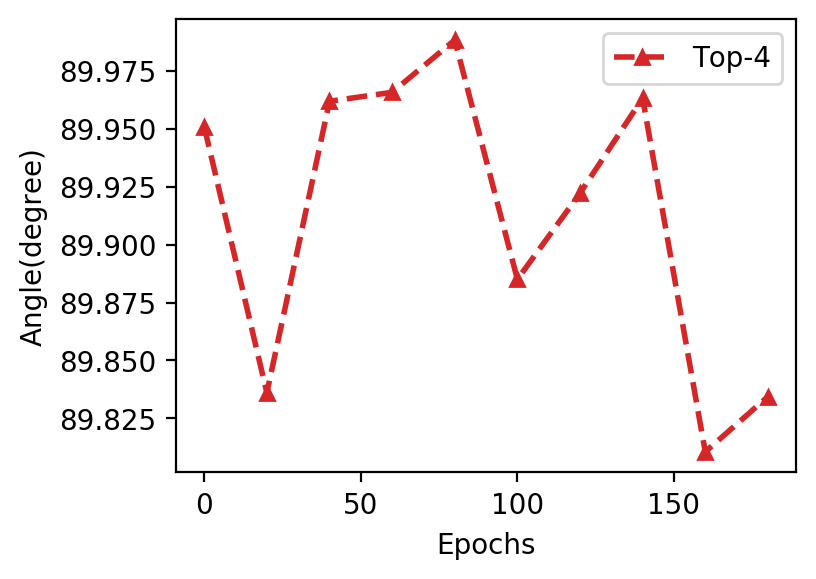}}\\
\subfloat[Top 5 singular vector.]
{\includegraphics[width=0.5\textwidth]{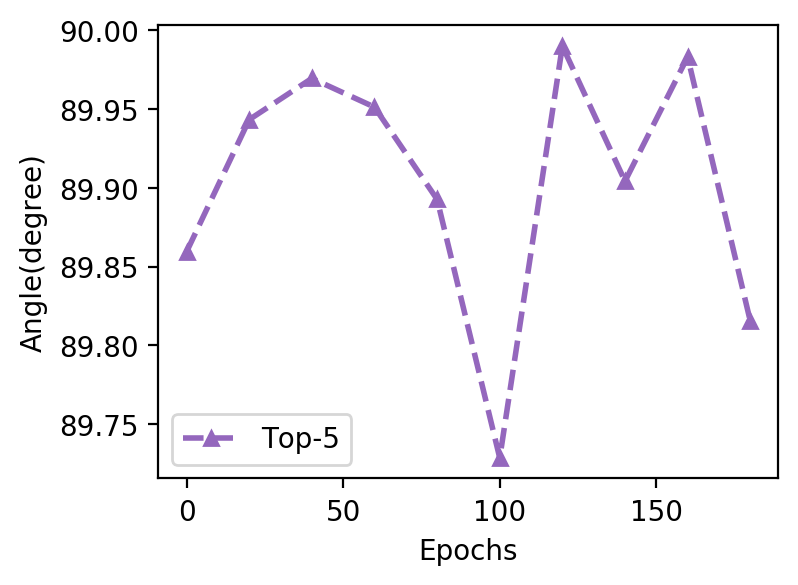}}
\subfloat[Top 6 singular vector.]
{\includegraphics[width=0.5\textwidth]{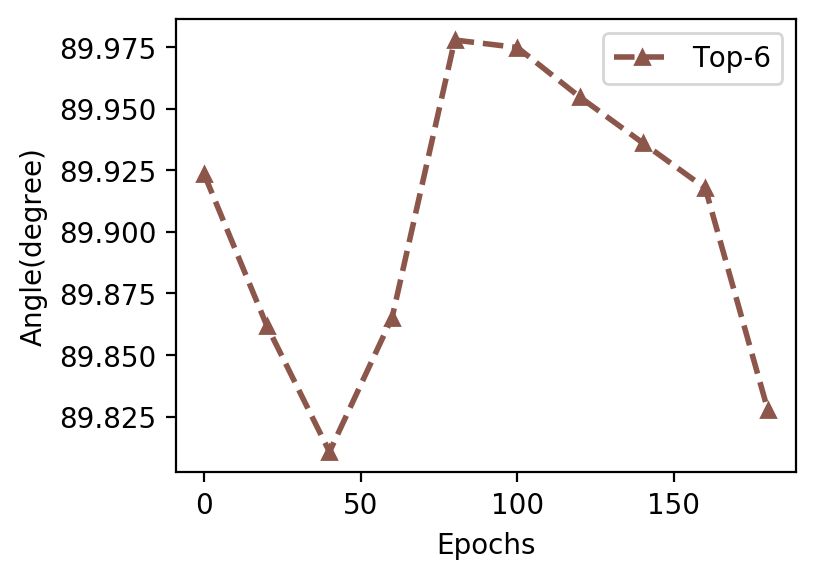}}
\caption{{Angles between the top-k left singular vector ($k=1$ is the $\mathcal{P}$-vector) of the training and well-trained models over the number of epochs in the training process (Resnet-50, CIFAR-10). Note the first plot point refers to the feature matrix after trained for one epoch.} }
\label{fig:Analysis_k}
\end{figure*}

Please refer to Figure~\ref{fig:Analysis_k} for the results of experiments carried out on CIFAR-10 datasets using ResNet-50.

\subsection{Log-Log Plots that correlate the $\mathcal{P}$-vector Angles and the Model Performance}
\begin{figure*}
\centering
\subfloat[VggNet16 (Train)]{\includegraphics[width=0.5\textwidth]{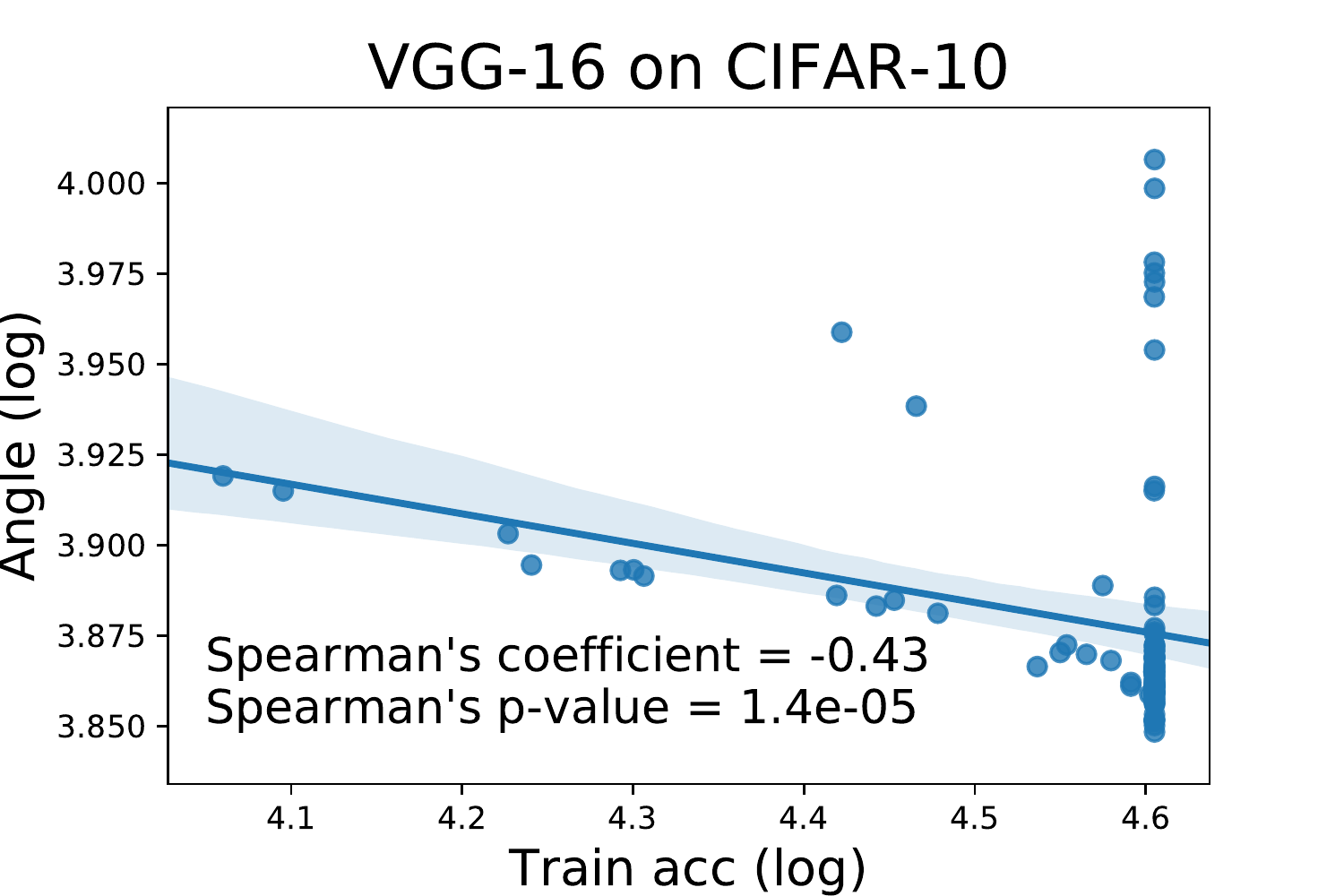}}
\subfloat[VggNet16 (Test)]{\includegraphics[width=0.5\textwidth]{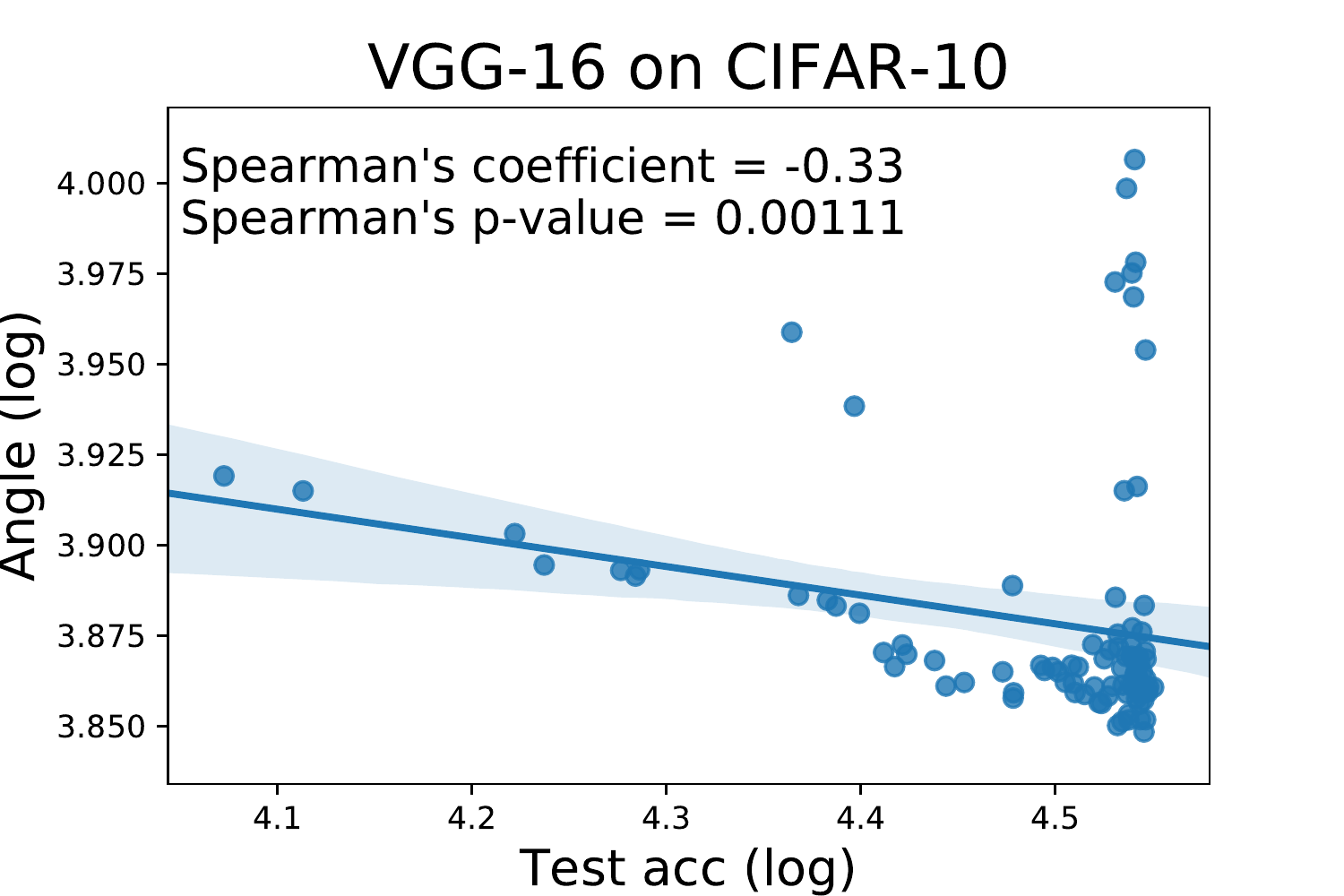}}\\
\subfloat[ResNet20 (Train)]{\includegraphics[width=0.5\textwidth]{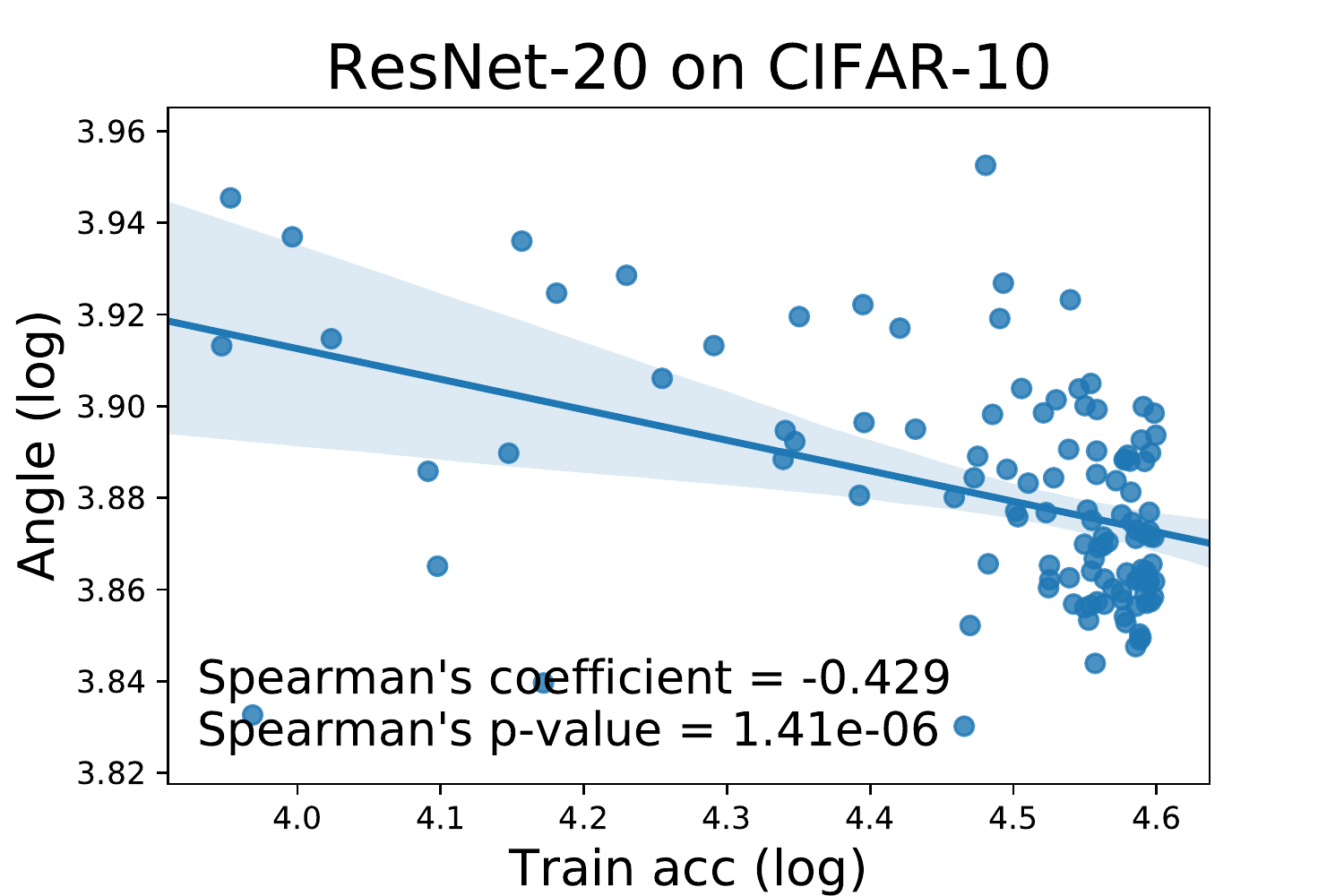}}
\subfloat[ResNet20 (Test)]{\includegraphics[width=0.5\textwidth]{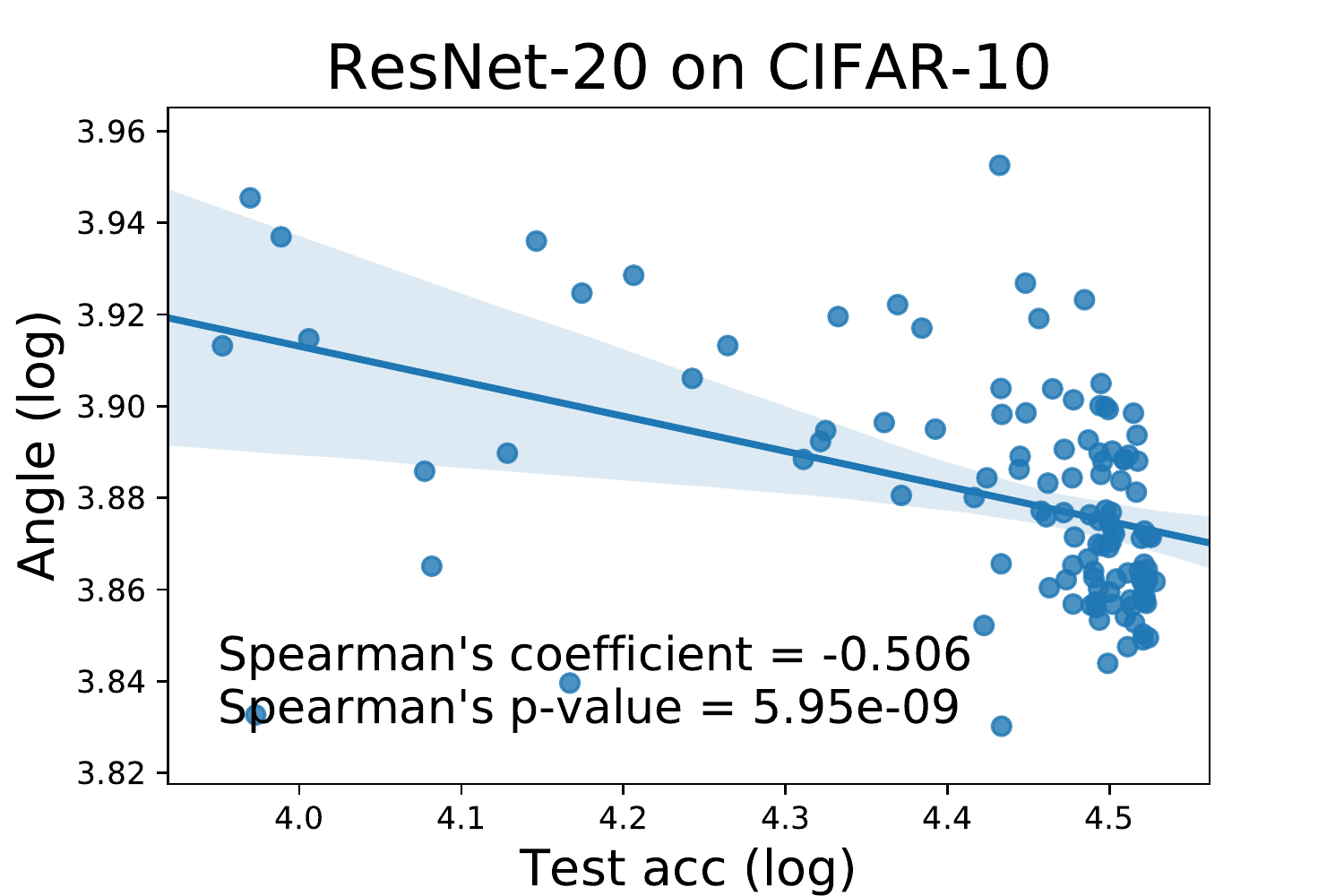}}\\
\subfloat[ResNet56 (Train)]{\includegraphics[width=0.5\textwidth]{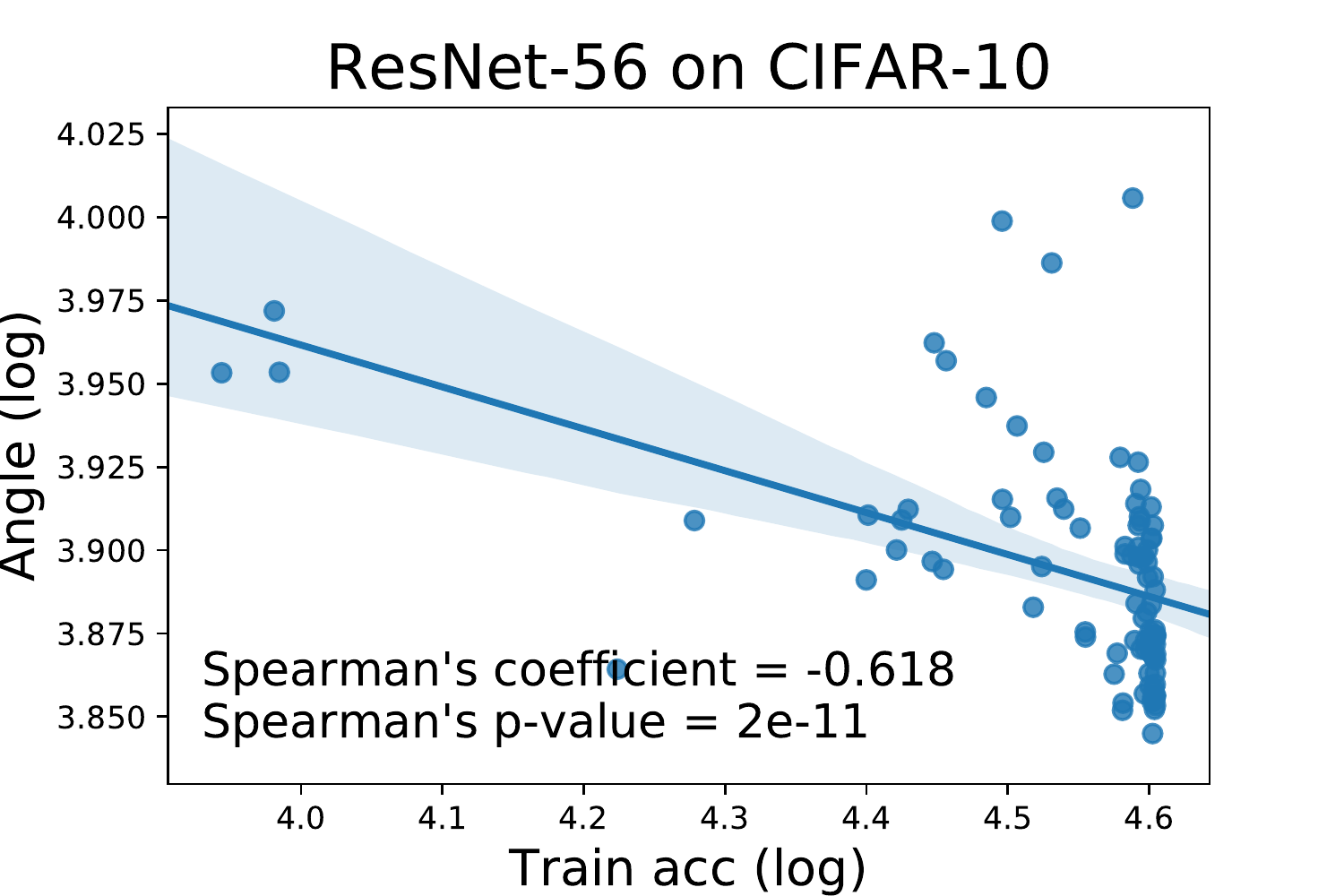}}
\subfloat[ResNet56 (Test)]{\includegraphics[width=0.5\textwidth]{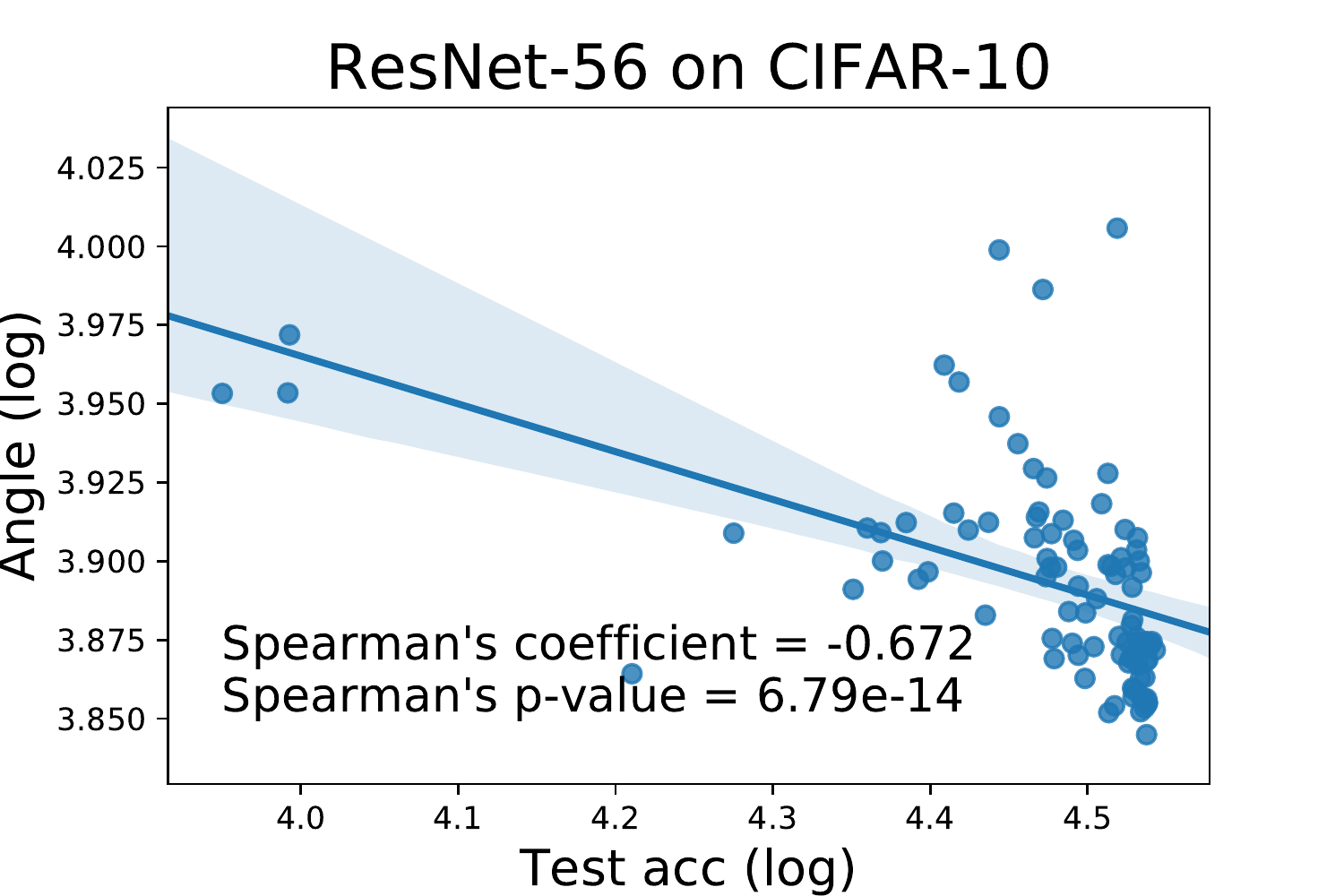}}
\caption{{Log-Log Plots: Correlations between the model performance (training and testing accuracy in log range) and the angels (in log range) between model and data $\mathcal{P}$-vectors using CIFAR-10 datasets.} }
\label{fig:append_log_claim3-2}
\end{figure*}

Please refer to Figure~\ref{fig:append_log_claim3-2} for the log-log plot of the results based on CIFAR-10 datasets.

\begin{figure*}
\centering
\subfloat[ResNet20 (Train)]{\includegraphics[width=0.5\textwidth]{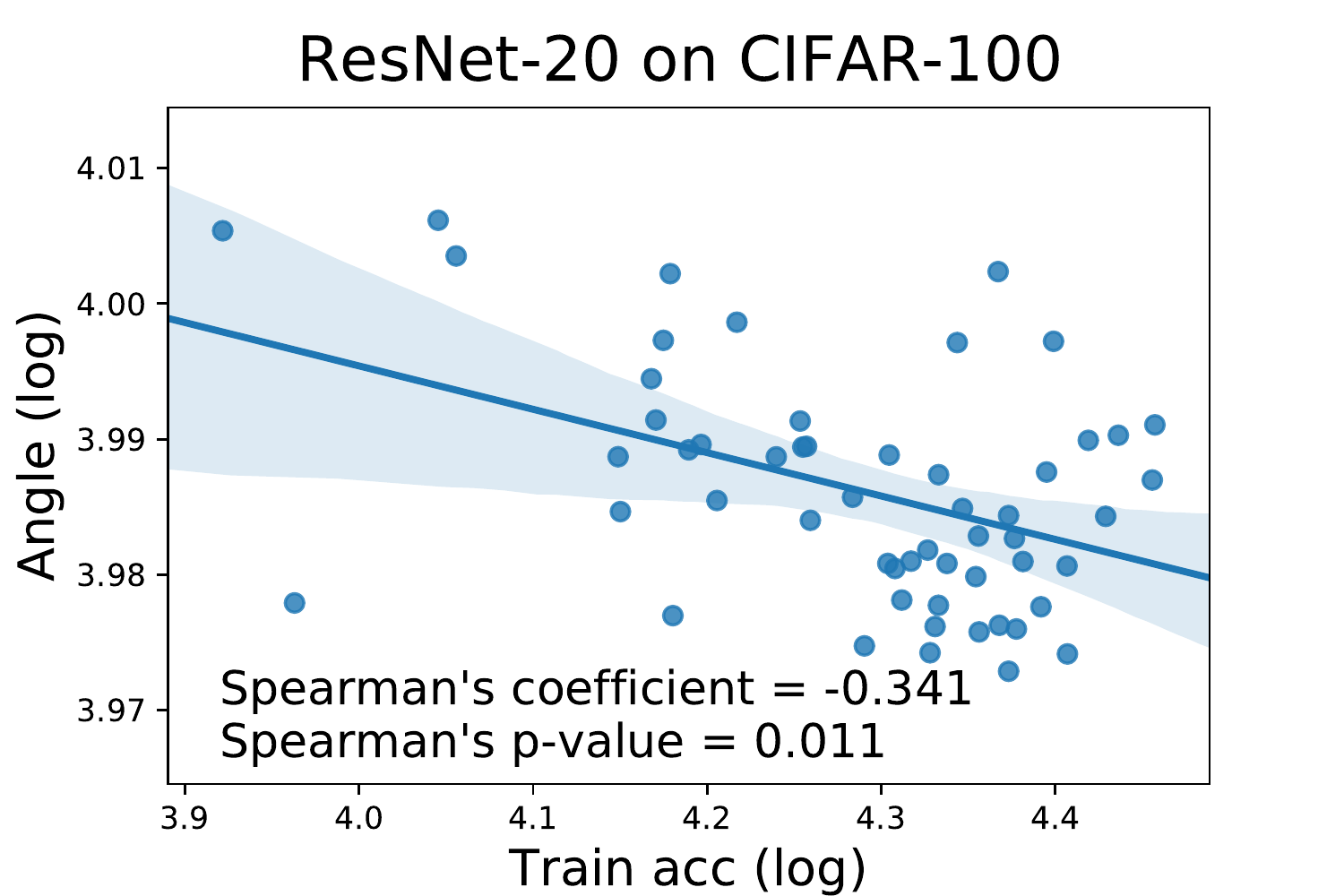}}
\subfloat[ResNet20 (Test)]{\includegraphics[width=0.5\textwidth]{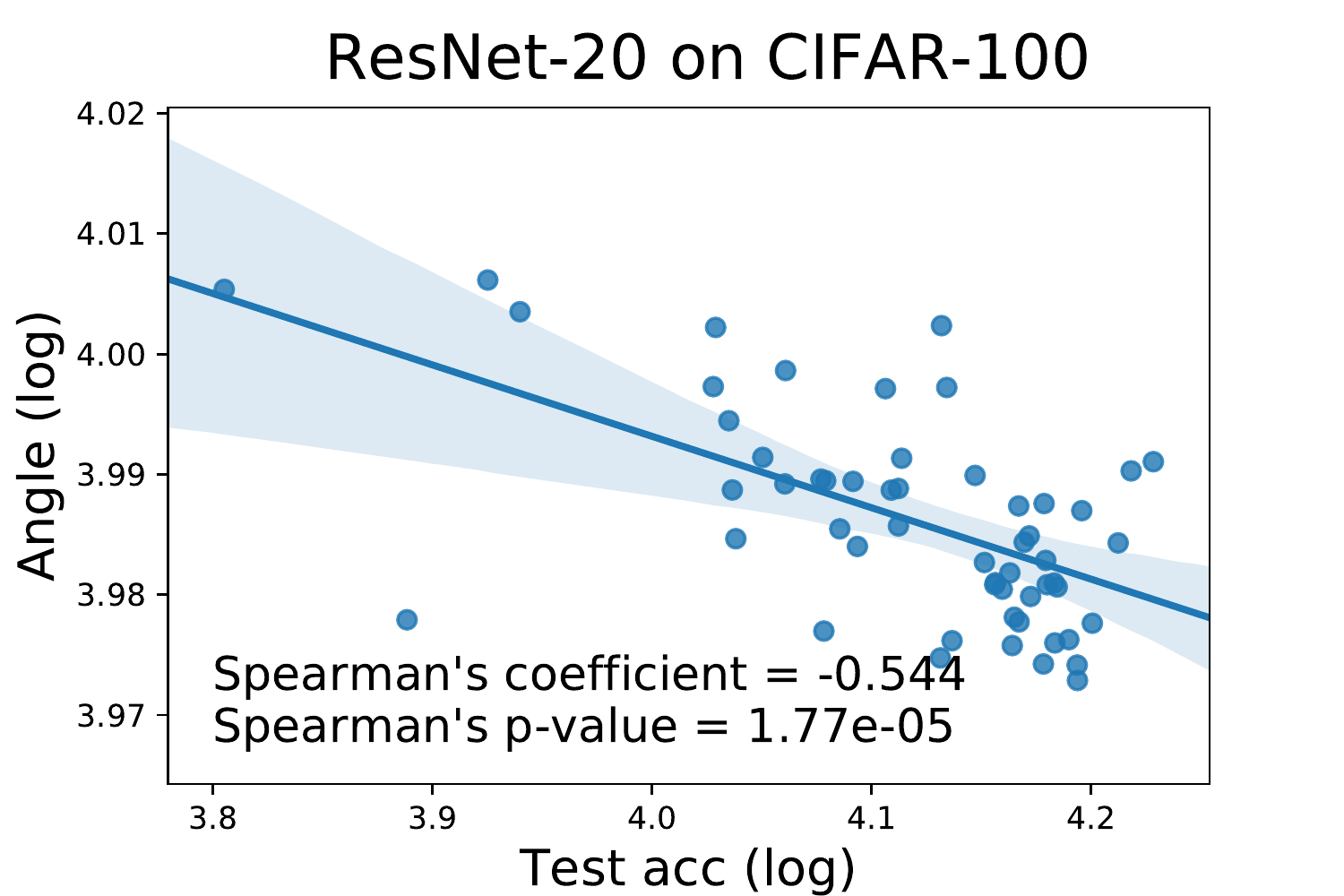}}\\
\subfloat[ResNet56 (Train)]{\includegraphics[width=0.5\textwidth]{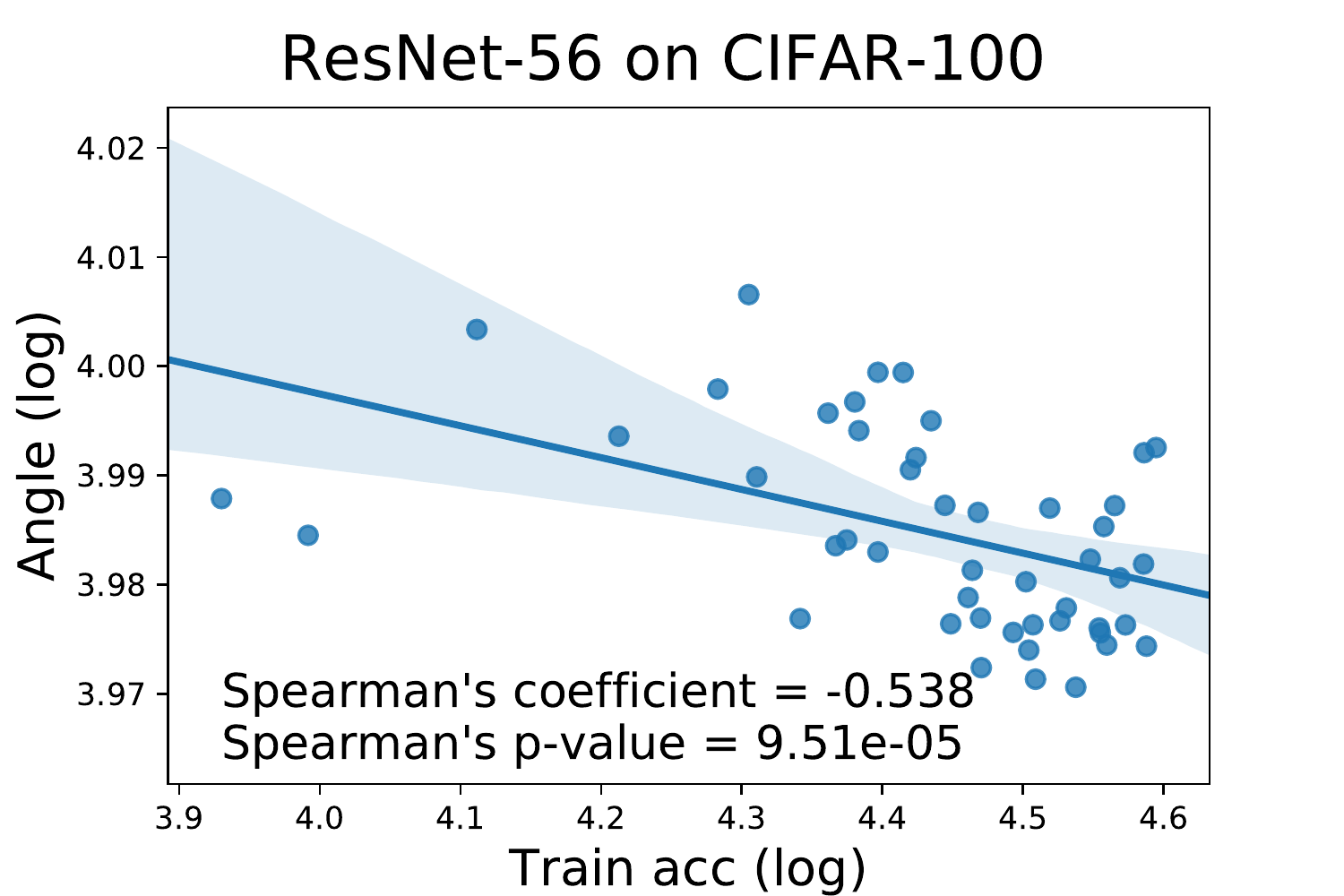}}
\subfloat[ResNet56 (Test)]{\includegraphics[width=0.5\textwidth]{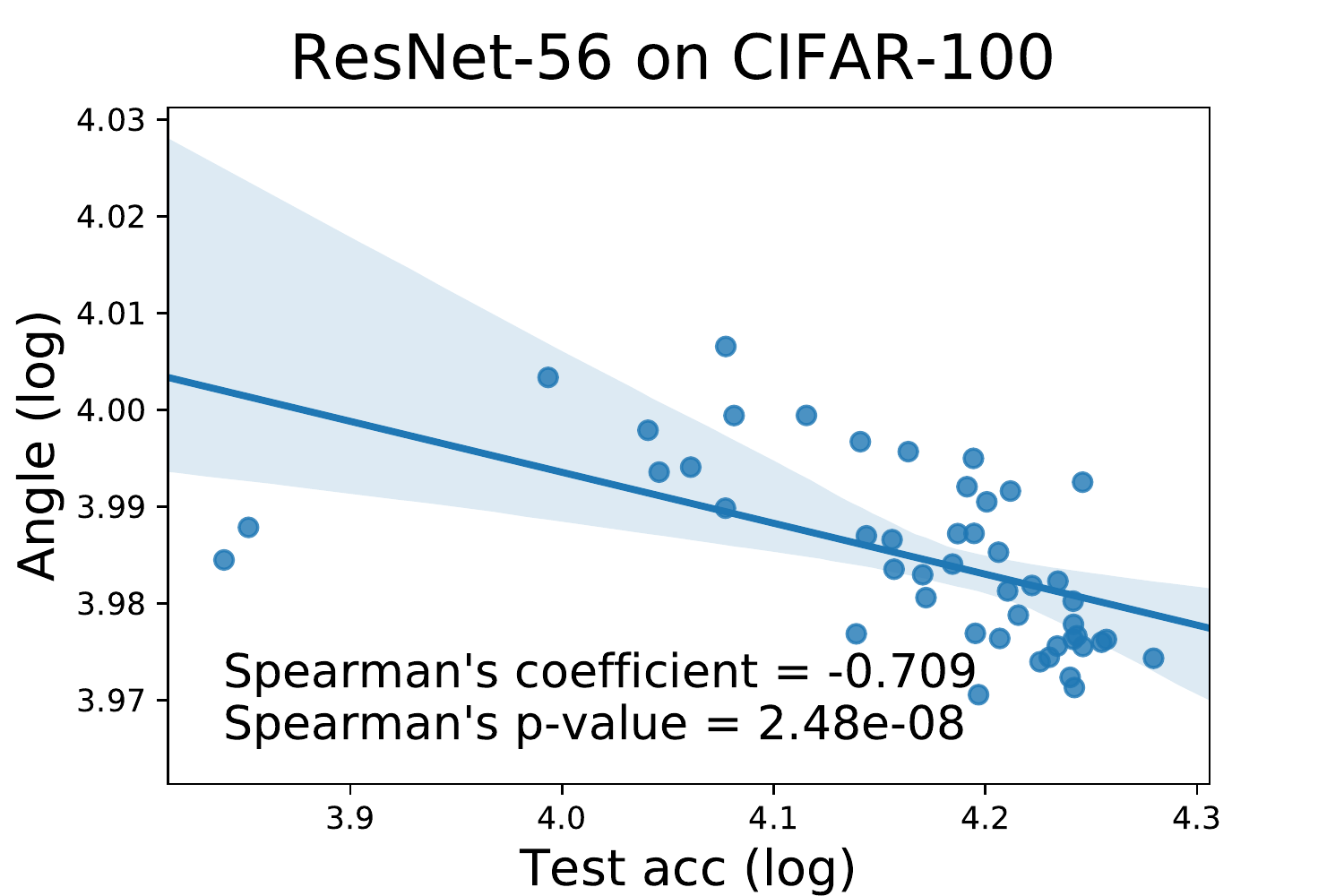}}\\
\subfloat[ResNet110 (Train)]{\includegraphics[width=0.5\textwidth]{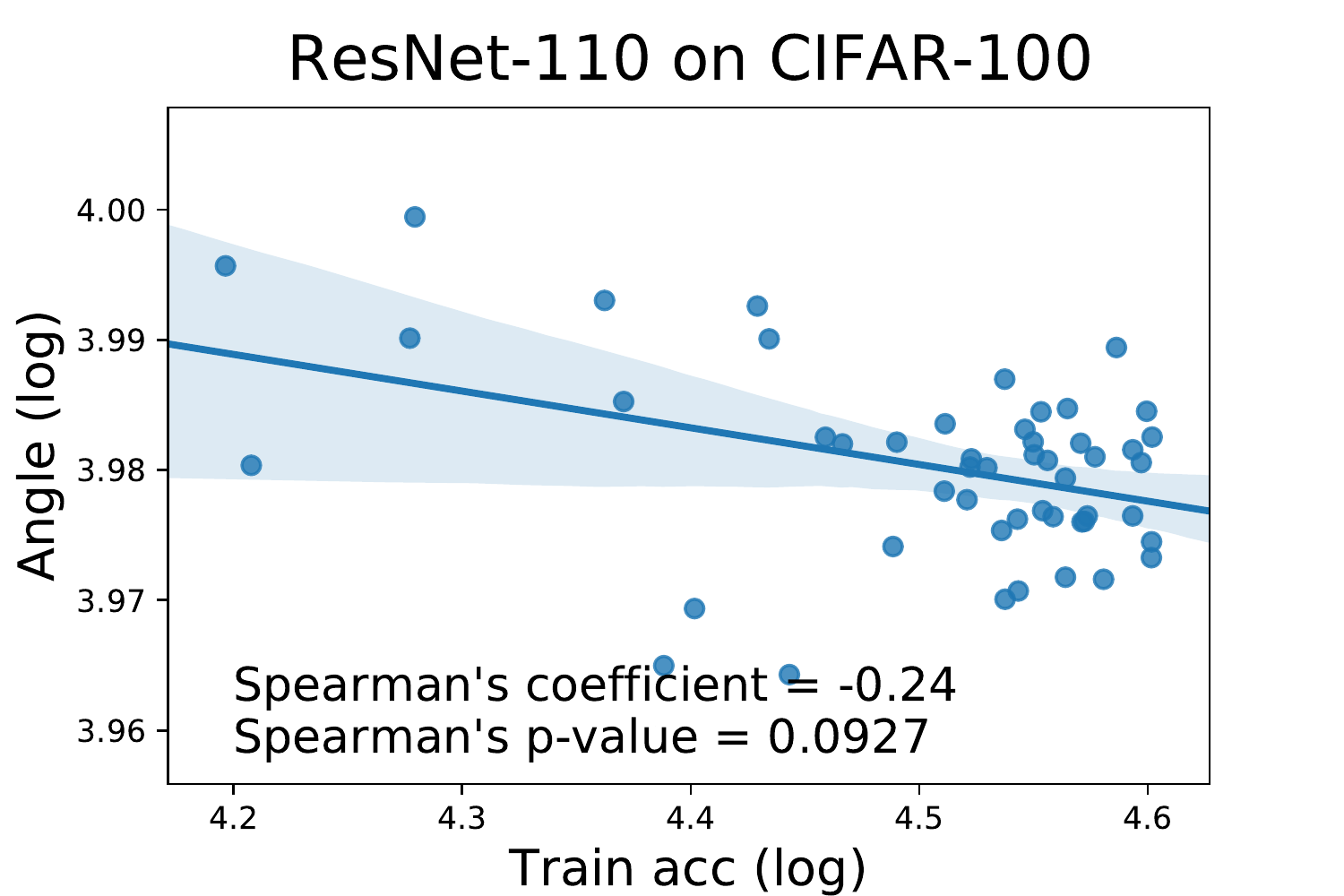}}
\subfloat[ResNet110 (Test)]{\includegraphics[width=0.5\textwidth]{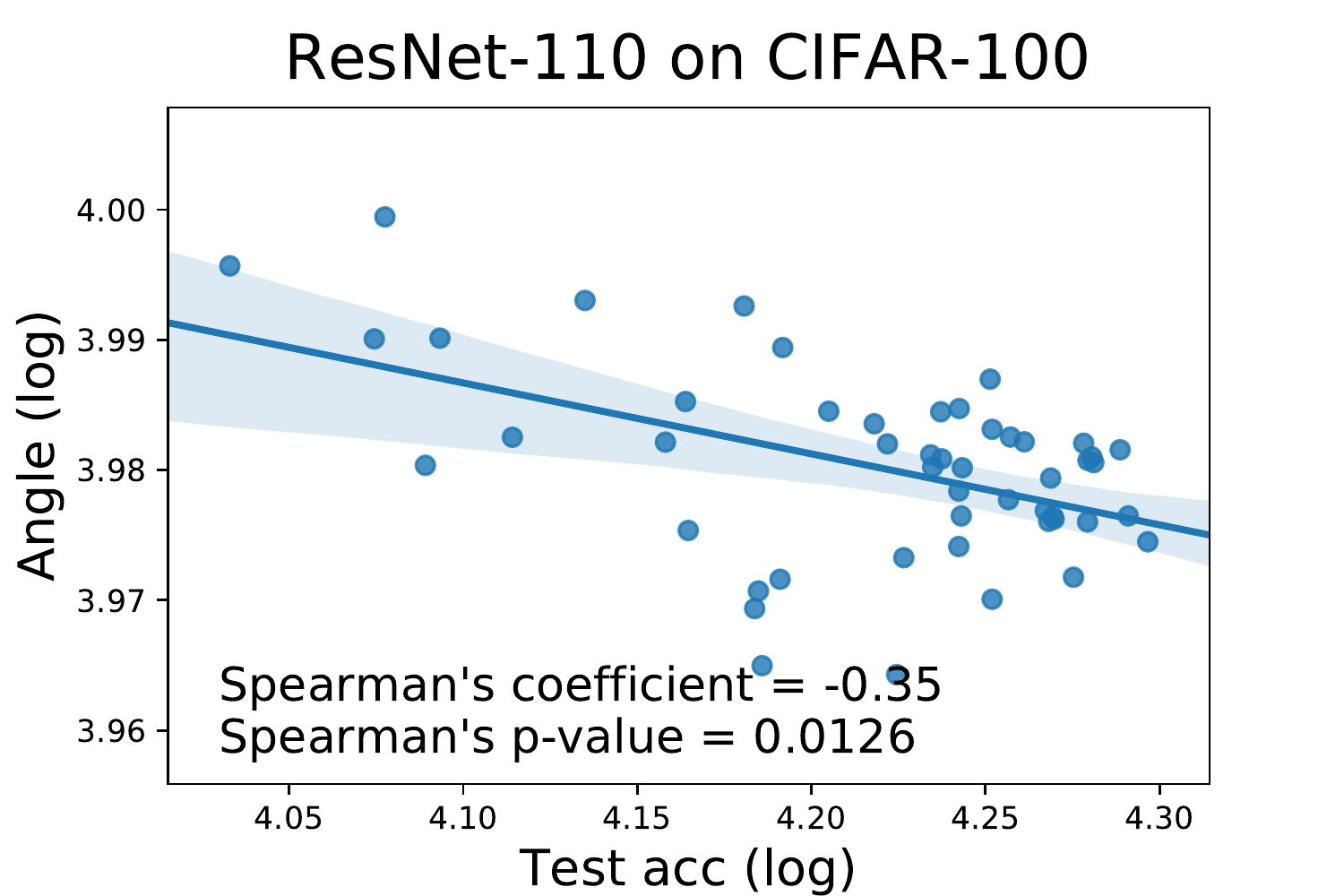}}
\caption{{Log-Log Plots: Correlations between the model performance (training and testing accuracy in log range) and the angels (in log range) between model and data $\mathcal{P}$-vectors using CIFAR-100 datasets.} }
\label{fig:append_log_claim3-2-2}
\end{figure*}


\subsection{Explained variance analysis of Top-k singular vector}

\begin{figure}
    \centering
    \includegraphics[width=0.5\textwidth]{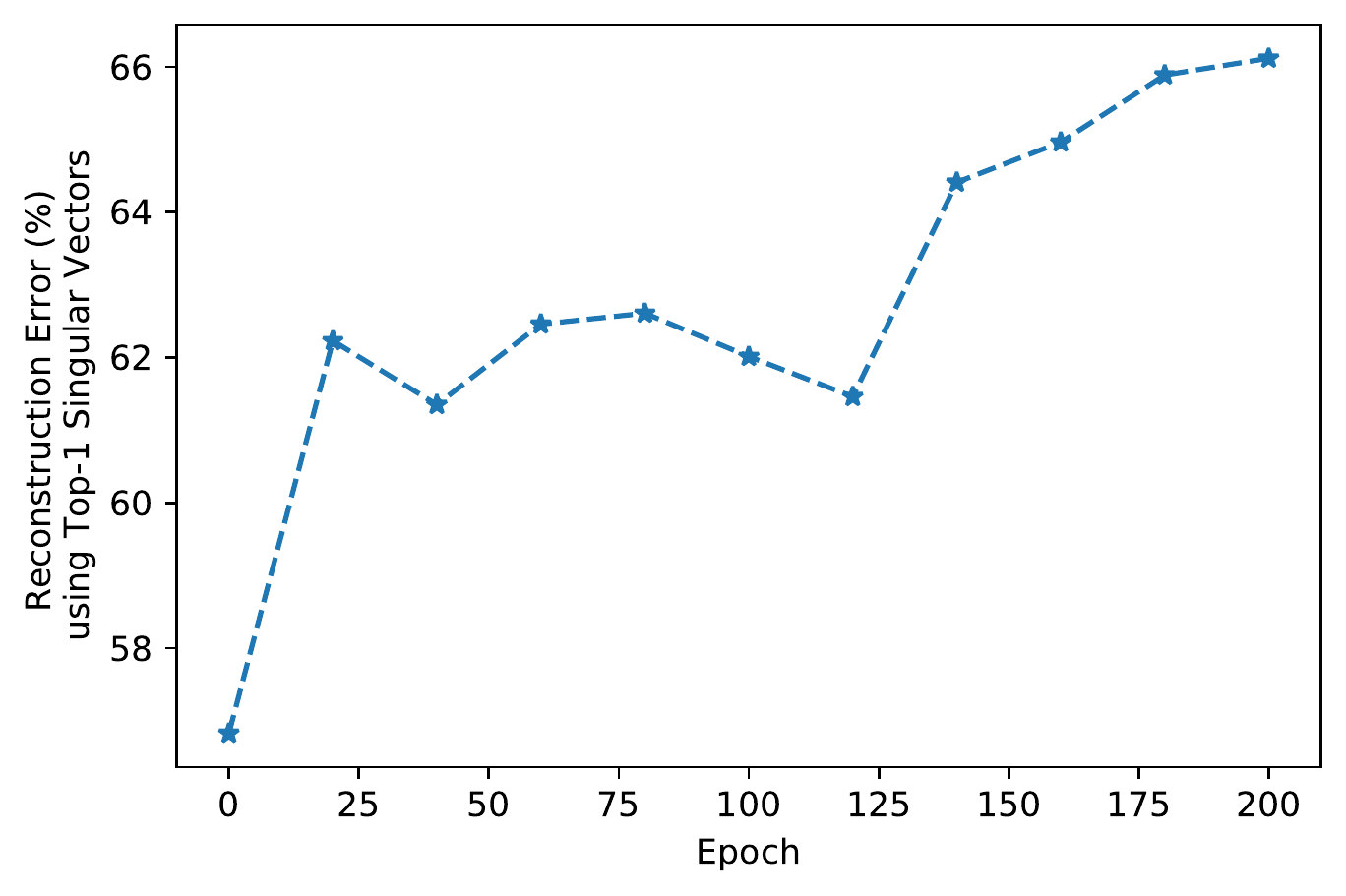}
    \caption{The reconstruction error of approximation with top-1 singular vector through the training process.}
    \label{fig:ratio}
\end{figure}

Explained Variances and the approximation error of the top-$k$ dimensional subspace $E=\|X-U_k \Sigma_k V_k^T\|_F^2$ of top-$k$ singular vectors are shown in Table. \ref{tab:exp_var}, where $X$ would be the feature matrices, and $U_k$, $\Sigma_k$, $V_k$ are the first  columns in the result of SVD, here we analysis only the changes of top $k=1$ singular vectors through different training checkpoints. The changing procedure for the reconstruction error of approximation is further shown in Figure.\ref{fig:ratio}. Also, the variance ratio analysis with various $k$ values is shown in Table.\ref{tab:var-ratio.}.
 
\begin{table}
\centering
\begin{tabular}{c|c|c|c}
\hline
 Epoch &       $E$ &  $|X_k|_F^2$&  Ratio of $E$ \\
\hline
     0 &  246.33 &     485.18 &       56.83 \\
    20 &  699.95 &    1115.90 &       62.23 \\
    40 &  674.80 &    1106.20 &       61.35 \\
    60 &  689.86 &    1115.35 &       62.46 \\
    80 &  851.61 &    1364.77 &       62.61 \\
   100 &  880.21 &    1434.66 &       62.01 \\
   120 &  867.66 &    1418.27 &       61.46 \\
   140 &  964.99 &    1507.03 &       64.41 \\
   160 &  968.36 &    1497.79 &       64.96 \\
   180 &  980.35 &    1494.50 &       65.89 \\
   200 &  980.86 &    1490.68 &       66.12 \\
\hline
\end{tabular}
\caption{Explained variance and reconstruction analysis of top-1 singular vector through the training process.}
\label{tab:exp_var}
\end{table}

\begin{table}
\centering
\begin{tabular}{c|c|c|c}
\hline
    k & Var. ratio(\%) & k & Var. ratio(\%) \\
\hline
     1 & 56.74  & 7 & 88.91 \\
     2 & 64.89  & 8 & 92.52 \\
     3 & 70.67  & 9 & 95.77 \\
     4 & 75.87  & 10 & 98.87 \\
     5 & 80.63  & 11 & 98.96 \\
     6 & 84.89  & 12 & 99.03 \\
\hline
\end{tabular}
\caption{Explained Variances of top-k singular vectors.}
\label{tab:var-ratio.}
\end{table}

\end{document}